\pdfoutput=1

\documentclass[11pt]{article}

\usepackage{acl}

\usepackage{times}
\usepackage{latexsym}

\usepackage[T1]{fontenc}
\usepackage{url}

\usepackage[utf8]{inputenc}

\usepackage{microtype}

\usepackage{inconsolata}

\usepackage{graphicx}
\usepackage{subcaption}
\usepackage{colortbl}
\usepackage{booktabs}
\usepackage{verbatim}
\usepackage{lipsum}
\usepackage{nicematrix}
\usepackage{amssymb}
\usepackage{enumitem}

\title{When Only Time Will Tell: Interpreting How Transformers Process\\ Local Ambiguities Through the Lens of Restart-Incrementality}

\author{Brielen Madureira$^{\mathbf{1,\thanks{Equal contribution.}}}$ \hspace{10mm}  Patrick Kahardipraja$^{\mathbf{2,\footnotemark[1]}}$  \hspace{10mm}  David Schlangen$^{\mathbf{1, 3}}$ \\
$^{\mathbf{1}}$Computational Linguistics, Department of Linguistics, University of Potsdam, Germany \\
$^{\mathbf{2}}$Fraunhofer Heinrich Hertz Institute, Berlin, Germany \\
$^{\mathbf{3}}$German Research Center for Artificial Intelligence (DFKI), Berlin, Germany \\
  \texttt{\{madureiralasota,david.schlangen\}@uni-potsdam.de}, \\ \texttt{patrick.kahardipraja@hhi.fraunhofer.de}}

\begin{document}
\maketitle
\begin{abstract}
Incremental models that process sentences one token at a time will sometimes encounter points where more than one interpretation is possible. Causal models are forced to output one interpretation and continue, whereas models that can revise may edit their previous output as the ambiguity is resolved. In this work, we look at how restart-incremental Transformers build and update internal states, in an effort to shed light on what processes cause revisions not viable in autoregressive models. We propose an interpretable way to analyse the incremental states, showing that their sequential structure encodes information on the garden path effect and its resolution. Our method brings insights on various bidirectional encoders for contextualised meaning representation and dependency parsing, contributing to show their advantage over causal models when it comes to revisions.\footnote{Our code is available at: \url{https://github.com/briemadu/restart-inc-ambiguities}}
\end{abstract}

\section{Introduction}
\label{sec:intro}
\textit{This is the honey}... even if we stopped mid-sentence here, you would likely have created a partial interpretation of this prefix considering \textit{the honey} as (the beginning of) a noun phrase. It could have many continuations, \textit{e.g.}~\textit{that skunks like}, or \textit{produced by stingless bees}. But what if the next token is another noun, as \textit{bee}? A semantic parser would have to revise its previous hypothesis to accommodate the fact that \textit{honey} has become a modifier of \textit{bee}.

\begin{figure}[ht]
    \centering
    \includegraphics[trim={0cm 3cm 0cm 0cm},clip,width=\columnwidth,page=1]{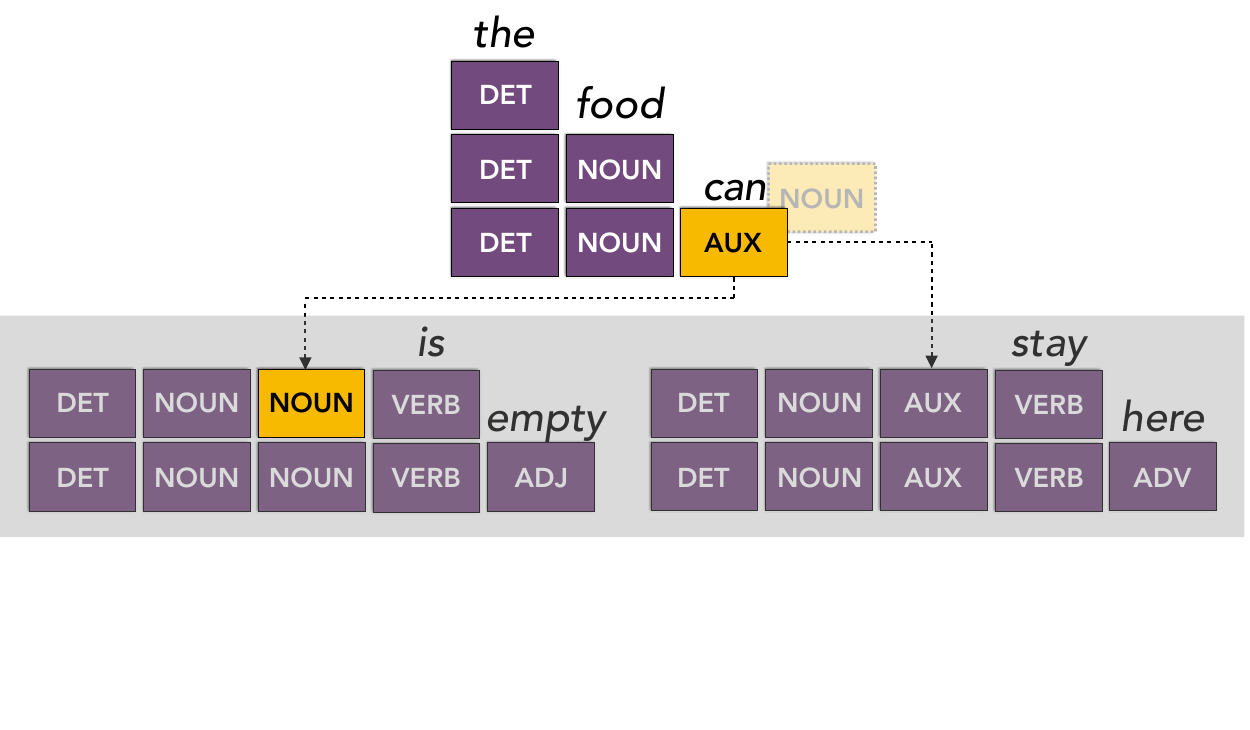}
    \caption{A prefix with multiple valid continuations. A causal decoder is forced to output only one POS-tag for the token \textit{can} at this point and cannot change it anymore, even if its internal state encodes the local ambiguity. In contrast, a restart-incremental model can perform revisions and would thus be able to recover if the selected label turned out to be incorrect (as in left).}
    \label{fig:example}
\end{figure}

Bidirectional NLP models (\textit{i.e.}~those that encode linguistic input using both its left and right context) have transformed the field of computational linguistics. But that has come at the cost of cognitive plausibility in various aspects, in particular establishing a disregard for language's temporal structure. While humans process language one increment at a time \citep{marslen1973linguistic,altmann1988interaction,levelt1993speaking}, BiLSTMs \citep{graves2005framewise} and Transformers \citep{vaswani2017attention} have mechanisms that rely on access to the complete input sequence, making them unsuitable as off-the-shelf components for incremental applications. Although their unidirectional counterparts, LSTMs and autoregressive (\textit{aka} causal) Transformers, are equipped with the possibility of incremental decoding (making predictions relying only on left context), their token representations are static and thus not updated as incoming tokens arrive \citep{eisape-etal-2022-probing}, because the underlying encoding is unidirectional. This places them at the disadvantage of not being able to revise, which is a desired property to recover from mistakes or local ambiguities \citep{schlangen2011general,madureira-etal-2023-road}, as shown in Figure \ref{fig:example}.

The restart incremental (RI) paradigm\footnote{Or qualitative incrementality \citep{kilger1995incremental}.}  \citep{schlangen2011general} circumvents this issue by adding an interface upon any model, making it work incrementally by processing prefixes from scratch whenever a new input increment arrives \citep{beuck-etal-2011-decision}. Even though it is computationally costly, RI has been effectively studied in simultaneous MT \citep{arivazhagan-etal-2020-translation,sen-etal-2023-self}, dialogue systems \citep{khouzaimi-etal-2014-easy}, disfluency detection \citep{chen-etal-2022-teaching} and NLU pipelines \citep{rafla2019incrementalizing}. Besides, it has the advantage of incorporating recent increments into past predictions, revising hypotheses when deemed necessary.

A recent line of work has investigated restart-incrementality for sequence labelling by profiling bidirectional encoders \citep{madureira-schlangen-2020-incremental,kahardipraja-etal-2021-towards} and modelling and evaluating revision policies \citep{kahardipraja-etal-2023-tapir,kaushal-etal-2023-efficient,madureira-etal-2023-road}. Still, evaluations so far have been performed in a black box fashion: Only the relations between input tokens and output labels have been considered. The output labels as top positions of the softmax operation 
are only one product of many intermediate computations that have not been examined. 

Therefore, we still do not know why RI models revise when they revise. What happens internally in the model's mechanisms to encode the need to edit previous outputs? How can (static) causal representations be affected under RI bidirectionality? Can we predict whether the model will recover from a wrong interpretation? To answer such questions, we need a shift to glass box interpretability methods that can shed light on the dynamics of updates in the internal states which lead to output revisions. Linguistically motivated analyses are also required to examine the behaviour of bidirectional models with respect to specific phenomena known to cause reinterpretations, like garden path constructions.

Thus, our present contributions to make RI  models more transparent and explainable are \texttt{(a)} a formalisation of RI sequential processors as transition systems that create structured step-by-step constructions not present in causal models; \texttt{(b)} a proposal of interpretability methods for the internal mechanisms of RI models; and \texttt{(c)} an analysis of the strategies employed by various models on stimuli containing local ambiguities, for which a well-defined motivation for reanalysis is known.

\section{Related Work}
\label{sec:lit}

\paragraph{Alternatives to Restart Incrementality} Monotonic decoding is advantageous to avoid output instability but comes with the downside of not recovering if there is a genuine reason to revise. Some attempts to overcome this issue are adapting back-propagation to update the internal representation of the output \citep{qin-etal-2020-back} or gradient-based methods to update the cached internal representations \citep{yoshida-gimpel-2021-reconsidering-past} without changing the model's parameters. Although this alleviates the strict monotonicity, it requires adapting the model's implementation. RI, on the other hand, is available to anyone in possession of any model, as it only requires re-running it as is each time. Via recomputations, bidirectional models innately incorporate new increments into its states and revise if needed \citep{kahardipraja-etal-2023-tapir}. For causal models, revisions can occur if one resorts to beam search methods (see \emph{e.g.}~\citet{leblond-etal-2021-machine}), traversing the sequence of static states generated at each token to make predictions at each time step. However, this technique does not reside within the model's own mechanisms. Besides, if the correct sequence falls out of the chosen beam size at some point, the needed revision signal will be lost. Performing the search requires enough memory and additional processing time, which may become an issue for longer sentences. 

\paragraph{Benefits of Bidirectionality} Several works have studied the effects of bidirectionality in LMs. \citet{artetxe-etal-2022-role} show that bidirectional attention is beneficial for some tasks like infilling and fine-tuning (but detrimental for others like next token prediction), while bidirectional context is beneficial when used in conjunction with bidirectional attention. \citet{springer2024repetition} propose using repetitions to allow that early tokens incorporate later tokens in their representations in LLMs. \citet{dukic2024not} also show advantages of applying layer-wise removal of the causal mask during LLM fine-tuning.

\paragraph{Interpretability} Various methods have been proposed to analyse how neural networks encode linguistic information \citep{belinkov-glass-2019-analysis}, with Transformers front and centre, \textit{e.g.}~in how information flows in its self-attention \citep{abnar-zuidema-2020-quantifying} and how its predictions are refined or affected by previous tokens layer after layer \citep{geva-etal-2022-transformer, ferrando-etal-2023-explaining,belrose2023eliciting,din2023jump}. Voluminous BERTology works exist  \citep{clark-etal-2019-bert,rogers-etal-2020-primer}, in particular investigating its attention mechanisms, what it means for an embedding to be contextualised instead of static and what layers encode what type of linguistic structure. Another angle is to look at how linguistic information is dynamically learnt throughout the training regime of LMs \citep{saphra-lopez-2019-understanding,chen2023sudden}.
The realm of incremental processing has also been active. \citet{ulmer-etal-2019-assessing} propose methods to interpret how RNNs incrementally encode, integrate and retain information. The incremental aptitude of neural networks to encode syntactic information has also been examined with the aid of psycholinguistic experimental methodologies \citep{futrell-etal-2019-neural, wilcox-etal-2021-targeted}. 

\paragraph{Cognitive Motivation} An evident test bed are stimuli containing local ambiguities to evaluate if models exhibit garden path effects. In psycholinguistics, a traditional debate holds between the early-commitment with eventual reanalysis \textit{vs}.~beam search approaches, where multiple hypotheses are kept in parallel, known as two and one-stage accounts, respectively \citep{van2021single}. In this sense, the RI paradigm can be viewed as a two-stage account performing multiple forward-reanalyses \citep{frazier-rayner-1982}.

\paragraph{Transformers and Ambiguities} Recent findings point to models encoding multiple hypotheses in face of local ambiguities. \citet{aina-linzen-2021-language} probe to what extent autoregressive LMs encode multiple syntactic analyses by assessing the probability assigned to each interpretation as they generate continuations of a prefix, finding that multiple interpretations are followed in parallel. Methods relying on surprisal theory show that the magnitude of the garden path effect in autoregressive and RNN LMs underestimate human behaviour \citep{van2021single,arehalli-etal-2022-syntactic}. Other works study BERT-based models using differences in surprisal and attention between unambiguous and locally ambiguous sentences \citep{lee2022bert,lee2023decoding}.

\paragraph{Other analyses} Other glass box analyses have been put forward. \citet{irwin-etal-2023-bert} use probing tasks with garden paths to study how the BERT family assigns semantic roles in QA tasks. \citet{jurayj-etal-2022-garden} define vector similarity methods as an improvement over surprisal to shed light on how GPT-2 traverses garden paths, finding periods of ambiguity in the hidden states that do not always surface. \citet{lindborg2021meaning} study how meaning is built word by word by GPT-2 looking at the connection between the size of the updates in its output activations and the N400 psychometric in humans.

\paragraph{Speculating Continuations} Through structural probes, \citet{hewitt-manning-2019-structural} show that it is possible to recover syntax trees from ELMo and BERT embeddings using linear transformations. \citet{eisape-etal-2022-probing} extend the probe to incremental settings and conclude that the internal representations of autoregressive LMs encode syntactic uncertainty that can be explored by future tokens, which would be a reason why such models perform well even without access to future words. It is also possible that monotonic models work well due to speculation about the future. \citet{pal-etal-2023-future} use GPT's internal states to predict future tokens, finding that some layers partially anticipate subsequent tokens. But this speculation may lead to wrong paths and is not always desirable. \citet{kitaev-etal-2022-learned}, for instance, explore non-speculative incrementality to induce syntactic representations free of speculation.

As we see, in previous studies, analyses happened either at autoregressive mode or with bidirectional access to the whole sequence. Our focus is to tailor interpretability methods for inspecting the emergent properties of bidirectional models under \textit{restart incrementality}, beyond output labels.

\section{Formalisation}
\label{sec:formalisation}
A restart-incremental model is constructed upon an underlying (non-incremental) model, making it perform a sequence of re-computations as the input is processed increment by increment \citep{schlangen2011general,beuck-etal-2011-decision}. We propose a general formalisation of this procedure, detail its structures for sequential processing and discuss their interpretation for bidirectional models. 

\paragraph{Restart-Incrementality} Let $M: w \mapsto o$ be a (non-incremental) model that maps an input $w$ to an output $o$ by computing internal states $s$. Generally speaking, restart-incrementality is an interface $I$ around $M$, such that $I(M)$ can be fed individual tokens $w_i$, which---until the interface is reset---are taken as continuations of the sequence of tokens fed so far. Hence, at time step $t$, $I(M)$ will be provided with the token $w_t$. Internally, the interface assembles the \textit{prefix} $w_1, \dots, w_t$ (denoted as $w^t$ below), which it feeds to $M$, to compute the corresponding sequences $s^t = (s_1^t, \dots, s_t^t)$ and $o^t = (o_1^t, \dots, o_t^t)$.\footnote{By growing incrementally, each prefix changes the context of each prior input token, which is why the elements of state and output sequences need to be indexed with the timestep in superscript; \emph{e.g.} $s_1^4$ is the internal representation of token 1, at the time when the first 4 tokens were available.}

This way, from the perspective of $M$, it is always a sequence of tokens that is being processed from scratch, independently from any prior calls of $M$, while from the perspective of $I(M)$, with each call a single token is added. While for the further downstream processing, only the output revisions (that is, the elements where $o^{t+1}$ differs from $o^t$) may be of interest,\footnote{As in the IU model of \citet{schlangen2011general}, implemented in InproTK \citep{kennington-etal-2014-inprotks}.} for the purpose of our analysis the sequences 
$s^1, s^2, \dots, s^n$ corresponding to each prefix are kept in memory by $I(M)$. 
$I(M)$ can hence be regarded as a transition system where states are updated based on the action of recomputing with the integration of a new input increment.

\vspace{0.3cm}
\paragraph{Triangular Structures} To represent the memory built by $I(M)$, we can extend the 2D chart of outputs in \citet{madureira-etal-2023-road} with a third dimension, filling it with state sequences as follows: 

\vspace{0.5cm}
\begin{center}
    $\begin{NiceArray}{ccccc|ccccc} 

        \toprule
        & & & & w_1                         & s_1^1     &           &           &           &         \\ 
        & & &  w_1 & w_2                    & s_1^2     &   s_2^2   &           &           &         \\ 
        & & w_1 & w_2 & w_3               & s_1^3     &   s_2^3   &  s_3^3    &           &         \\ 
             &   \iddots  &  \iddots   &  \iddots   &                           \vdots    & \vdots    &   \vdots  &  \vdots   &  \ddots   &         \\
        w_1 & w_2 & w_3 & \ldots & w_n       & s_1^n     & s_2^n     &  s_3^n    & \cdots    & s_n^{n} \\   \bottomrule
    \end{NiceArray}$
\end{center}
\vspace{0.5cm}

\begin{figure}[t]
    \centering
    \includegraphics[trim={0cm 0.85cm 14cm 0cm},clip,width=\columnwidth,page=1]{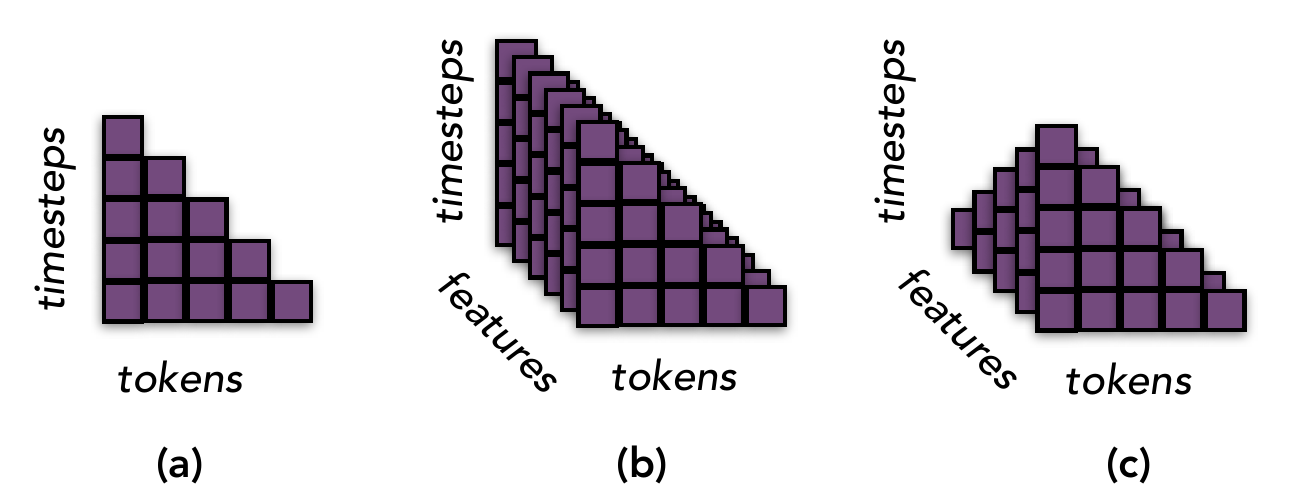}
    \vspace{0.1cm}
    \caption{Triangular structures representing states built step by step in restart-incremental sequential processing.}
    \label{fig:pyramids}
\end{figure}

The right portion represents how the states evolve from time step to time step for an input sequence of $n$ tokens. The last state sequence coincides with the sequence $M$ creates when the full input is available. 
Figure \ref{fig:pyramids} illustrates three types of resulting structures. Although they are actually multidimensional arrays, with some abuse of nomenclature, we will reference them by the triangular prisms they resemble. In the right triangle in (a), each token is assigned one value per time step.
In the right triangular prism (b), each time step produces a vector with a fixed number of dimensions for each token (for instance, a probability distribution over labels or an embedding). In the truncated triangular prism (c), at each time step, a vector the same size of the current prefix is built for each token (for example, attention scores or dependency arcs over the input). 

These structures allow us to inspect emergent properties by comparing time steps. We can specifically look at the dynamics of the $s^j_k$ for a fixed $k$, \textit{i.e.}~the states corresponding to the same input token, but over the re-computations at each time step.
Note that, when models are bidirectional, this is a much richer process than what happens in autoregressive or monotonic decoders, where elements in the main diagonal are computed once and kept fixed for all subsequent steps. Here, all $s$ can keep changing as right context gets integrated.\footnote{The main diagonal contains another form of causality: Although its elements also see no right context, they are computed considering bidirectional representations of the left tokens available. In causal models, all states are unidirectional.}

\vspace{0.3cm}
\section{Method}
\label{sec:method}
We now introduce our method and motivate its interpretations. Once we have extracted states as a RI model processes a sentence, we can analyse the dynamics of this process, \textit{i.e.}~how do these vectors evolve over time. The 3D structures (b) and (c) from Figure~\ref{fig:pyramids} can be converted into 2D as in (a) if we apply a metric over the features dimension, summarising it into one value (\textit{e.g.}~Shannon's entropy or divergence over distributions and similarity or distance metrics for embeddings). With such scores, we can then examine whether the behaviour of their variation aligns with output edits.

If we look at columns of the triangular structures, we can analyse a sequence of states assigned to one token step by step. If we consider the rows, we see the effect that the recently added token had on the current prefix. This analysis can in principle be applied to any sentence. However, to shed light on how restart-incremental models edit the output and how they engage in reinterpretation, it is useful to analyse what happens when the linguistic input is known to contain local ambiguities. That way, we have a genuine motivation to expect revisions and can study how states change at key positions.

Let $v_{jk}^i$ be the value assigned for feature $k$ of token $j$ at time step $i$ in a sequence with $n$ tokens. We can compare the differences between $v_{jk}^i$ and four relevant positions with clear interpretations:

\begin{itemize}
    \item the value $v_{jk}^j$ at the main diagonal, \textit{i.e.} the initial interpretation of token $j$ when it was first observed without any right context

    \item the value $v_{jk}^{(i-1)}$ at the previous time step, \textit{i.e.} the change in the interpretation of token $j$ caused by the most recently added token $w_i$

    \item the value $v_{jk}^n$ at the last time step, \textit{i.e.} the final (gold standard) interpretation of token $j$ when it is observed with the whole context

    \item the value $v_{jk}^{(i+d)}$, where $i+d$ is the position when the disambiguating token occurs
 
\end{itemize}

\vspace{0.2cm}

\begin{figure}[t]
    \centering
    \includegraphics[trim={0cm 2.2cm 16cm 0cm},clip,width=\columnwidth,page=1]{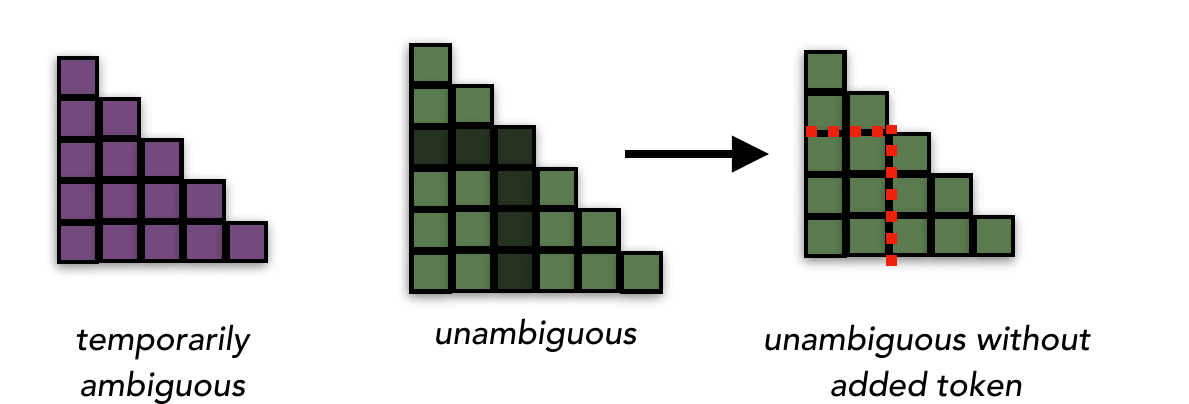}
    \caption{We realign tokens, after computing the states, by removing the states of additional token(s) from the triangular structure. That way we can directly compare the states of a locally ambiguous sentence with its unambiguous counterpart.}
    \label{fig:gp-effect}
\end{figure}

A classic approach in studies using surprisal or reading time as a measure of processing difficulty is to consider the difference of the measure in the disambiguating region of an unambiguous sentence and a locally ambiguous sentence as the garden path effect \citep{van2021single,arehalli-etal-2022-syntactic,huang2023surprisal}. Drawing a parallel with that approach, we compute triangular structures for the stimulus and its baseline. After removing the extra tokens, we have two structures with the same dimensions, as shown in Figure \ref{fig:gp-effect}. To give an example, such a pairing would be \textit{``The professor noticed the grant gained more attention''} and \textit{``The professor noticed (that) the grant gained more attention''}, with the token in parentheses  removed \textbf{after computing the states} to make the structures directly comparable.

\vspace{0.3cm}
\section{Analysis}
\label{sec:analysis}
Overall, we are interested in how future tokens affect states of past tokens and how such changes surface as output revisions. Our main hypothesis is that RI models are led down the garden path when they first encounter a local ambiguity without further right context but, as the disambiguating region is integrated, states are updated to absorb the new interpretation. We provide insights into updates that may be allowing them to recover. 

\paragraph{Scope} We apply our method to two RI scenarios. Firstly, we look at the construction of meaning representations in bidirectional LMs (§\ref{sec:analysis-meaning}). We assume that the representations of a token encode its meaning in the available context, tracking how it evolves step by step and across all the layers. If there is a shift in the meaning of a token, we expect to observe variation over a control reference in the corresponding states. The second scenario is dependency parsing as sequence labelling with arcs and relations (§\ref{sec:analysis-parsing}) \citep{spoustova2010dependency, strzyz-etal-2019-viable}. In this case, we have the output labels as an external signal of the concrete decisions made by the model. We investigate whether changes in divergence of the attention scores and of the distribution over labels or arcs align with output edits and with the resolution of the ambiguity.

\begin{figure}[h]
    \centering
    \includegraphics[trim={0cm 19cm 8cm 0cm},clip,width=\columnwidth,page=1]{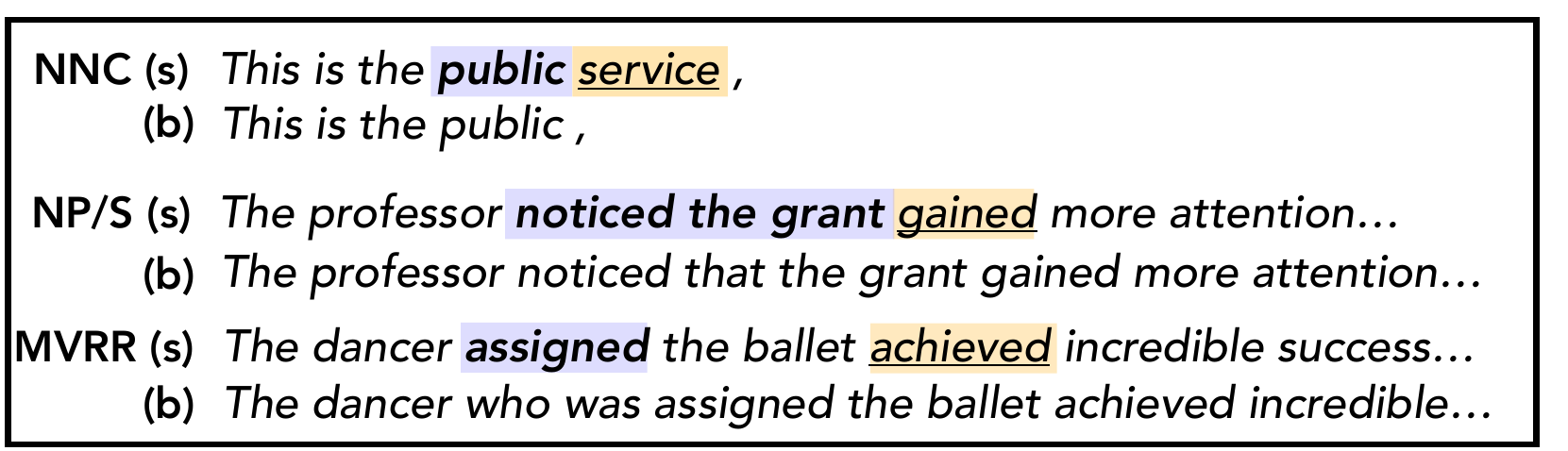}
    \caption{The three types of stimuli (s) and their corresponding reference baselines (b) used in our analyses. The words in lilac/bold are locally ambiguous until the underlined/yellow token is observed.}
    \label{fig:temp-amb}
\end{figure}

\paragraph{Material}  We study three kinds of local ambiguities in English: 24 instances of Direct Object/Sentential Complement (NP/S) and of Main Verb/Reduced Relative ambiguity (MVRR) garden paths from \citet{huang2023surprisal} and 281 instances of noun-noun compounds (NNC) from \citet{garcia-etal-2021-probing} with a fixed context. Examples of each type are shown in Figure \ref{fig:temp-amb}. NNC has a very localised need for revision, where the immediate next token changes the interpretation of the preceding noun. In NP/S and MVRR, the disambiguating token appears after a NP, with a more broad syntactic and semantic shift of all the prefix. Please see the original publications for detailed motivations.

\vspace{0.2cm}

\subsection{General Effects of Right Context}

We start by showing, in Table \ref{tab:variation}, that even in the baseline stimuli (where no major revisions are expected), a token's state keeps being updated as more right tokens are integrated. For meaning, we compute the cosine distance between a state and its previous version at the last layer of BERT, to isolate how the latest token $w_{t+i}$ affects $s_t^{t+i-1}$. For dependency parsing, we measure the variation of entropy of the arc distribution also with respect to its previous state for the biaffine parser \citep{dozat2017deep} with RoBERTa. On average, there is a considerable effect when the immediately next token is added, which decreases gradually as more right context is observed in almost all cases. 

\vspace{0.2cm}
\begin{table}[ht!]
\centering
\small
{\setlength{\tabcolsep}{3pt}
\begin{tabular}{l c c c c c c c }
    \toprule
         & $t\!+\!1$ & $t\!+\!2$ & $t\!+\!3$ & $t\!+\!4$ & $t\!+\!5$ & $t\!+\!6$ & $t\!+\!7$ \\
    \midrule
    \textbf{Meaning} & & & & & & \\
        \hspace{0.2cm} MVRR & 0.38 & 0.09 & 0.05 & 0.04 & 0.04 & 0.03 & 0.03 \\
        \hspace{0.2cm} NPS & 0.39 & 0.10 & 0.07 & 0.05 & 0.03 & 0.03 & 0.02 \\
        \hspace{0.2cm} NNC & 0.34 & 0.12 & 0.15 & 0.13 & - & - & - \\
    \midrule
    \textbf{DP} & & & & & & \\
    \hspace{0.2cm} MVRR & 0.11 & 0.02 & 0.01 & 0.02 & 0.01 & 0.00 & 0.00 \\
    \hspace{0.2cm} NPS & 0.19 & 0.06 & 0.05 & 0.01 & 0.00 & 0.00 & 0.00 \\
    \hspace{0.2cm} NNC & 0.14 & 0.21 & 0.29 & 0.08 & - & - & - \\
    \bottomrule
\end{tabular}
}
\caption{Average effect of token $w_{t+i}$ (each column) on previous states $s_t$ over all baseline stimuli (\textit{i.e.}~here, the effect is measured without the influence of the garden path structure). 
}
\label{tab:variation}

\end{table}

\subsection{Incremental Construction of Meaning}
\label{sec:analysis-meaning}
For this part, we extract the hidden states for all layers of pretrained bidirectional transformer LMs. We show results for BERT \citep{devlin-etal-2019-bert} here and RoBERTa \citep{liu2019roberta} in Appendix \ref{sec:appendix_meaning}. We also study static embeddings in causal models, namely GPT-2 \citep{radford2019language} here and OPT \citep{zhang2022opt} in Appendix \ref{sec:appendix_meaning}. We measure how much states are updated, by computing their cosine distance to a reference time step.

\begin{figure}[t]
    \centering
    \includegraphics[trim={0cm 0cm 0cm 0cm},clip,width=\columnwidth,page=1]{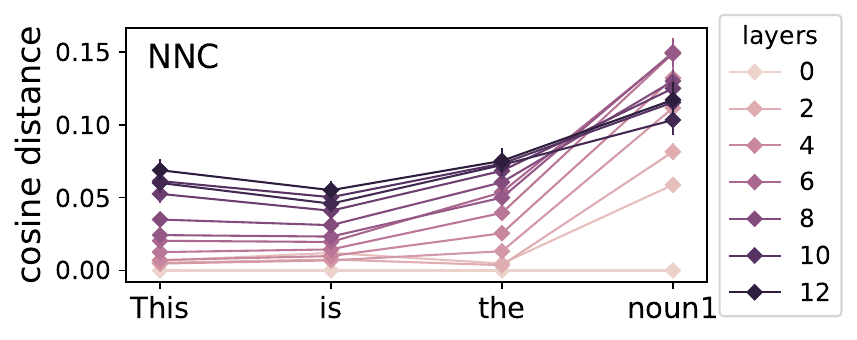}
    \caption{BERT's mean effect of the second noun on the tokens in the prefix. Absolute difference over baseline.}
    \label{fig:nnc-bert}
\end{figure}

\paragraph{NNC} In Figure \ref{fig:nnc-bert}, we zoom in at what happens to the prefix when the second noun is added (the fifth time step) by computing the cosine distance between states $s_i^4$ and $s_i^5$, for $i=1, \dots, 4$. We subtract from it a baseline case where a comma is observed instead. This controls for keeping the meaning of the first noun as a NP head versus it becoming a modifier of the second noun. The results  show that the second noun affects the meaning of all previous tokens, more than the baseline, and the effect is even larger for the first noun, in all layers but especially in middle ones. In the last layer, the mean cosine distance between the initial state of the first noun and its updated version when the second noun is observed is almost 0.41, considerably above the corresponding 0.34 in Table \ref{tab:variation}. 

\begin{figure}[h]
    \centering
    \frame{\includegraphics[trim={0cm 0cm 0cm 0cm},clip,width=0.49\columnwidth,page=1]{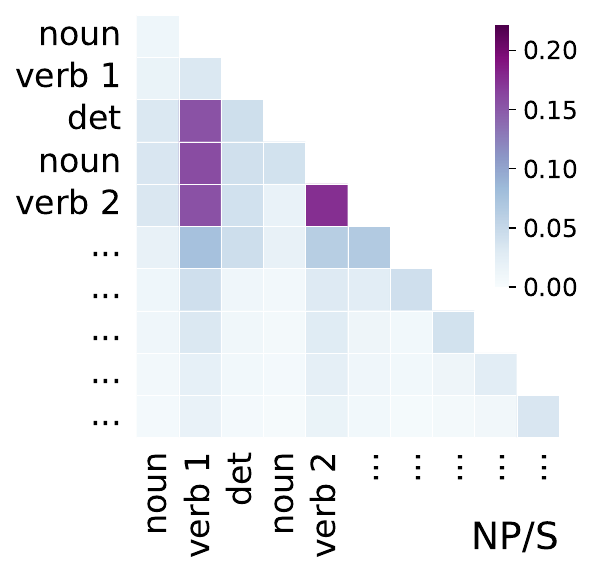}}
    \frame{\includegraphics[trim={0cm 0cm 0cm 0cm},clip,width=0.49\columnwidth,page=1]{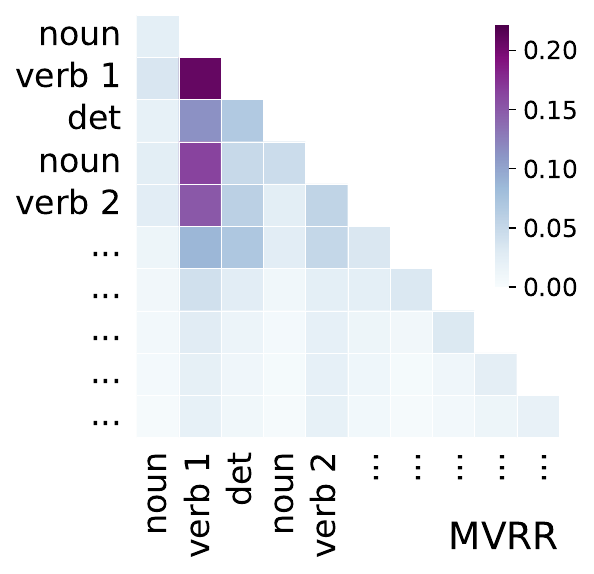}}
    
    \caption{Average absolute difference between stimulus and baseline in distance of each representation to its final state at layer 9 of BERT.}
    \label{fig:classic-gp-bert}
\end{figure}

\paragraph{NP/S} In these stimuli, humans can first interpret the NP as direct object, but only until disambiguating region is observed, when it becomes a sentential complement. The final meaning of the stimulus is the same as the baseline's. Thus here we measure how distant each $s_i^j$ is from its target interpretation $s_i^N$ at the final time step $N$. Then we compute the absolute difference between the distances for the stimulus and the baseline. In Figure \ref{fig:classic-gp-bert} (left), we show the effect around the disambiguating token for the layer where it is most prominent. The representation of the first verb diverges from its final meaning as the NP is processed. After the disambiguating region is integrated, this difference almost disappears. This suggests that the model first builds an initial interpretation for the stimulus but, one token after the second verb, it turns to its final meaning, as in the unambiguous case. BERT seems to load the semantics of the argument on the first verb, so the noun does not encode so much what its role is in the two variations of the sentence. We also observe this in almost all layers, especially the middle to upper ones (see Appendix \ref{sec:appendix_meaning}).

\vspace{0.2cm}

\paragraph{MVRR} For stimuli where the first verb becomes a reduced relative, we perform the same type of analysis as in NP/S and observe similar results. In Figure \ref{fig:classic-gp-bert} (right), a similar pattern occurs, but the divergence starts earlier, at the initial step where the first verb is observed. Again, this is evidence that the initial representation of the verb did not fully encode what would be the final reduced relative form and is revised in the face of upcoming tokens.

The mean variation on the first verb once the second disambiguating token is observed is 0.15 and 0.14 for NP/S and MVRR, well above the average variation at $t+4$ in Table \ref{tab:variation}. The conclusion so far is that future tokens do affect the meaning representation of previous tokens and more considerably so when the model has a linguistic motivation to revise its states. 

\begin{figure}[h]
    \centering
    \frame{\includegraphics[trim={0cm 0cm 0cm 0cm},clip,width=0.49\columnwidth,page=1]{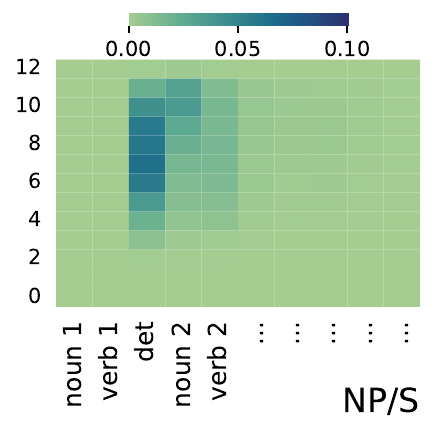}}
    \frame{\includegraphics[trim={0cm 0cm 0cm 0cm},clip,width=0.49\columnwidth,page=1]{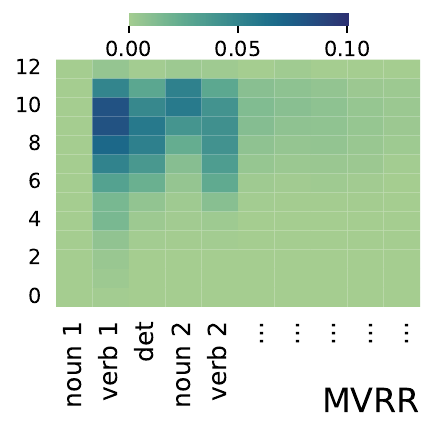}}
    
    \caption{Distance between the causal embeddings of the stimulus and the baseline, after taking the absolute difference of the expected variation by a counterpart pair with an unambiguous verb, for all layers of GPT-2.}
    \label{fig:causal-gpt2}
\end{figure}

\paragraph{Causal Models} We now look at what happens with static embeddings by directly computing the distance from the states in the ambiguous stimulus to the corresponding states in the baseline, which disambiguates the meaning in advance. We subtract from that the distance of a similar pair of sentences with an unambiguous first verb (\textit{given} for MVRR and \textit{said} for NP/S, see Appendix \ref{sec:appendix_meaning} for the detailed formulation) to account for the effect of one sentence having more tokens than the other. The remaining absolute difference is shown in Figure \ref{fig:causal-gpt2}. The intermediate layers encode a difference, which should be due to non-commitment in the ambiguous stimulus to what will turn out to be the ``true'' analysis. That can be interpreted as how the model would revise, if it could. In OPT (Appendix \ref{sec:appendix_meaning}), the difference remains and can affect its predictions. For GPT-2's last layer, however, the distance is close to 0 in both pairs, so that the actual states used for downstream decisions are practically the same in the stimulus and the baseline, despite their potentially different unfolding meanings. 

\subsection{Incremental Dependency Parsing}
\label{sec:analysis-parsing}
Aside from being a fundamental aspect of human language processing, incremental parsing is also useful in applications such as simultaneous translation \citep{ryu-etal-2006-simultaneous} and disfluency detection \citep{honnibal-johnson-2014-joint}. We investigate two dependency parsers in a RI setting: 1) the biaffine parser \citep{dozat2017deep} trained on top of RoBERTa and fine-tuned on PTB \citep{marcus-etal-1993-building} and 2) the DiaParser \citep{attardi-etal-2021-biaffine}, which uses ELECTRA \citep{clark2020electra} and is fine-tuned on PTB and EWT \citep{silveira-etal-2014-gold}. Here we only show findings for the biaffine parser as we find similar results to occur in DiaParser. See Appendix \ref{sec:appendix_parsing} for a full comparison.

Our focus is on the self-attention mechanism of the parsers, from which the dependency arcs and labels are directly derived. Both parsers handle parsing and labelling decisions sequentially, selecting the labels for each arc only after ensuring the well-formedness of the tree via the MST algorithm \citep{chu-liu, edmonds1967optimum}. As such, an independent analysis of the dependency labels is not possible, as they depend on the predicted arcs, which in turn may also change from time to time as more tokens are observed. Hence we analyse both dependency arcs and labels in a joint fashion. We measure how attention distributions evolve across time steps, by computing the Jensen-Shannon divergence (JSD) with respect to the last, previous, or first (the main diagonal) time step as a reference. We also consider how the arc may change due to future tokens. See Appendix \ref{sec:appendix_parsing} for full details.

\begin{figure}[h]
    \centering
    \frame{\includegraphics[trim={0cm 0cm 0cm 0cm},clip,width=0.49\columnwidth,page=1]{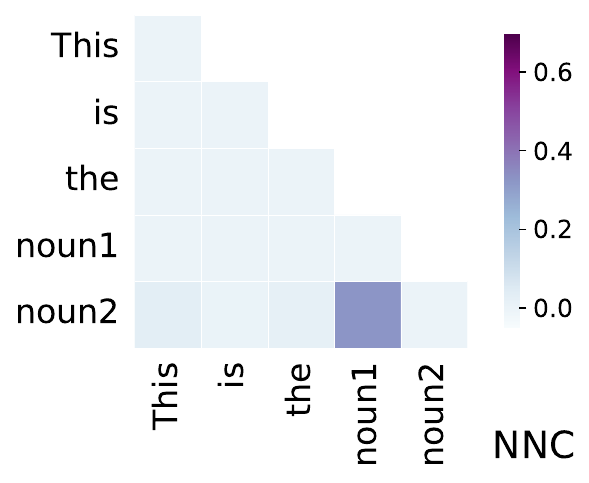}}
    \frame{\includegraphics[trim={0cm 0cm 0cm 0cm},clip,width=0.49\columnwidth,page=1]{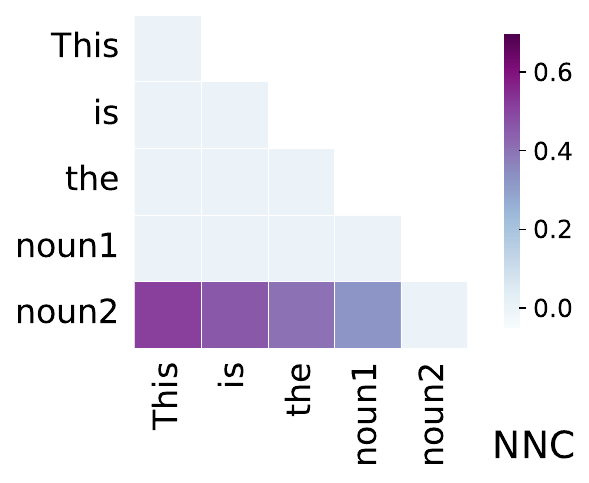}}
    
    \caption{Average absolute difference of JSD between stimulus and baseline on NNC. Left: first step reference. Right: previous step reference.}
    \label{fig:nnc-dep}
\end{figure}

\paragraph{NNC} We isolate the effect of the local ambiguity by computing the absolute difference of JSD between the stimulus and the baseline in a similar manner to (§\ref{sec:analysis-meaning}). Using the first time step as a reference, we glean how far the label distribution of the first noun shifts from its origin when it acts as a modifier for the second noun as opposed to encountering a comma, where it retains its interpretation as the NP head. We also compute the difference with respect to the previous step to investigate how the distribution shift causes the dependency structure to change through time. Both are shown in Figure \ref{fig:nnc-dep}. On the left, we see that the second noun alters the original distribution of the first noun more than the baseline, as the head-dependent relation of the first noun changes. To the right, we find that the second noun affects not only the first noun, but also the distributions of all previous tokens as it replaces the first noun as the argument of \textit{is}. This is not present on the left figure. 

\vspace{0.2cm}

\begin{figure}[ht]
    \centering
    \frame{\includegraphics[trim={0cm 0cm 0cm 0cm},clip,width=0.49\columnwidth,page=1]{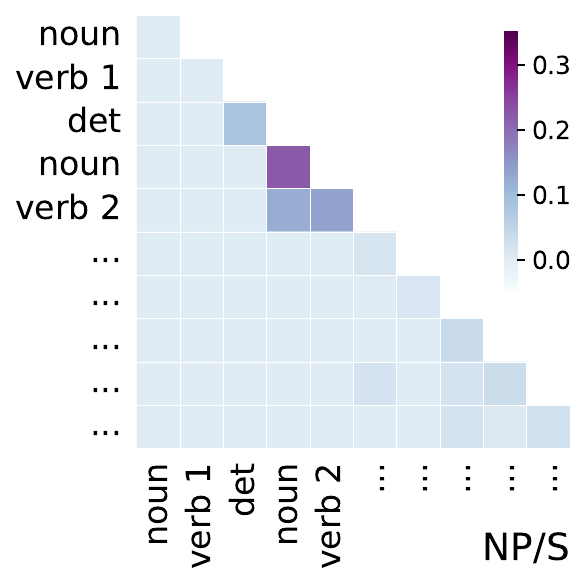}}
    \frame{\includegraphics[trim={0cm 0cm 0cm 0cm},clip,width=0.49\columnwidth,page=1]{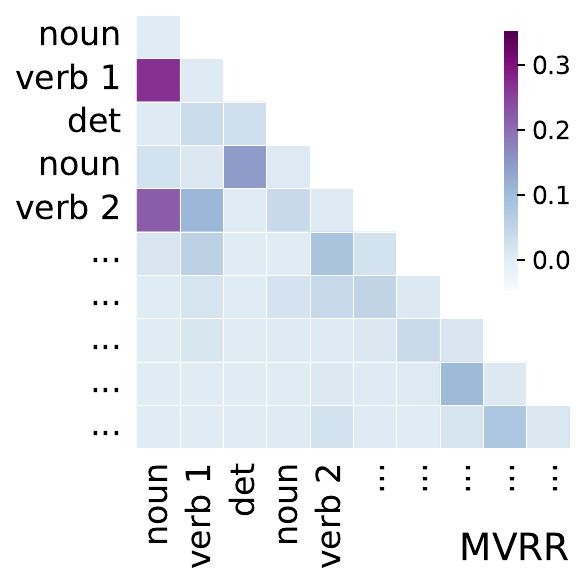}}
    
    \caption{Average absolute difference of JSD between stimulus and baseline on NP/S (left, final step reference) \& MVRR (right, previous step reference).}
    \label{fig:nps-mvrr-dep}
\end{figure}

\paragraph{NP/S} As the stimulus in the NP/S ambiguity has the same final interpretation as their unambiguous counterpart, we take the final step as a reference and compute the absolute difference of JSD between them, factoring out the complementiser \textit{that}. We find that the label distribution of the sentential complement diverges at the beginning between the stimuli and the baseline when compared to the final distribution (Figure \ref{fig:nps-mvrr-dep}, left). However, there is almost no difference between them after entering the disambiguating region, similar to (§\ref{sec:analysis-meaning}). 

\vspace{0.2cm}

\paragraph{MVRR} For MVRR, we are interested in how the dynamics of the parse evolve throughout the time axis. We subtract the JSD of the stimulus calculated wrt.\ the previous step from the baseline chart. In Figure \ref{fig:nps-mvrr-dep} (right), we notice peaks in the difference for the first noun when the first and second verbs are encountered. The first one occurs as the first verb can initially be parsed either as the main verb or the verb in the reduced relative clause, making the first noun be interpreted differently. However, this is reconciled in the second peak when the disambiguating region is incorporated to the stimulus' interpretation. We observe an effect similar to NP/S afterwards, as the ambiguity is resolved and the JSD difference vanishes.

\paragraph{Alignment with Edits}
We examine whether changes in label distributions due to future tokens also coincide with changes in arcs or labels. For instance, this happens in NNC when the first noun is revised from the argument of the verb to be the modifier of the second noun. To do this, we use the JSD chart with the previous step as a reference and assume that changes or edits happens if its value is bigger than a threshold $\tau=0.45\ln(2)$, where $\ln(2)$ is the divergence upper bound. Then we compute the Matthews correlation coefficient (MCC; \citealp{MATTHEWS1975442}) between the predictions and the edits (Table \ref{tab:alignment}). In general, we find that distribution shifts are positively correlated to edits, with a stronger magnitude for dependency arcs compared to labels. This suggests that shifts in distributions encode information that surface as revisions. 

\vspace{0.3cm}
\begin{table}[ht!]
\centering
\small
{\setlength{\tabcolsep}{3pt}
\begin{tabular}{l c c c c c c}
    \toprule
         & \multicolumn{2}{c}{NNC} & \multicolumn{2}{c}{NP/S} & \multicolumn{2}{c}{MVRR} \\
    \midrule
    \textbf{Arc} & S & B & S & B & S & B \\
        \cmidrule(lr){2-3} \cmidrule(lr){4-5} \cmidrule(lr){6-7}
        \hspace{0.2cm} Biaffine & 0.99 & 0.96 & 0.95 & 0.92 & 0.86 & 0.93  \\
        \hspace{0.2cm} DiaParser (EWT) & 0.97 & 1.00 & 0.95 & 0.92 & 0.80 & 0.94 \\
        \hspace{0.2cm} DiaParser (PTB) & 0.81 & 0.78 & 0.91 & 0.92 & 0.79 & 0.96 \\
    \midrule
    \textbf{Label} & S & B & S & B & S & B \\
        \cmidrule(lr){2-3} \cmidrule(lr){4-5} \cmidrule(lr){6-7}
        \hspace{0.2cm} Biaffine & 0.55 & 0.77 & 0.86 & 0.77 & 0.71 & 0.84  \\
        \hspace{0.2cm} DiaParser (EWT) & 0.37 & 0.53 & 0.78 & 0.70 & 0.61 & 0.75 \\
        \hspace{0.2cm} DiaParser (PTB) & 0.23 & 0.52 & 0.78 & 0.72 & 0.64 & 0.86 \\
    \bottomrule
\end{tabular}
}
\caption{MCC between distribution shifts and edits for dependency arcs and labels. S: stimulus \& B: baseline.
}
\label{tab:alignment}
\end{table}

\section{Discussion \& Conclusion}
\label{sec:discussion}
Our method has shown that a RI interface applied to bidirectional encoders yields models that build sequences of states with a rich dynamics of updates of past representations. This grants these models with \textit{revisability}, a property that is desirable in incremental systems \citep{schlangen2011general} but  not present in unidirectional models. While causal models must create a representation based only on left context and stick to their first commitment, RI bidirectional models can profit from incorporating right context and revisit its previous decisions. Our analysis has empirically revealed that the RI models we studied seem to run into the downsides of parsing, \textit{i.e.}~they are led down the garden path, but their initial internal representations are updated once the disambiguating region is processed, more than in the baselines. In other words, it seems that they make early commitments but then revise accordingly, as in two-stage approaches to language processing. 

In the analysis of contextualised embeddings, the effect is less pronounced on lower layers, but more prominent from middle to upper layers. This relates to the findings of \citet{tenney-etal-2019-bert} who suggest that upper layers (that are more semantic) can be used to disambiguate decisions in lower layers (that are more syntactic). This is also in line with works showing that middle to upper layers are most informative for some tasks \citep{liu-etal-2019-linguistic}.

Treating the triangular structures as internal states of a transition system constructed from graph-based dependency parsers allows us to uncover how relations between tokens change as an effect of future tokens. By using the divergence computed with various reference time steps as a measure, we show that attention distributions may evolve differently throughout the course of processing and how they can become similar again through the disambiguation process. The shift in distributions also indicates edits in dependency arcs and labels, and more importantly, enlightens us more about why RI models revise when they do.

While decoder-only LLMs have grown in popularity the past few years, we still see room for improvement, particularly regarding how their static representations can be updated (like RI models) as more tokens are  processed. This can benefit tasks that require non-monotonic reasoning \cite{KRAUS1990167}, regardless of whether it involves NLG or NLU. Natural language is also inherently ambiguous, although its spectrum of ambiguity depends on its use \citep{schlangen-2023-general} which necessitates this processing mode. Moreover, modern LLMs still struggle with this aspect of language understanding \citep{liu-etal-2023-afraid}.

\label{sec:conclusion}
Applying a RI interface may result in a non-monotonic output. However, there are certain cases where monotonicity is preferred. For instance, to reduce flicker in simultaneous machine translation models \citep{arivazhagan-etal-2020-translation, 9054585} or when a stable and immediate output is always needed for further processing in an incremental pipeline. In both cases, the lookahead strategy \citep{beuck-etal-2011-decision} can be used for disambiguation by considering tokens within the lookahead window. Alternative methods like beam search can also be employed, depending on how various properties such as monotonicity, timeliness, and decisiveness \citep{kohn-2018-incremental, beuck-etal-2011-decision} are prioritised in the use case.

One of Transformers' properties that does not align with human language processing, is how words are processed in parallel, even in autoregressive settings. Ideally, models of human language should also face similar input and linguistic challenges as humans do \citep{BLANK2023987}. We show that when facing such challenge in the form of garden-path sentences, Transformers with a RI interface still exhibit the ability to disambiguate using forward-reanalysis \citep{frazier-rayner-1982}. 

Future research can explore how the restart activity can serve as an indicator of updates that should not be immediately integrated into the output, or model a controller that is able to detect variations that lead to revisions and decide whether to delay outputs until a level of certainty has been reached. Additionally, it would be interesting to see how a RI interface can be used to investigate the capability of NLP models to deal with linguistic ambiguities in other languages than English, \emph{e.g.} the interpretation of relative clauses in Dutch \citep{wijnholds-moortgat-2023-structural}.

\section*{Limitations}
\label{sec:limitations}
Feeding prefixes to models pretrained on full sentences can lead to out-of-distribution issues. The results in \citet{madureira-schlangen-2020-incremental} showed that bidirectional models can exhibit a high relative correctness, which means that models often output the same labels for prefixes as they do for the full sentence, although BERT suffers more than others. This is exactly why it is important to understand how much they rely on the right context and study how and when revisions happen in this setting, with stimuli that is expected to induce a revision for human language processing. 

We have only considered short-range temporary ambiguities, \textit{i.e.}~those that are resolved up to 3 or 4 tokens into the future. It would be interesting to study whether meaningful revisions also occur when the distance between the temporarily ambiguous token and the disambiguating region is longer.

We observed that the standard deviation of the means we report can be large at some cases. That means that, for some sentences, the effect is less present and in others where it is more extreme. This may relate to what is the most likely early commitment given the lexical information of the ambiguous tokens. Further investigation could try to consider that. Larger samples would also be necessary to have better mean and std.~estimators. Related to that, we did not take into account the frequency of the vocabulary choices in these datasets. 

The garden path effects we found may not be so salient for all types of local ambiguities. Our initial analysis on the verb-noun ambiguity using the stimuli by \citet{aina-linzen-2021-language} (like the example in Figure \ref{fig:example}) led to inconclusive results. For such cases, maybe other metrics are required, especially because it is harder to isolate the effect of the ambiguity from the effect that the immediate next token has on its neighbour. Further investigation can also be done with ungrammatical sentences and other types of garden paths in the SAP benchmark.

As we discuss in the Appendix, some stimuli were excluded from the analysis of meaning due to rare tokens that require subtokenisation and cause misalignment in the incremental structures.

For the analysis of causal embeddings, we used different architectures. Ideally, we should also compare a fixed architecture, trained with all the same parameters and data, but once with causal masking and once with bidirectional access. That way we could directly compare how the RI bidirectional and causal embeddings differ. Besides, cosine  has some known drawbacks as a measure of similarity of embeddings, because it is sensitive to training data frequency and can underestimate similarity in comparison to human judgements \citep{zhou-etal-2022-problems}. We explored other metrics, but opted to report cosine similarity due to its intuitive interpretation and bounds. Future studies could further explore other metrics. See \citet{jurayj-etal-2022-garden} for a comparison.

We do not perform layer-level analysis for dependency parsing as both parsers use scalar mix of hidden representations from pre-trained LMs. While we only use graph-based dependency parsers, it would be interesting to apply our methods to transition-based parsers, as they works from left to right and are traditionally associated with the notion of incrementality \citep{nivre-2008-algorithms}. In our preliminary analysis, we also compare Shannon's entropy on attention distributions for dependency arcs. However, the results are inconclusive.

\section*{Acknowledgements}
We thank the anonymous reviewers for their valuable comments and suggestions. We also thank Sophia Rauh for the initial analysis on the incremental labels and for the annotation of token positions in the stimuli. This work is partially funded by the Deutsche Forschungsgemeinschaft (DFG, German Research Foundation) – 423217434 ("RECOLAGE") grant.

\bibliography{anthology,custom}

\begin{thebibliography}{84}
\expandafter\ifx\csname natexlab\endcsname\relax\def\natexlab#1{#1}\fi

\bibitem[{Abnar and Zuidema(2020)}]{abnar-zuidema-2020-quantifying}
Samira Abnar and Willem Zuidema. 2020.
\newblock \href {https://doi.org/10.18653/v1/2020.acl-main.385} {Quantifying attention flow in transformers}.
\newblock In \emph{Proceedings of the 58th Annual Meeting of the Association for Computational Linguistics}, pages 4190--4197, Online. Association for Computational Linguistics.

\bibitem[{Aina and Linzen(2021)}]{aina-linzen-2021-language}
Laura Aina and Tal Linzen. 2021.
\newblock \href {https://doi.org/10.18653/v1/2021.blackboxnlp-1.4} {The language model understood the prompt was ambiguous: Probing syntactic uncertainty through generation}.
\newblock In \emph{Proceedings of the Fourth BlackboxNLP Workshop on Analyzing and Interpreting Neural Networks for NLP}, pages 42--57, Punta Cana, Dominican Republic. Association for Computational Linguistics.

\bibitem[{Altmann and Steedman(1988)}]{altmann1988interaction}
Gerry Altmann and Mark Steedman. 1988.
\newblock \href {https://doi.org/10.1016/0010-0277(88)90020-0} {Interaction with context during human sentence processing}.
\newblock \emph{Cognition}, 30(3):191--238.

\bibitem[{Arehalli et~al.(2022)Arehalli, Dillon, and Linzen}]{arehalli-etal-2022-syntactic}
Suhas Arehalli, Brian Dillon, and Tal Linzen. 2022.
\newblock \href {https://doi.org/10.18653/v1/2022.conll-1.20} {Syntactic surprisal from neural models predicts, but underestimates, human processing difficulty from syntactic ambiguities}.
\newblock In \emph{Proceedings of the 26th Conference on Computational Natural Language Learning (CoNLL)}, pages 301--313, Abu Dhabi, United Arab Emirates (Hybrid). Association for Computational Linguistics.

\bibitem[{Arivazhagan et~al.(2020{\natexlab{a}})Arivazhagan, Cherry, Macherey, and Foster}]{arivazhagan-etal-2020-translation}
Naveen Arivazhagan, Colin Cherry, Wolfgang Macherey, and George Foster. 2020{\natexlab{a}}.
\newblock \href {https://doi.org/10.18653/v1/2020.iwslt-1.27} {Re-translation versus streaming for simultaneous translation}.
\newblock In \emph{Proceedings of the 17th International Conference on Spoken Language Translation}, pages 220--227, Online. Association for Computational Linguistics.

\bibitem[{Arivazhagan et~al.(2020{\natexlab{b}})Arivazhagan, Cherry, Te, Macherey, Baljekar, and Foster}]{9054585}
Naveen Arivazhagan, Colin Cherry, I~Te, Wolfgang Macherey, Pallavi Baljekar, and George Foster. 2020{\natexlab{b}}.
\newblock \href {https://doi.org/10.1109/ICASSP40776.2020.9054585} {Re-translation strategies for long form, simultaneous, spoken language translation}.
\newblock In \emph{ICASSP 2020 - 2020 IEEE International Conference on Acoustics, Speech and Signal Processing (ICASSP)}, pages 7919--7923.

\bibitem[{Artetxe et~al.(2022)Artetxe, Du, Goyal, Zettlemoyer, and Stoyanov}]{artetxe-etal-2022-role}
Mikel Artetxe, Jingfei Du, Naman Goyal, Luke Zettlemoyer, and Veselin Stoyanov. 2022.
\newblock \href {https://doi.org/10.18653/v1/2022.findings-emnlp.293} {On the role of bidirectionality in language model pre-training}.
\newblock In \emph{Findings of the Association for Computational Linguistics: EMNLP 2022}, pages 3973--3985, Abu Dhabi, United Arab Emirates. Association for Computational Linguistics.

\bibitem[{Attardi et~al.(2021)Attardi, Sartiano, and Simi}]{attardi-etal-2021-biaffine}
Giuseppe Attardi, Daniele Sartiano, and Maria Simi. 2021.
\newblock \href {https://doi.org/10.18653/v1/2021.iwpt-1.19} {Biaffine dependency and semantic graph parsing for {E}nhanced{U}niversal dependencies}.
\newblock In \emph{Proceedings of the 17th International Conference on Parsing Technologies and the IWPT 2021 Shared Task on Parsing into Enhanced Universal Dependencies (IWPT 2021)}, pages 184--188, Online. Association for Computational Linguistics.

\bibitem[{Belinkov and Glass(2019)}]{belinkov-glass-2019-analysis}
Yonatan Belinkov and James Glass. 2019.
\newblock \href {https://doi.org/10.1162/tacl_a_00254} {Analysis methods in neural language processing: A survey}.
\newblock \emph{Transactions of the Association for Computational Linguistics}, 7:49--72.

\bibitem[{Belrose et~al.(2023)Belrose, Furman, Smith, Halawi, Ostrovsky, McKinney, Biderman, and Steinhardt}]{belrose2023eliciting}
Nora Belrose, Zach Furman, Logan Smith, Danny Halawi, Igor Ostrovsky, Lev McKinney, Stella Biderman, and Jacob Steinhardt. 2023.
\newblock \href {https://doi.org/10.48550/arXiv.2303.08112} {Eliciting latent predictions from transformers with the tuned lens}.
\newblock \emph{arXiv preprint arXiv:2303.08112}.

\bibitem[{Beuck et~al.(2011)Beuck, K{\"o}hn, and Menzel}]{beuck-etal-2011-decision}
Niels Beuck, Arne K{\"o}hn, and Wolfgang Menzel. 2011.
\newblock \href {https://aclanthology.org/W11-4605} {Decision strategies for incremental {POS} tagging}.
\newblock In \emph{Proceedings of the 18th Nordic Conference of Computational Linguistics ({NODALIDA} 2011)}, pages 26--33, Riga, Latvia. Northern European Association for Language Technology (NEALT).

\bibitem[{Bever(1970)}]{bever1970}
Thomas Bever. 1970.
\newblock \href {https://doi.org/10.1093/acprof:oso/9780199677139.003.0001} {\emph{The Cognitive Basis for Linguistic Structures}}, pages 279--352.

\bibitem[{Blank(2023)}]{BLANK2023987}
Idan~A. Blank. 2023.
\newblock \href {https://doi.org/https://doi.org/10.1016/j.tics.2023.08.006} {What are large language models supposed to model?}
\newblock \emph{Trends in Cognitive Sciences}, 27(11):987--989.

\bibitem[{Chen et~al.(2024)Chen, Shwartz-Ziv, Cho, Leavitt, and Saphra}]{chen2023sudden}
Angelica Chen, Ravid Shwartz-Ziv, Kyunghyun Cho, Matthew~L Leavitt, and Naomi Saphra. 2024.
\newblock \href {https://openreview.net/forum?id=MO5PiKHELW} {Sudden drops in the loss: Syntax acquisition, phase transitions, and simplicity bias in {MLM}s}.
\newblock In \emph{The Twelfth International Conference on Learning Representations}.

\bibitem[{Chen et~al.(2022)Chen, Zayats, Walker, and Padfield}]{chen-etal-2022-teaching}
Angelica Chen, Vicky Zayats, Daniel Walker, and Dirk Padfield. 2022.
\newblock \href {https://doi.org/10.18653/v1/2022.naacl-main.60} {Teaching {BERT} to wait: Balancing accuracy and latency for streaming disfluency detection}.
\newblock In \emph{Proceedings of the 2022 Conference of the North American Chapter of the Association for Computational Linguistics: Human Language Technologies}, pages 827--838, Seattle, United States. Association for Computational Linguistics.

\bibitem[{Chu and Liu(1965)}]{chu-liu}
Yeong-Jin Chu and Tseng-Hong Liu. 1965.
\newblock \href {https://cir.nii.ac.jp/crid/1571135650708759040} {On the shortest arborescence of a directed graph}.
\newblock \emph{Science Sinica}, 14:1396--1400.

\bibitem[{Clark et~al.(2019)Clark, Khandelwal, Levy, and Manning}]{clark-etal-2019-bert}
Kevin Clark, Urvashi Khandelwal, Omer Levy, and Christopher~D. Manning. 2019.
\newblock \href {https://doi.org/10.18653/v1/W19-4828} {What does {BERT} look at? an analysis of {BERT}{'}s attention}.
\newblock In \emph{Proceedings of the 2019 ACL Workshop BlackboxNLP: Analyzing and Interpreting Neural Networks for NLP}, pages 276--286, Florence, Italy. Association for Computational Linguistics.

\bibitem[{Clark et~al.(2020)Clark, Luong, Le, and Manning}]{clark2020electra}
Kevin Clark, Minh-Thang Luong, Quoc~V. Le, and Christopher~D. Manning. 2020.
\newblock \href {https://openreview.net/forum?id=r1xMH1BtvB} {Electra: Pre-training text encoders as discriminators rather than generators}.
\newblock In \emph{International Conference on Learning Representations}.

\bibitem[{Devlin et~al.(2019)Devlin, Chang, Lee, and Toutanova}]{devlin-etal-2019-bert}
Jacob Devlin, Ming-Wei Chang, Kenton Lee, and Kristina Toutanova. 2019.
\newblock \href {https://doi.org/10.18653/v1/N19-1423} {{BERT}: Pre-training of deep bidirectional transformers for language understanding}.
\newblock In \emph{Proceedings of the 2019 Conference of the North {A}merican Chapter of the Association for Computational Linguistics: Human Language Technologies, Volume 1 (Long and Short Papers)}, pages 4171--4186, Minneapolis, Minnesota. Association for Computational Linguistics.

\bibitem[{Dozat and Manning(2017)}]{dozat2017deep}
Timothy Dozat and Christopher~D. Manning. 2017.
\newblock \href {https://openreview.net/forum?id=Hk95PK9le} {Deep biaffine attention for neural dependency parsing}.
\newblock In \emph{International Conference on Learning Representations}.

\bibitem[{Duki{\'c} and {\v{S}}najder(2024)}]{dukic2024not}
David Duki{\'c} and Jan {\v{S}}najder. 2024.
\newblock \href {https://doi.org/10.48550/arXiv.2401.14556} {Looking right is sometimes right: Investigating the capabilities of decoder-only llms for sequence labeling}.
\newblock \emph{arXiv preprint arXiv:2401.14556}.

\bibitem[{Edmonds(1967)}]{edmonds1967optimum}
Jack Edmonds. 1967.
\newblock Optimum branchings.
\newblock \emph{Journal of Research of the national Bureau of Standards B}, 71(4):233--240.

\bibitem[{Eisape et~al.(2022)Eisape, Gangireddy, Levy, and Kim}]{eisape-etal-2022-probing}
Tiwalayo Eisape, Vineet Gangireddy, Roger Levy, and Yoon Kim. 2022.
\newblock \href {https://doi.org/10.18653/v1/2022.findings-emnlp.203} {Probing for incremental parse states in autoregressive language models}.
\newblock In \emph{Findings of the Association for Computational Linguistics: EMNLP 2022}, pages 2801--2813, Abu Dhabi, United Arab Emirates. Association for Computational Linguistics.

\bibitem[{Ferrando et~al.(2023)Ferrando, G{\'a}llego, Tsiamas, and Costa-juss{\`a}}]{ferrando-etal-2023-explaining}
Javier Ferrando, Gerard~I. G{\'a}llego, Ioannis Tsiamas, and Marta~R. Costa-juss{\`a}. 2023.
\newblock \href {https://doi.org/10.18653/v1/2023.acl-long.301} {Explaining how transformers use context to build predictions}.
\newblock In \emph{Proceedings of the 61st Annual Meeting of the Association for Computational Linguistics (Volume 1: Long Papers)}, pages 5486--5513, Toronto, Canada. Association for Computational Linguistics.

\bibitem[{Frazier(1979)}]{frazier1979comprehending}
Lyn Frazier. 1979.
\newblock \emph{On comprehending sentences: Syntactic parsing strategies}.
\newblock Ph.D. thesis, University of Connecticut.

\bibitem[{Frazier and Rayner(1982)}]{frazier-rayner-1982}
Lyn Frazier and Keith Rayner. 1982.
\newblock \href {https://doi.org/https://doi.org/10.1016/0010-0285(82)90008-1} {Making and correcting errors during sentence comprehension: Eye movements in the analysis of structurally ambiguous sentences}.
\newblock \emph{Cognitive Psychology}, 14(2):178--210.

\bibitem[{Futrell et~al.(2019)Futrell, Wilcox, Morita, Qian, Ballesteros, and Levy}]{futrell-etal-2019-neural}
Richard Futrell, Ethan Wilcox, Takashi Morita, Peng Qian, Miguel Ballesteros, and Roger Levy. 2019.
\newblock \href {https://doi.org/10.18653/v1/N19-1004} {Neural language models as psycholinguistic subjects: Representations of syntactic state}.
\newblock In \emph{Proceedings of the 2019 Conference of the North {A}merican Chapter of the Association for Computational Linguistics: Human Language Technologies, Volume 1 (Long and Short Papers)}, pages 32--42, Minneapolis, Minnesota. Association for Computational Linguistics.

\bibitem[{Garcia et~al.(2021)Garcia, Kramer~Vieira, Scarton, Idiart, and Villavicencio}]{garcia-etal-2021-probing}
Marcos Garcia, Tiago Kramer~Vieira, Carolina Scarton, Marco Idiart, and Aline Villavicencio. 2021.
\newblock \href {https://doi.org/10.18653/v1/2021.eacl-main.310} {Probing for idiomaticity in vector space models}.
\newblock In \emph{Proceedings of the 16th Conference of the European Chapter of the Association for Computational Linguistics: Main Volume}, pages 3551--3564, Online. Association for Computational Linguistics.

\bibitem[{Geva et~al.(2022)Geva, Caciularu, Wang, and Goldberg}]{geva-etal-2022-transformer}
Mor Geva, Avi Caciularu, Kevin Wang, and Yoav Goldberg. 2022.
\newblock \href {https://doi.org/10.18653/v1/2022.emnlp-main.3} {Transformer feed-forward layers build predictions by promoting concepts in the vocabulary space}.
\newblock In \emph{Proceedings of the 2022 Conference on Empirical Methods in Natural Language Processing}, pages 30--45, Abu Dhabi, United Arab Emirates. Association for Computational Linguistics.

\bibitem[{Graves and Schmidhuber(2005)}]{graves2005framewise}
Alex Graves and J{\"u}rgen Schmidhuber. 2005.
\newblock \href {https://doi.org/10.1016/j.neunet.2005.06.042} {Framewise phoneme classification with bidirectional lstm and other neural network architectures}.
\newblock \emph{Neural networks}, 18(5-6):602--610.

\bibitem[{Hewitt and Manning(2019)}]{hewitt-manning-2019-structural}
John Hewitt and Christopher~D. Manning. 2019.
\newblock \href {https://doi.org/10.18653/v1/N19-1419} {{A} structural probe for finding syntax in word representations}.
\newblock In \emph{Proceedings of the 2019 Conference of the North {A}merican Chapter of the Association for Computational Linguistics: Human Language Technologies, Volume 1 (Long and Short Papers)}, pages 4129--4138, Minneapolis, Minnesota. Association for Computational Linguistics.

\bibitem[{Honnibal and Johnson(2014)}]{honnibal-johnson-2014-joint}
Matthew Honnibal and Mark Johnson. 2014.
\newblock \href {https://doi.org/10.1162/tacl_a_00171} {Joint incremental disfluency detection and dependency parsing}.
\newblock \emph{Transactions of the Association for Computational Linguistics}, 2:131--142.

\bibitem[{Hrycyk et~al.(2021)Hrycyk, Zarcone, and Hahn}]{hrycyk-etal-2021-fast}
Lianna Hrycyk, Alessandra Zarcone, and Luzian Hahn. 2021.
\newblock \href {https://doi.org/10.18653/v1/2021.nlp4convai-1.6} {Not so fast, classifier {--} accuracy and entropy reduction in incremental intent classification}.
\newblock In \emph{Proceedings of the 3rd Workshop on Natural Language Processing for Conversational AI}, pages 52--67, Online. Association for Computational Linguistics.

\bibitem[{Huang et~al.(2023)Huang, Arehalli, Kugemoto, Muxica, Prasad, Dillon, and Linzen}]{huang2023surprisal}
Kuan-Jung Huang, Suhas Arehalli, Mari Kugemoto, Christian Muxica, Grusha Prasad, Brian Dillon, and Tal Linzen. 2023.
\newblock \href {https://doi.org/10.31234/osf.io/z38u6} {Surprisal does not explain syntactic disambiguation difficulty: evidence from a large-scale benchmark}.

\bibitem[{Irwin et~al.(2023)Irwin, Wilson, and Marantz}]{irwin-etal-2023-bert}
Tovah Irwin, Kyra Wilson, and Alec Marantz. 2023.
\newblock \href {https://doi.org/10.18653/v1/2023.eacl-main.235} {{BERT} shows garden path effects}.
\newblock In \emph{Proceedings of the 17th Conference of the European Chapter of the Association for Computational Linguistics}, pages 3220--3232, Dubrovnik, Croatia. Association for Computational Linguistics.

\bibitem[{Jurayj et~al.(2022)Jurayj, Rudman, and Eickhoff}]{jurayj-etal-2022-garden}
William Jurayj, William Rudman, and Carsten Eickhoff. 2022.
\newblock \href {https://doi.org/10.18653/v1/2022.blackboxnlp-1.25} {Garden path traversal in {GPT}-2}.
\newblock In \emph{Proceedings of the Fifth BlackboxNLP Workshop on Analyzing and Interpreting Neural Networks for NLP}, pages 305--313, Abu Dhabi, United Arab Emirates (Hybrid). Association for Computational Linguistics.

\bibitem[{Kahardipraja et~al.(2021)Kahardipraja, Madureira, and Schlangen}]{kahardipraja-etal-2021-towards}
Patrick Kahardipraja, Brielen Madureira, and David Schlangen. 2021.
\newblock \href {https://doi.org/10.18653/v1/2021.emnlp-main.90} {Towards incremental transformers: An empirical analysis of transformer models for incremental {NLU}}.
\newblock In \emph{Proceedings of the 2021 Conference on Empirical Methods in Natural Language Processing}, pages 1178--1189, Online and Punta Cana, Dominican Republic. Association for Computational Linguistics.

\bibitem[{Kahardipraja et~al.(2023)Kahardipraja, Madureira, and Schlangen}]{kahardipraja-etal-2023-tapir}
Patrick Kahardipraja, Brielen Madureira, and David Schlangen. 2023.
\newblock \href {https://doi.org/10.18653/v1/2023.findings-acl.257} {{TAPIR}: Learning adaptive revision for incremental natural language understanding with a two-pass model}.
\newblock In \emph{Findings of the Association for Computational Linguistics: ACL 2023}, pages 4173--4197, Toronto, Canada. Association for Computational Linguistics.

\bibitem[{Kaushal et~al.(2023)Kaushal, Gupta, Upadhyay, and Faruqui}]{kaushal-etal-2023-efficient}
Ayush Kaushal, Aditya Gupta, Shyam Upadhyay, and Manaal Faruqui. 2023.
\newblock \href {https://doi.org/10.18653/v1/2023.eacl-main.31} {Efficient encoders for streaming sequence tagging}.
\newblock In \emph{Proceedings of the 17th Conference of the European Chapter of the Association for Computational Linguistics}, pages 418--429, Dubrovnik, Croatia. Association for Computational Linguistics.

\bibitem[{Kennington et~al.(2014)Kennington, Kousidis, and Schlangen}]{kennington-etal-2014-inprotks}
Casey Kennington, Spyros Kousidis, and David Schlangen. 2014.
\newblock \href {https://doi.org/10.3115/v1/W14-4312} {{I}npro{TK}s: A toolkit for incremental situated processing}.
\newblock In \emph{Proceedings of the 15th Annual Meeting of the Special Interest Group on Discourse and Dialogue ({SIGDIAL})}, pages 84--88, Philadelphia, PA, U.S.A. Association for Computational Linguistics.

\bibitem[{Khouzaimi et~al.(2014)Khouzaimi, Laroche, and Lefevre}]{khouzaimi-etal-2014-easy}
Hatim Khouzaimi, Romain Laroche, and Fabrice Lefevre. 2014.
\newblock \href {https://doi.org/10.3115/v1/W14-4314} {An easy method to make dialogue systems incremental}.
\newblock In \emph{Proceedings of the 15th Annual Meeting of the Special Interest Group on Discourse and Dialogue ({SIGDIAL})}, pages 98--107, Philadelphia, PA, U.S.A. Association for Computational Linguistics.

\bibitem[{Kilger and Finkler(1995)}]{kilger1995incremental}
Anne Kilger and Wolfgang Finkler. 1995.
\newblock \href {http://dx.doi.org/10.22028/D291-24965} {Incremental generation for real-time applications}.
\newblock \emph{DFKI Verbmobil Research Report RR-95-11}.

\bibitem[{Kitaev et~al.(2022)Kitaev, Lu, and Klein}]{kitaev-etal-2022-learned}
Nikita Kitaev, Thomas Lu, and Dan Klein. 2022.
\newblock \href {https://doi.org/10.18653/v1/2022.acl-long.220} {Learned incremental representations for parsing}.
\newblock In \emph{Proceedings of the 60th Annual Meeting of the Association for Computational Linguistics (Volume 1: Long Papers)}, pages 3086--3095, Dublin, Ireland. Association for Computational Linguistics.

\bibitem[{K{\"o}hn(2018)}]{kohn-2018-incremental}
Arne K{\"o}hn. 2018.
\newblock \href {https://aclanthology.org/C18-1253} {Incremental natural language processing: Challenges, strategies, and evaluation}.
\newblock In \emph{Proceedings of the 27th International Conference on Computational Linguistics}, pages 2990--3003, Santa Fe, New Mexico, USA. Association for Computational Linguistics.

\bibitem[{Kraus et~al.(1990)Kraus, Lehmann, and Magidor}]{KRAUS1990167}
Sarit Kraus, Daniel Lehmann, and Menachem Magidor. 1990.
\newblock \href {https://doi.org/https://doi.org/10.1016/0004-3702(90)90101-5} {Nonmonotonic reasoning, preferential models and cumulative logics}.
\newblock \emph{Artificial Intelligence}, 44(1):167--207.

\bibitem[{Leblond et~al.(2021)Leblond, Alayrac, Sifre, Pislar, Jean-Baptiste, Antonoglou, Simonyan, and Vinyals}]{leblond-etal-2021-machine}
R{\'e}mi Leblond, Jean-Baptiste Alayrac, Laurent Sifre, Miruna Pislar, Lespiau Jean-Baptiste, Ioannis Antonoglou, Karen Simonyan, and Oriol Vinyals. 2021.
\newblock \href {https://doi.org/10.18653/v1/2021.emnlp-main.662} {Machine translation decoding beyond beam search}.
\newblock In \emph{Proceedings of the 2021 Conference on Empirical Methods in Natural Language Processing}, pages 8410--8434, Online and Punta Cana, Dominican Republic. Association for Computational Linguistics.

\bibitem[{Lee and Shin(2023)}]{lee2023decoding}
Jonghyun Lee and Jeong-Ah Shin. 2023.
\newblock Decoding bert’s internal processing of garden-path structures through attention maps.
\newblock \emph{Korean Journal of English Language and Linguistics}, 23:461--481.

\bibitem[{Lee et~al.(2022)Lee, Shin, and Park}]{lee2022bert}
Jonghyun Lee, Jeong-Ah Shin, and Myung-Kwan Park. 2022.
\newblock (al)bert down the garden path: Psycholinguistic experiments for pre-trained language models.
\newblock \emph{Korean Journal of English Language and Linguistics}, 22:1033--1050.

\bibitem[{Levelt(1993)}]{levelt1993speaking}
Willem~JM Levelt. 1993.
\newblock \href {https://doi.org/10.7551/mitpress/6393.001.0001} {\emph{Speaking: From intention to articulation}}.
\newblock MIT press.

\bibitem[{Lindborg and Rabovsky(2021)}]{lindborg2021meaning}
Alma Lindborg and Milena Rabovsky. 2021.
\newblock \href {https://escholarship.org/uc/item/6d71c9sj} {Meaning in brains and machines: Internal activation update in large-scale language model partially reflects the n400 brain potential}.
\newblock In \emph{Proceedings of the annual meeting of the cognitive science society}, volume~43.

\bibitem[{Liu et~al.(2023)Liu, Wu, Michael, Suhr, West, Koller, Swayamdipta, Smith, and Choi}]{liu-etal-2023-afraid}
Alisa Liu, Zhaofeng Wu, Julian Michael, Alane Suhr, Peter West, Alexander Koller, Swabha Swayamdipta, Noah Smith, and Yejin Choi. 2023.
\newblock \href {https://doi.org/10.18653/v1/2023.emnlp-main.51} {We{'}re afraid language models aren{'}t modeling ambiguity}.
\newblock In \emph{Proceedings of the 2023 Conference on Empirical Methods in Natural Language Processing}, pages 790--807, Singapore. Association for Computational Linguistics.

\bibitem[{Liu et~al.(2019{\natexlab{a}})Liu, Gardner, Belinkov, Peters, and Smith}]{liu-etal-2019-linguistic}
Nelson~F. Liu, Matt Gardner, Yonatan Belinkov, Matthew~E. Peters, and Noah~A. Smith. 2019{\natexlab{a}}.
\newblock \href {https://doi.org/10.18653/v1/N19-1112} {Linguistic knowledge and transferability of contextual representations}.
\newblock In \emph{Proceedings of the 2019 Conference of the North {A}merican Chapter of the Association for Computational Linguistics: Human Language Technologies, Volume 1 (Long and Short Papers)}, pages 1073--1094, Minneapolis, Minnesota. Association for Computational Linguistics.

\bibitem[{Liu et~al.(2019{\natexlab{b}})Liu, Ott, Goyal, Du, Joshi, Chen, Levy, Lewis, Zettlemoyer, and Stoyanov}]{liu2019roberta}
Yinhan Liu, Myle Ott, Naman Goyal, Jingfei Du, Mandar Joshi, Danqi Chen, Omer Levy, Mike Lewis, Luke Zettlemoyer, and Veselin Stoyanov. 2019{\natexlab{b}}.
\newblock \href {http://arxiv.org/abs/1907.11692} {Roberta: A robustly optimized bert pretraining approach}.

\bibitem[{Madureira et~al.(2023)Madureira, Kahardipraja, and Schlangen}]{madureira-etal-2023-road}
Brielen Madureira, Patrick Kahardipraja, and David Schlangen. 2023.
\newblock \href {https://aclanthology.org/2023.sigdial-1.14} {The road to quality is paved with good revisions: A detailed evaluation methodology for revision policies in incremental sequence labelling}.
\newblock In \emph{Proceedings of the 24th Meeting of the Special Interest Group on Discourse and Dialogue}, pages 156--167, Prague, Czechia. Association for Computational Linguistics.

\bibitem[{Madureira and Schlangen(2020)}]{madureira-schlangen-2020-incremental}
Brielen Madureira and David Schlangen. 2020.
\newblock \href {https://doi.org/10.18653/v1/2020.emnlp-main.26} {Incremental processing in the age of non-incremental encoders: An empirical assessment of bidirectional models for incremental {NLU}}.
\newblock In \emph{Proceedings of the 2020 Conference on Empirical Methods in Natural Language Processing (EMNLP)}, pages 357--374, Online. Association for Computational Linguistics.

\bibitem[{Marcus et~al.(1993)Marcus, Santorini, and Marcinkiewicz}]{marcus-etal-1993-building}
Mitchell~P. Marcus, Beatrice Santorini, and Mary~Ann Marcinkiewicz. 1993.
\newblock \href {https://aclanthology.org/J93-2004} {Building a large annotated corpus of {E}nglish: The {P}enn {T}reebank}.
\newblock \emph{Computational Linguistics}, 19(2):313--330.

\bibitem[{Marslen-Wilson(1973)}]{marslen1973linguistic}
William Marslen-Wilson. 1973.
\newblock \href {https://doi.org/10.1038/244522a0} {Linguistic structure and speech shadowing at very short latencies}.
\newblock \emph{Nature}, 244(5417):522--523.

\bibitem[{Matthews(1975)}]{MATTHEWS1975442}
B.W. Matthews. 1975.
\newblock \href {https://doi.org/https://doi.org/10.1016/0005-2795(75)90109-9} {Comparison of the predicted and observed secondary structure of t4 phage lysozyme}.
\newblock \emph{Biochimica et Biophysica Acta (BBA) - Protein Structure}, 405(2):442--451.

\bibitem[{Nivre(2008)}]{nivre-2008-algorithms}
Joakim Nivre. 2008.
\newblock \href {https://doi.org/10.1162/coli.07-056-R1-07-027} {Algorithms for deterministic incremental dependency parsing}.
\newblock \emph{Computational Linguistics}, 34(4):513--553.

\bibitem[{Pal et~al.(2023)Pal, Sun, Yuan, Wallace, and Bau}]{pal-etal-2023-future}
Koyena Pal, Jiuding Sun, Andrew Yuan, Byron Wallace, and David Bau. 2023.
\newblock \href {https://doi.org/10.18653/v1/2023.conll-1.37} {Future lens: Anticipating subsequent tokens from a single hidden state}.
\newblock In \emph{Proceedings of the 27th Conference on Computational Natural Language Learning (CoNLL)}, pages 548--560, Singapore. Association for Computational Linguistics.

\bibitem[{Qi et~al.(2020)Qi, Zhang, Zhang, Bolton, and Manning}]{qi2020stanza}
Peng Qi, Yuhao Zhang, Yuhui Zhang, Jason Bolton, and Christopher~D. Manning. 2020.
\newblock Stanza: A {Python} natural language processing toolkit for many human languages.
\newblock In \emph{Proceedings of the 58th Annual Meeting of the Association for Computational Linguistics: System Demonstrations}.

\bibitem[{Qin et~al.(2020)Qin, Shwartz, West, Bhagavatula, Hwang, Le~Bras, Bosselut, and Choi}]{qin-etal-2020-back}
Lianhui Qin, Vered Shwartz, Peter West, Chandra Bhagavatula, Jena~D. Hwang, Ronan Le~Bras, Antoine Bosselut, and Yejin Choi. 2020.
\newblock \href {https://doi.org/10.18653/v1/2020.emnlp-main.58} {Back to the future: Unsupervised backprop-based decoding for counterfactual and abductive commonsense reasoning}.
\newblock In \emph{Proceedings of the 2020 Conference on Empirical Methods in Natural Language Processing (EMNLP)}, pages 794--805, Online. Association for Computational Linguistics.

\bibitem[{Radford et~al.(2019)Radford, Wu, Child, Luan, Amodei, and Sutskever}]{radford2019language}
Alec Radford, Jeff Wu, Rewon Child, David Luan, Dario Amodei, and Ilya Sutskever. 2019.
\newblock Language models are unsupervised multitask learners.

\bibitem[{Rafla and Kennington(2019)}]{rafla2019incrementalizing}
Andrew Rafla and Casey Kennington. 2019.
\newblock \href {https://doi.org/10.48550/arXiv.1907.05403} {Incrementalizing rasa's open-source natural language understanding pipeline}.
\newblock \emph{arXiv preprint arXiv:1907.05403}.

\bibitem[{Rogers et~al.(2020)Rogers, Kovaleva, and Rumshisky}]{rogers-etal-2020-primer}
Anna Rogers, Olga Kovaleva, and Anna Rumshisky. 2020.
\newblock \href {https://doi.org/10.1162/tacl_a_00349} {A primer in {BERT}ology: What we know about how {BERT} works}.
\newblock \emph{Transactions of the Association for Computational Linguistics}, 8:842--866.

\bibitem[{Ryu et~al.(2006)Ryu, Matsubara, and Inagaki}]{ryu-etal-2006-simultaneous}
Koichiro Ryu, Shigeki Matsubara, and Yasuyoshi Inagaki. 2006.
\newblock \href {https://aclanthology.org/P06-2088} {Simultaneous {E}nglish-{J}apanese spoken language translation based on incremental dependency parsing and transfer}.
\newblock In \emph{Proceedings of the {COLING}/{ACL} 2006 Main Conference Poster Sessions}, pages 683--690, Sydney, Australia. Association for Computational Linguistics.

\bibitem[{Saphra and Lopez(2019)}]{saphra-lopez-2019-understanding}
Naomi Saphra and Adam Lopez. 2019.
\newblock \href {https://doi.org/10.18653/v1/N19-1329} {Understanding learning dynamics of language models with {SVCCA}}.
\newblock In \emph{Proceedings of the 2019 Conference of the North {A}merican Chapter of the Association for Computational Linguistics: Human Language Technologies, Volume 1 (Long and Short Papers)}, pages 3257--3267, Minneapolis, Minnesota. Association for Computational Linguistics.

\bibitem[{Schlangen(2023)}]{schlangen-2023-general}
David Schlangen. 2023.
\newblock \href {https://doi.org/10.18653/v1/2023.findings-emnlp.591} {On general language understanding}.
\newblock In \emph{Findings of the Association for Computational Linguistics: EMNLP 2023}, pages 8818--8825, Singapore. Association for Computational Linguistics.

\bibitem[{Schlangen and Skantze(2011)}]{schlangen2011general}
David Schlangen and Gabriel Skantze. 2011.
\newblock \href {https://doi.org/10.5087/dad.2011.105} {A general, abstract model of incremental dialogue processing}.
\newblock \emph{Dialogue \& Discourse}, 2(1):83--111.

\bibitem[{Sen et~al.(2023)Sen, Sennrich, Zhang, and Haddow}]{sen-etal-2023-self}
Sukanta Sen, Rico Sennrich, Biao Zhang, and Barry Haddow. 2023.
\newblock \href {https://doi.org/10.18653/v1/2023.eacl-main.270} {Self-training reduces flicker in retranslation-based simultaneous translation}.
\newblock In \emph{Proceedings of the 17th Conference of the European Chapter of the Association for Computational Linguistics}, pages 3734--3744, Dubrovnik, Croatia. Association for Computational Linguistics.

\bibitem[{Silveira et~al.(2014)Silveira, Dozat, de~Marneffe, Bowman, Connor, Bauer, and Manning}]{silveira-etal-2014-gold}
Natalia Silveira, Timothy Dozat, Marie-Catherine de~Marneffe, Samuel Bowman, Miriam Connor, John Bauer, and Chris Manning. 2014.
\newblock \href {http://www.lrec-conf.org/proceedings/lrec2014/pdf/1089_Paper.pdf} {A gold standard dependency corpus for {E}nglish}.
\newblock In \emph{Proceedings of the Ninth International Conference on Language Resources and Evaluation ({LREC}'14)}, pages 2897--2904, Reykjavik, Iceland. European Language Resources Association (ELRA).

\bibitem[{Spoustová and Spousta(2010)}]{spoustova2010dependency}
Drahomíra Spoustová and Miroslav Spousta. 2010.
\newblock \href {https://doi.org/10.2478/v10108-010-0017-3} {Dependency parsing as a sequence labeling task}.
\newblock \emph{The Prague Bulletin of Mathematical Linguistics}, 94.

\bibitem[{Springer et~al.(2024)Springer, Kotha, Fried, Neubig, and Raghunathan}]{springer2024repetition}
Jacob~Mitchell Springer, Suhas Kotha, Daniel Fried, Graham Neubig, and Aditi Raghunathan. 2024.
\newblock \href {https://doi.org/10.48550/arXiv.2402.15449} {Repetition improves language model embeddings}.
\newblock \emph{arXiv preprint arXiv:2402.15449}.

\bibitem[{Strzyz et~al.(2019)Strzyz, Vilares, and G{\'o}mez-Rodr{\'\i}guez}]{strzyz-etal-2019-viable}
Michalina Strzyz, David Vilares, and Carlos G{\'o}mez-Rodr{\'\i}guez. 2019.
\newblock \href {https://doi.org/10.18653/v1/N19-1077} {Viable dependency parsing as sequence labeling}.
\newblock In \emph{Proceedings of the 2019 Conference of the North {A}merican Chapter of the Association for Computational Linguistics: Human Language Technologies, Volume 1 (Long and Short Papers)}, pages 717--723, Minneapolis, Minnesota. Association for Computational Linguistics.

\bibitem[{Tenney et~al.(2019)Tenney, Das, and Pavlick}]{tenney-etal-2019-bert}
Ian Tenney, Dipanjan Das, and Ellie Pavlick. 2019.
\newblock \href {https://doi.org/10.18653/v1/P19-1452} {{BERT} rediscovers the classical {NLP} pipeline}.
\newblock In \emph{Proceedings of the 57th Annual Meeting of the Association for Computational Linguistics}, pages 4593--4601, Florence, Italy. Association for Computational Linguistics.

\bibitem[{Ulmer et~al.(2019)Ulmer, Hupkes, and Bruni}]{ulmer-etal-2019-assessing}
Dennis Ulmer, Dieuwke Hupkes, and Elia Bruni. 2019.
\newblock \href {https://doi.org/10.18653/v1/W19-4324} {Assessing incrementality in sequence-to-sequence models}.
\newblock In \emph{Proceedings of the 4th Workshop on Representation Learning for NLP (RepL4NLP-2019)}, pages 209--217, Florence, Italy. Association for Computational Linguistics.

\bibitem[{Van~Schijndel and Linzen(2021)}]{van2021single}
Marten Van~Schijndel and Tal Linzen. 2021.
\newblock \href {https://doi.org/10.1111/cogs.12988} {Single-stage prediction models do not explain the magnitude of syntactic disambiguation difficulty}.
\newblock \emph{Cognitive science}, 45(6):e12988.

\bibitem[{Vaswani et~al.(2017)Vaswani, Shazeer, Parmar, Uszkoreit, Jones, Gomez, Kaiser, and Polosukhin}]{vaswani2017attention}
Ashish Vaswani, Noam Shazeer, Niki Parmar, Jakob Uszkoreit, Llion Jones, Aidan~N. Gomez, \L{}ukasz Kaiser, and Illia Polosukhin. 2017.
\newblock \href {https://dl.acm.org/doi/abs/10.5555/3295222.3295349} {Attention is all you need}.
\newblock In \emph{Proceedings of the 31st International Conference on Neural Information Processing Systems}, NIPS'17, page 6000–6010, Red Hook, NY, USA. Curran Associates Inc.

\bibitem[{Wijnholds and Moortgat(2023)}]{wijnholds-moortgat-2023-structural}
Gijs Wijnholds and Michael Moortgat. 2023.
\newblock \href {https://doi.org/10.18653/v1/2023.conll-1.11} {Structural ambiguity and its disambiguation in language model based parsers: the case of {D}utch clause relativization}.
\newblock In \emph{Proceedings of the 27th Conference on Computational Natural Language Learning (CoNLL)}, pages 155--164, Singapore. Association for Computational Linguistics.

\bibitem[{Wilcox et~al.(2021)Wilcox, Vani, and Levy}]{wilcox-etal-2021-targeted}
Ethan Wilcox, Pranali Vani, and Roger Levy. 2021.
\newblock \href {https://doi.org/10.18653/v1/2021.acl-long.76} {A targeted assessment of incremental processing in neural language models and humans}.
\newblock In \emph{Proceedings of the 59th Annual Meeting of the Association for Computational Linguistics and the 11th International Joint Conference on Natural Language Processing (Volume 1: Long Papers)}, pages 939--952, Online. Association for Computational Linguistics.

\bibitem[{Yom~Din et~al.(2024)Yom~Din, Karidi, Choshen, and Geva}]{din2023jump}
Alexander Yom~Din, Taelin Karidi, Leshem Choshen, and Mor Geva. 2024.
\newblock \href {https://aclanthology.org/2024.lrec-main.840} {Jump to conclusions: Short-cutting transformers with linear transformations}.
\newblock In \emph{Proceedings of the 2024 Joint International Conference on Computational Linguistics, Language Resources and Evaluation (LREC-COLING 2024)}, pages 9615--9625, Torino, Italia. ELRA and ICCL.

\bibitem[{Yoshida and Gimpel(2021)}]{yoshida-gimpel-2021-reconsidering-past}
Davis Yoshida and Kevin Gimpel. 2021.
\newblock \href {https://doi.org/10.18653/v1/2021.findings-emnlp.346} {Reconsidering the past: Optimizing hidden states in language models}.
\newblock In \emph{Findings of the Association for Computational Linguistics: EMNLP 2021}, pages 4099--4105, Punta Cana, Dominican Republic. Association for Computational Linguistics.

\bibitem[{Zhang et~al.(2022)Zhang, Roller, Goyal, Artetxe, Chen, Chen, Dewan, Diab, Li, Lin et~al.}]{zhang2022opt}
Susan Zhang, Stephen Roller, Naman Goyal, Mikel Artetxe, Moya Chen, Shuohui Chen, Christopher Dewan, Mona Diab, Xian Li, Xi~Victoria Lin, et~al. 2022.
\newblock \href {https://doi.org/10.48550/arXiv.2205.01068} {Opt: Open pre-trained transformer language models}.
\newblock \emph{arXiv preprint arXiv:2205.01068}.

\bibitem[{Zhou et~al.(2022)Zhou, Ethayarajh, Card, and Jurafsky}]{zhou-etal-2022-problems}
Kaitlyn Zhou, Kawin Ethayarajh, Dallas Card, and Dan Jurafsky. 2022.
\newblock \href {https://doi.org/10.18653/v1/2022.acl-short.45} {Problems with cosine as a measure of embedding similarity for high frequency words}.
\newblock In \emph{Proceedings of the 60th Annual Meeting of the Association for Computational Linguistics (Volume 2: Short Papers)}, pages 401--423, Dublin, Ireland. Association for Computational Linguistics.

\end{thebibliography}

\newpage
\appendix

\section{Appendix}
\label{sec:appendix}
In this section, we provide some additional details and results.

\paragraph{Classic garden paths} The NP/S garden path was originally discussed by \citet{frazier1979comprehending} and the MVRR by \citet{bever1970}. We did not use the other forms of garden path in the SAP benchmark \citep{huang2023surprisal} because the type of revision would be harder to isolate or, in the case of NP/Z and subject-verb agreement mismatch, the stimulus is not a completely well-formed construction in written text.

\paragraph{NNC} The template we selected for the stimuli is \texttt{This is a noun1 noun2,} and for the baseline reference is \texttt{This is a noun1,\ }. This was chosen to be a not very informative left context in order to focus on the effect of the NP construction.

\paragraph{Licenses} The NNC stimuli\footnote{\url{https://github.com/marcospln/noun_compound_senses}} are released without a license. The NP/S and MVRR stimuli\footnote{\url{https://github.com/caplabnyu/sapbenchmark}} and the repository as a whole is under the MIT license. BERT and ELECTRA are released under Apache 2.0. RoBERTa and GPT-2 are under MIT. OPT is under a custom OPT-175B license agreement. The dependency parsing libraries we used are under the MIT license.

\section{Details: Incremental Construction of Meaning}
\label{sec:appendix_meaning}
We use the pretrained model checkpoints and corresponding tokenizers available on HuggingFace: \texttt{bert-base-uncased},\footnote{\url{https://huggingface.co/bert-base-uncased}} \texttt{roberta-base},\footnote{\url{https://huggingface.co/roberta-base}} \texttt{gpt2},\footnote{\url{https://huggingface.co/gpt2}} and \texttt{facebook/opt-125m}.\footnote{\url{https://huggingface.co/facebook/opt-125m}}

We did not include the stimuli for which tokens were split into subtokens, because that creates misalignment in the triangular structures and thus requires workarounds. Because of that, the samples differed slightly for each model. This is not a problem in our analysis because we are not ranking the performance of the models; what matters is the intrinsic behaviour of each model separately. The number of excluded instances is shown in Table \ref{tab:excluded}.

\begin{table}[ht!]
\centering
\small
{\setlength{\tabcolsep}{3pt}
\begin{tabular}{l l c}
    \toprule
    \textbf{model} & \textbf{source} & \textbf{$n$ excluded} \\
    \midrule
    BERT & NP/S & 5 \\
    BERT & MVRR & 4 \\
    BERT & NNC & 14 \\
    RoBERTa & NP/S & 1 \\
    RoBERTa & MVRR & 0 \\
    RoBERTa & NNC & 23 \\

    opt & NP/S & 1 \\
    opt & MVRR & 0 \\
    opt & NNC & 23 \\

    gpt2 & NP/S & 1 \\
    gpt2 & MVRR & 0 \\
    gpt2 & NNC & 23 \\

    \bottomrule
\end{tabular}
}
\caption{Number of excluded instances for each model and type of stimuli due to subtokenisation.}
\label{tab:excluded}

\end{table}

Cosine distance was chosen because that should capture meaning variations in the embedding space. It is also possible to use other distance metrics like Manhattan or Euclidean distance. We used the \texttt{paired\_distances} method from the \texttt{sklearn.metrics.pairwise} package\footnote{\url{https://scikit-learn.org/stable/modules/generated/sklearn.metrics.pairwise.paired_distances.html\#sklearn.metrics.pairwise.paired_distances}} to compute the cosine distance.

\subsection{Computations}

We now describe in greater detail the computation steps we took to derive the plots the results they illustrate.

Each sentence was fed to each model prefix by prefix and after each step we saved the hidden representations. We stored the representations of all layers using 4D arrays with dimensions number of layers, sequence length (time step), sequence length (token position), embedding dimension. All models had an initial embedding layer and 12 encoding layers.

\paragraph{NNC} We extracted the states for all stimuli and their corresponding baseline. Then, we computed the cosine distance of each state to its own version in the preceding time step. For the main diagonal (\textit{i.e.}~the first time a token's state is constructed), we set this value to be 0 for a better visualisation. This resulted in a 2D lower diagonal matrix where each cell contains one value. Let $c_s$ and $c_b$ be these charts for the stimulus and its baseline, respectively. $c_s$ has one row and one column more than $c_b$, since the baseline does not contain the second noun and is thus one token shorter. We create $c_s'$ by deleting the last row and last column in $c_s$ (which correspond to the comma) and then compute the absolute difference $d=|c_s' - c_b|$. We then average the numbers in $d$ over all stimulus+baseline pairs for each layer.

\paragraph{NP/S and MVRR} We extracted the states for all stimuli and their corresponding baseline. Then, we computed the cosine distance of each state to its own version in the last time step, which represents its meaning given the full sentence as context. This resulted in a 2D lower diagonal matrix where each cell contains one value. Let $c_s$ and $c_b$ be these charts for the stimulus and its baseline, respectively. Here, $c_b$ has more rows/columns due to the added tokens that disambiguate the verb (\textit{i.e.}~\textit{that} or \textit{who was}). We create $c_b'$ by deleting the extra row(s) and column(s) in $c_b$ (which correspond to the added tokens) and then compute the absolute difference $d=|c_s - c_b'|$. We then average the numbers in $d$ over all stimulus+baseline pairs for each layer. Some sentences start with a \texttt{det adj noun} and other with \texttt{det noun}. For the plots, we remove the initial rows/columns and begin the chart at the aligned noun.

\paragraph{Causal} For the analysis of causal embeddings, we use four variations of each sentence: (a) the stimulus with a temporary ambiguity; (b) the baseline, which disambiguates the role of the verb and NP in advance; (c) the stimulus with the first verb replaced with an unambiguous verb; (d) the baseline with the first verb replaced by the same unambiguous verb. For NP/S, we use the verb \textit{said} and for MVRR, \textit{given}. For example, for NP/S:

\begin{enumerate}[label=\textbf{(\alph*)}]
    \item The new doctor demonstrated the operation appeared increasingly likely to succeed.
    \item The new doctor demonstrated that the operation appeared increasingly likely to succeed.
    \item The new doctor said the operation appeared increasingly likely to succeed.
    \item The new doctor said that the operation appeared increasingly likely to succeed.
\end{enumerate}

and for MVRR:

\begin{enumerate}[label=\textbf{(\alph*)}]
    \item The professor awarded the grant gained more attention from marine biologists.
    \item The professor who was awarded the grant gained more attention from marine biologists.
    \item The professor given the grant gained more attention from marine biologists.
    \item The professor who was given the grant gained more attention from marine biologists.
\end{enumerate}

We first compute the distance of the states of the tokens in (a) to their corresponding states in (b) (\textit{i.e.}~the states of the added tokens are ignored). Let us call the resulting vector $d_{ab}$. To account for the expected variation due to the different number of tokens, we do the same for (c) and (d), to be used as a reference, and get $d_{cd}$. Then, we take the absolute difference $|d_{ab} - d_{cd}|$ and average values over all stimuli.

To conclude, we detail the procedure to compute the numbers in Table \ref{tab:variation}. We first create triangular structures for all baseline sentences by computing the cosine distance of each state to its own version in the immediately preceding time step. The cell at row $i$ and column $j$ in the right triangle thus contain a value representing how much token $w_i$ has affected the state $s_j$ of token $w_j$, with $i>j$. We then average the distances for all tokens that have a corresponding state $1, 2, \dots, 7$ steps into the future. In practice, this means collecting values at each sub-diagonal $-1, -2, \dots, -7$ and computing the average over baseline stimuli.

\subsection{Additional Results}

\begin{figure}[ht]
    \centering
    \includegraphics[trim={0cm 0cm 0cm 0cm},clip,width=\columnwidth,page=1]{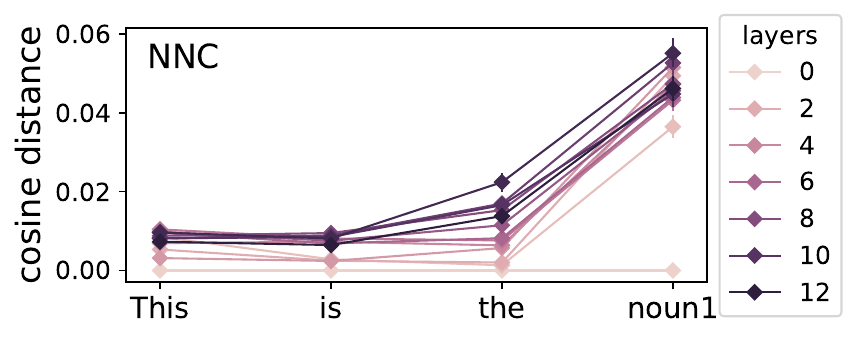}
    \caption{\textbf{NNC}. Average (absolute) effect over a baseline of the second noun on the tokens in the prefix for RoBERTa.}
    \label{fig:nnc-roberta}
\end{figure}

\begin{figure*}[ht]
    \centering
    \includegraphics[trim={0cm 0cm 0cm 0cm},clip,width=\linewidth,page=1]{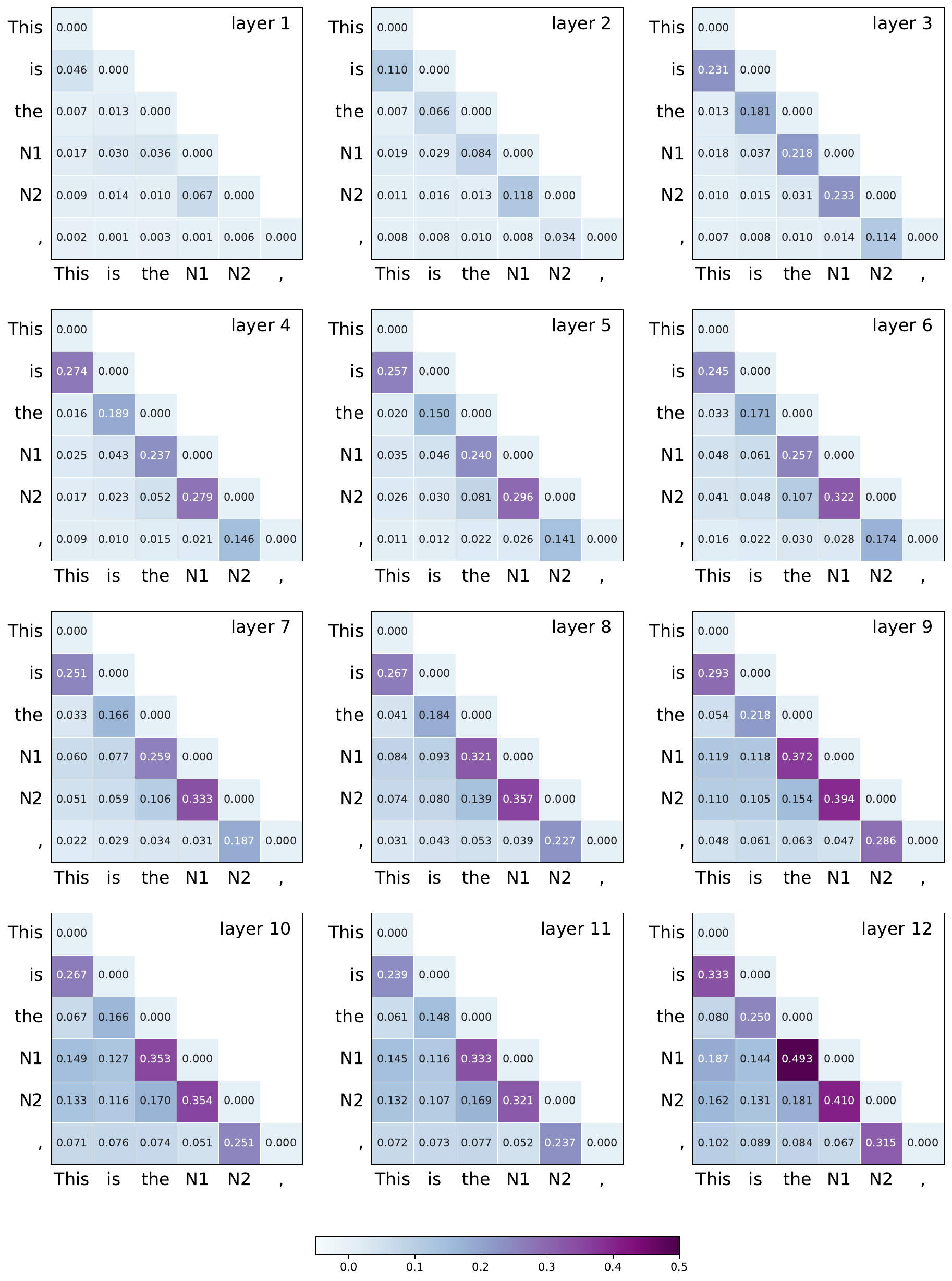}
    \caption{Overview of the mean cosine distance of a contextual embedding to its state in the preceding time step, for all layers of BERT, averaged over all NNC stimuli.}
    \label{fig:nnc-overview-bert}
\end{figure*}

\begin{figure*}[ht]
    \centering
    \includegraphics[trim={0cm 0cm 0cm 0cm},clip,width=\linewidth,page=1]{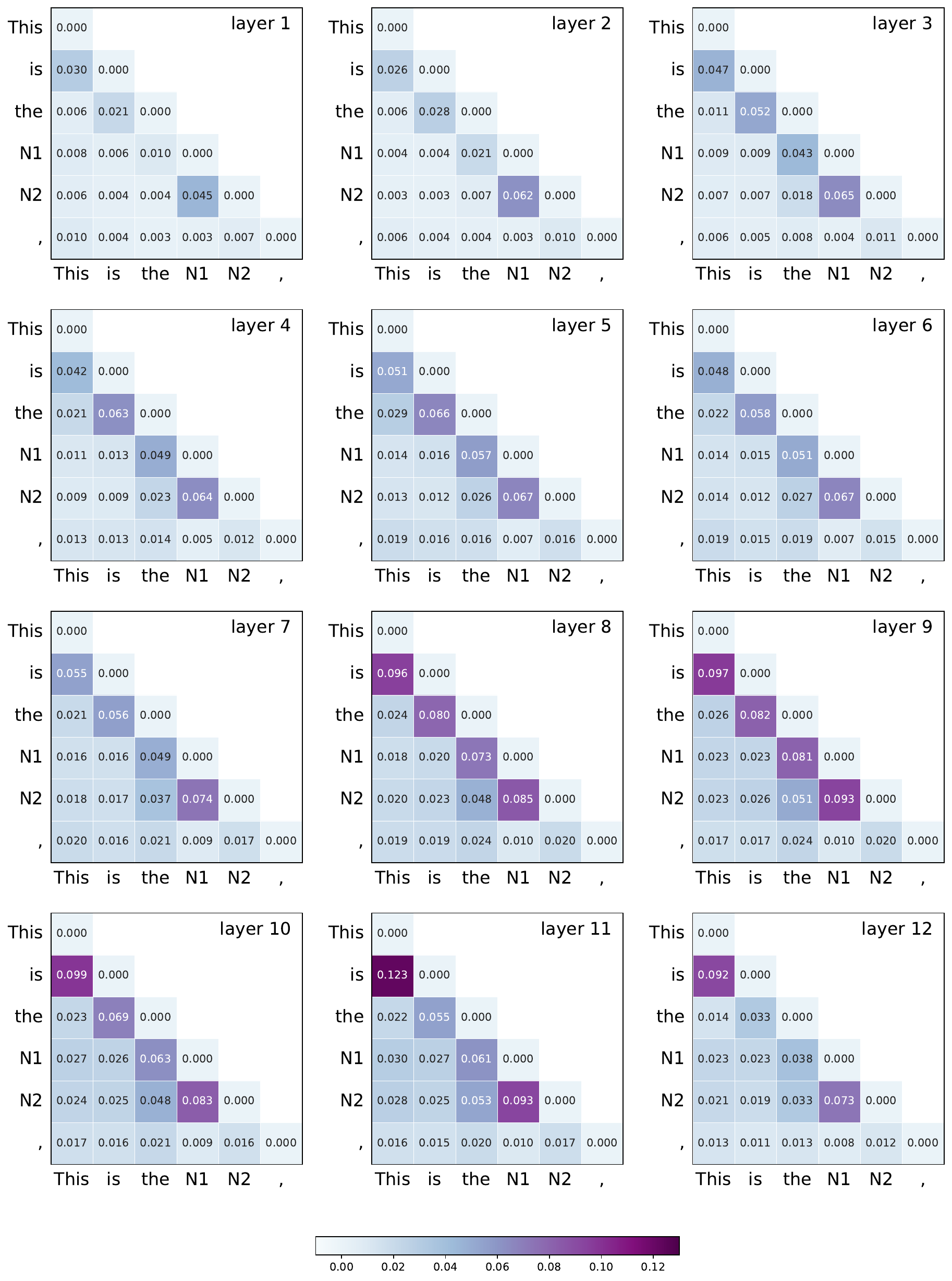}
    \caption{Overview of the mean cosine distance of a contextual embedding to its state in the preceding time step, for all layers of RoBERTa, averaged over all NNC stimuli.}
    \label{fig:nnc-overview-roberta}
\end{figure*}

In BERT, for layers 1 to 10, the maximum average change occurs at the first noun when the second noun is observed. For the upper layers, it is the second most, but still with a mean distance of almost 0.41, considerably above the corresponding 0.34 in Table \ref{tab:variation}. In the full overview in Figure \ref{fig:nnc-overview-bert} we see that all previous tokens are affected by all right tokens, especially during the construction of the NP. In Figure \ref{fig:nnc-roberta}, we see that RoBERTa has a similar effect on the NNC stimuli as BERT: the second noun influences the whole prefix, especially the first noun. Here, however, the magnitude of the effect is smaller, roughly half that of BERT. Besides, the upper layers have effects as high as the middle layers, different from BERT, where two of the middle layers stand out. Figure \ref{fig:nnc-overview-roberta} shows again that the magnitude of the state distances is smaller for RoBERTa. Up to layer 7, the effect of the second noun on the first one is the largest. However, in the upper layers, this no longer holds: The largest variation occurs for \textit{this} when \textit{is} is observed. 

\begin{figure*}[ht]
    \centering
    \includegraphics[trim={0cm 0cm 0cm 0cm},clip,width=\linewidth,page=1]{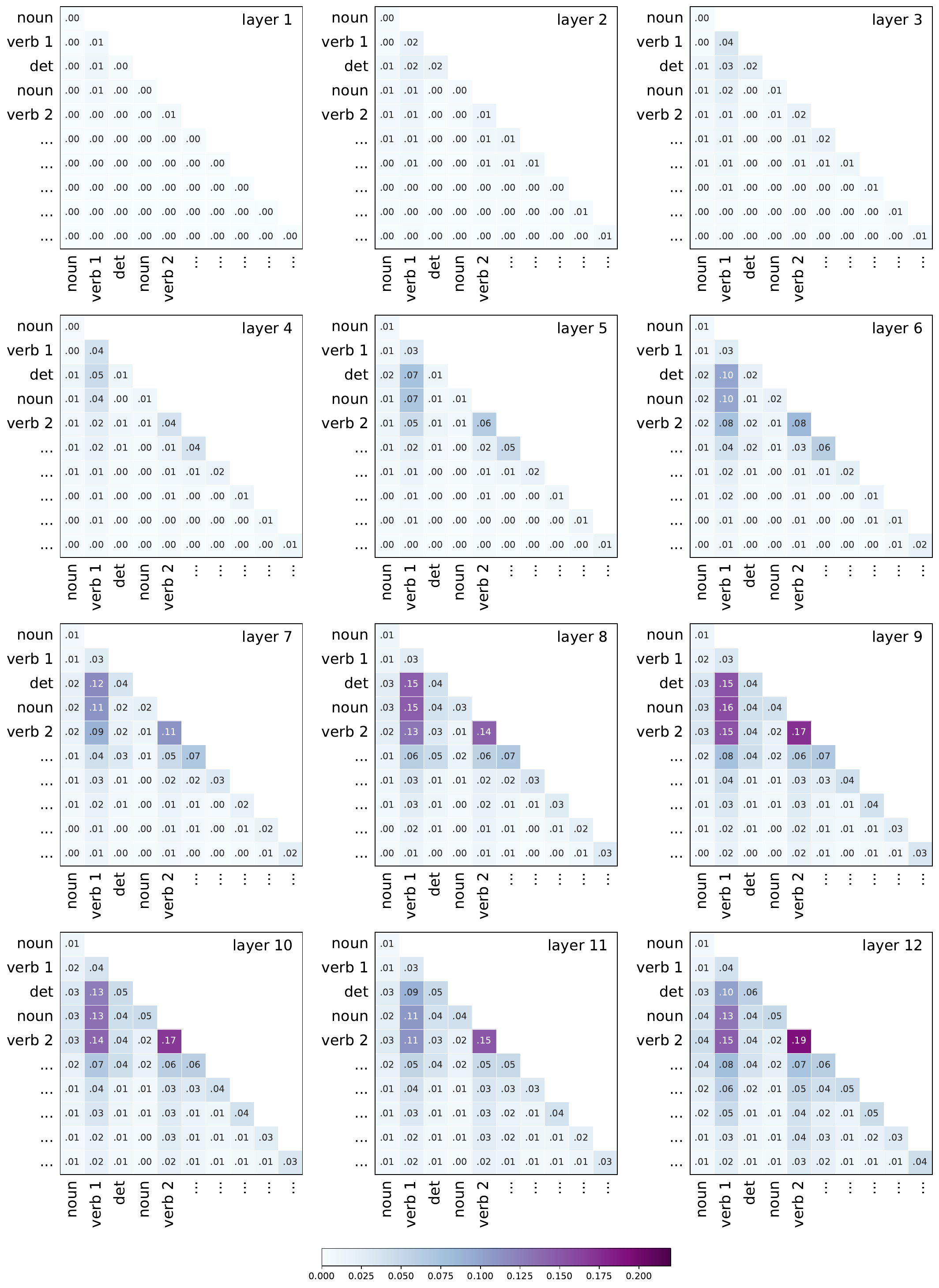}
    \caption{Overview of the cosine distance of a contextual embedding to its final state (last row). The numbers represent the absolute difference over the unambiguous baseline, for all layers of BERT, averaged over all NP/S stimuli.}
    \label{fig:nps-overview-bert}
\end{figure*}

\begin{figure*}[ht]
    \centering
    \includegraphics[trim={0cm 0cm 0cm 0cm},clip,width=\linewidth,page=1]{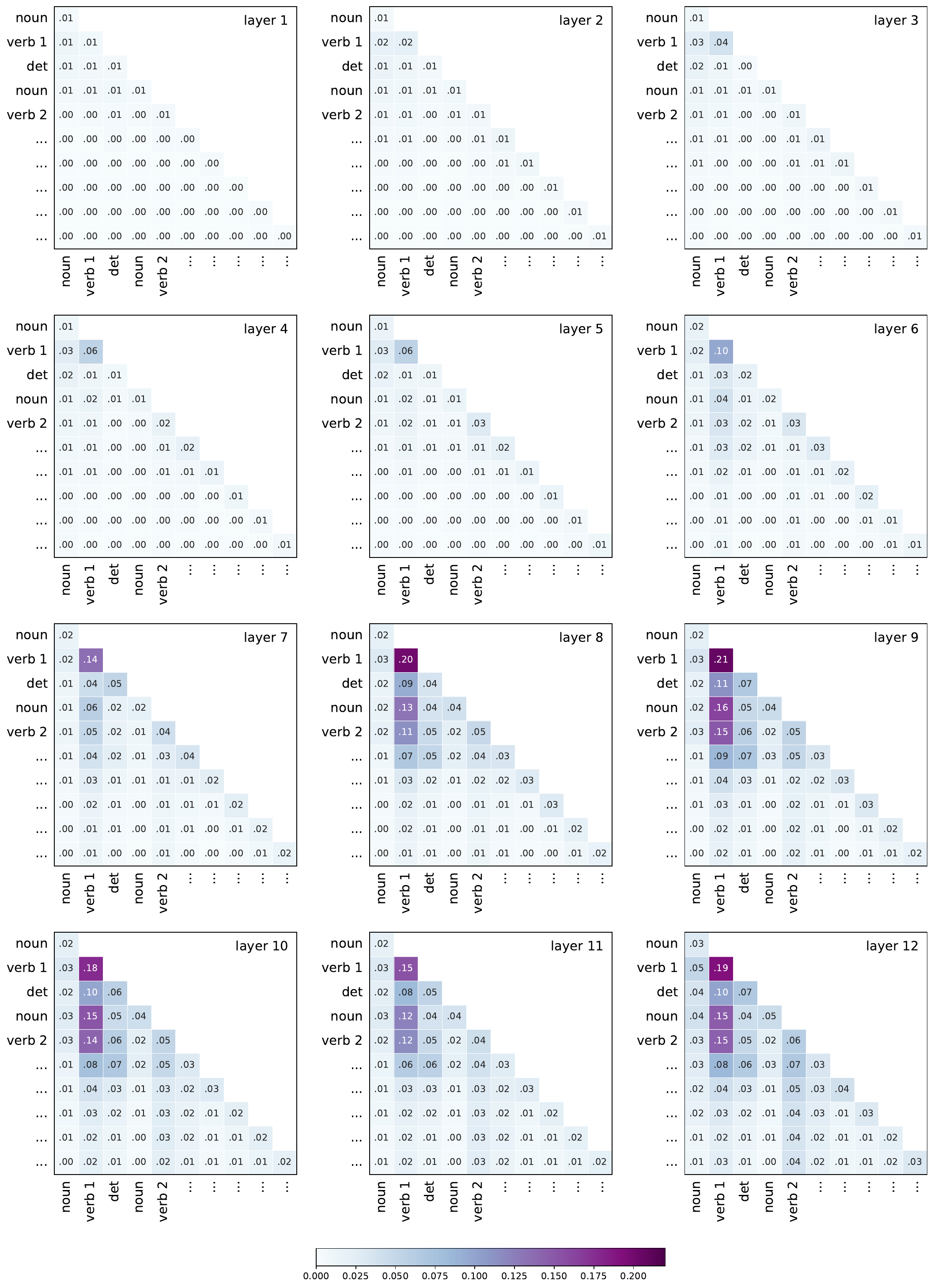}
    \caption{Overview of the cosine distance of a contextual embedding to its final state (last row). The numbers represent the absolute difference over the unambiguous baseline, for all layers of BERT, averaged over all MVRR stimuli.}
    \label{fig:mvrr-overview-bert}
\end{figure*}

\begin{figure*}[ht]
    \centering
    \includegraphics[trim={0cm 0cm 0cm 0cm},clip,width=\linewidth,page=1]{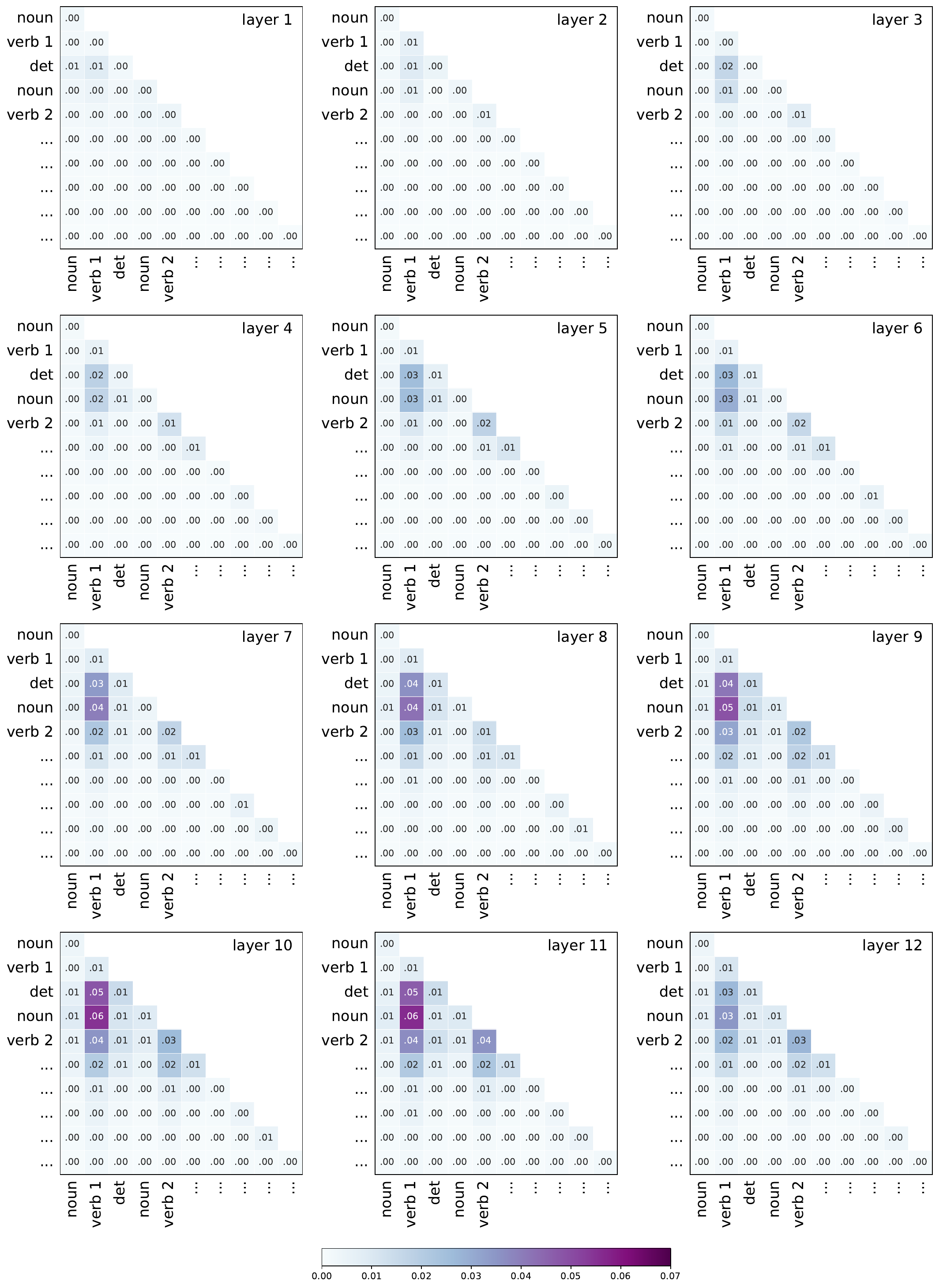}
    \caption{Overview of the cosine distance of a contextual embedding to its final state (last row). The numbers represent the absolute difference over the unambiguous baseline, for all layers of RoBERTa, averaged over all NP/S stimuli.}
    \label{fig:nps-overview-roberta}
\end{figure*}

\begin{figure*}[ht]
    \centering
    \includegraphics[trim={0cm 0cm 0cm 0cm},clip,width=\linewidth,page=1]{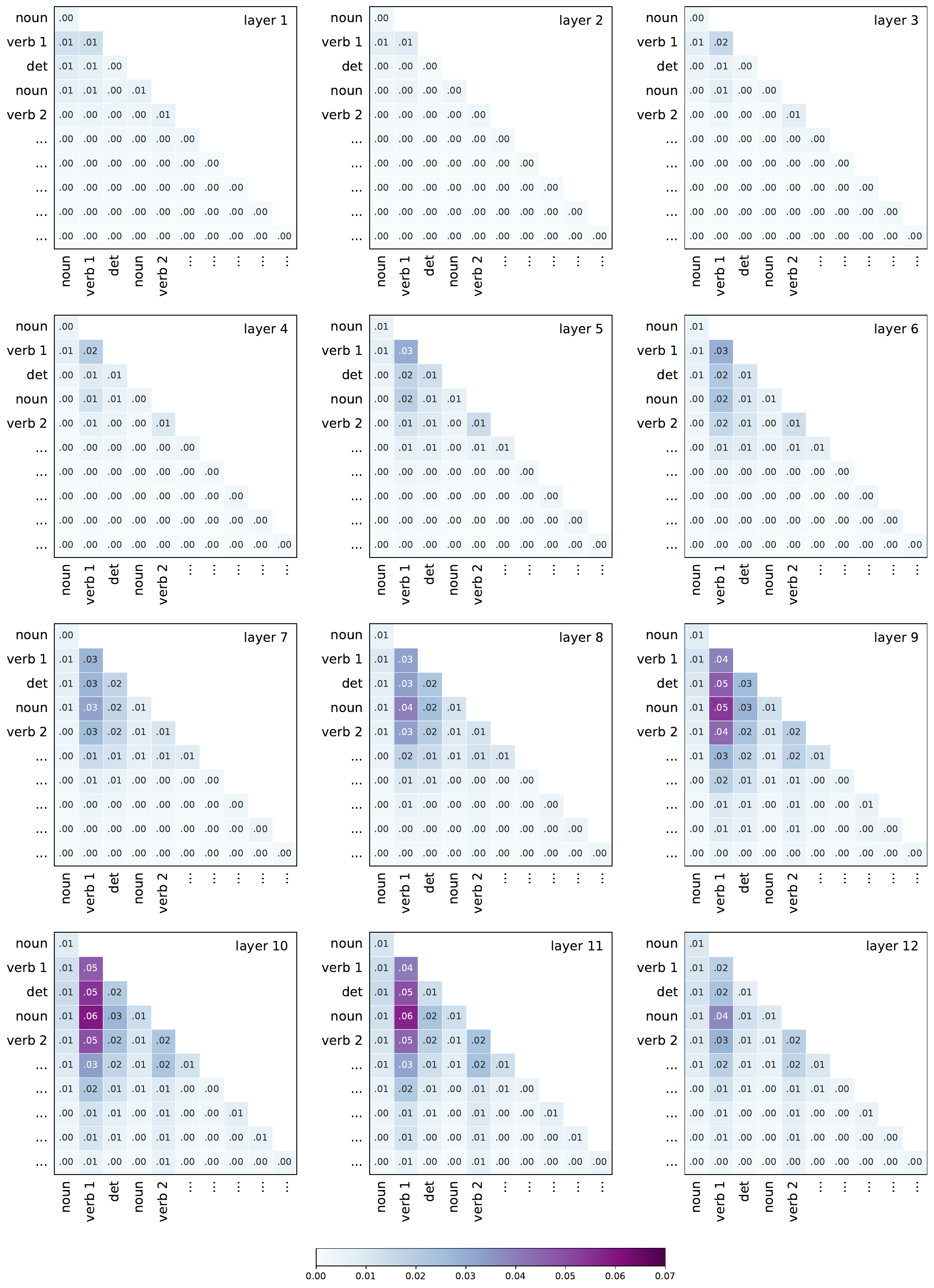}
    \caption{Overview of the cosine distance of a contextual embedding to its final state (last row). The numbers represent the absolute difference over the unambiguous baseline, for all layers of RoBERTa, averaged over all MVRR stimuli.}
    \label{fig:mvrr-overview-roberta}
\end{figure*}

Figures \ref{fig:nps-overview-bert} and \ref{fig:mvrr-overview-bert} show the full results of the NP/S and MVRR cases for BERT. For RoBERTa, the patterns of behaviour are similar: The representation of the first verb differs from what will be the final one, until the second verb, and one token after that, are integrated. Again, while BERT has an effect of around 0.15, RoBERTa's effect is around 0.05, as illustrated in Figure \ref{fig:nps-overview-roberta} and \ref{fig:mvrr-overview-roberta}. In both types of ambiguity, the effect becomes smaller in the last layer for RoBERTa, while it persists in BERT. More investigation is needed to shed light on what conceptual differences between the models cause the different behaviours.

\begin{figure}[ht]
    \centering
    \frame{\includegraphics[trim={0cm 0cm 0cm 0cm},clip,width=0.49\columnwidth,page=1]{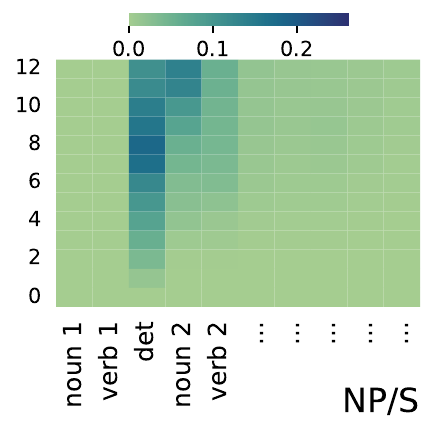}}
    \frame{\includegraphics[trim={0cm 0cm 0cm 0cm},clip,width=0.49\columnwidth,page=1]{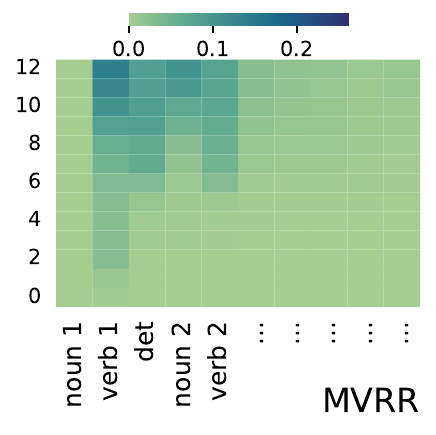}}
    
    \caption{Distance between the causal embeddings of the stimulus and the baseline, after subtracting the expected variation by a counterpart pair with an unambiguous verb, for all layers of OPT.}
    \label{fig:causal-opt}
\end{figure}

\begin{figure}[ht]
    \centering
    \frame{\includegraphics[trim={0cm 0cm 0cm 0cm},clip,width=0.49\columnwidth,page=1]{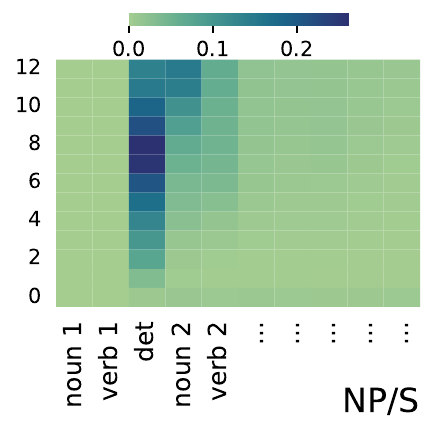}}
    \frame{\includegraphics[trim={0cm 0cm 0cm 0cm},clip,width=0.49\columnwidth,page=1]{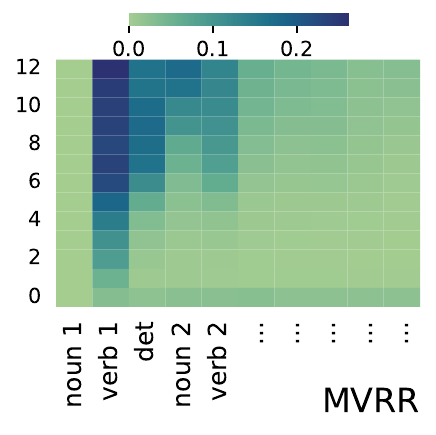}}
    
    \caption{Distance between the causal embeddings of the stimulus and the baseline, for all layers of OPT.}
    \label{fig:causal-opt-dist}
\end{figure}

\begin{figure}[ht]
    \centering
    \frame{\includegraphics[trim={0cm 0cm 0cm 0cm},clip,width=0.49\columnwidth,page=1]{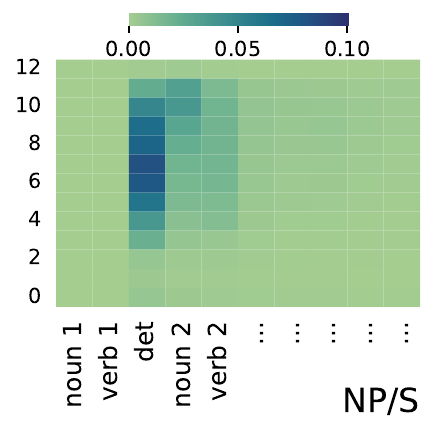}}
    \frame{\includegraphics[trim={0cm 0cm 0cm 0cm},clip,width=0.49\columnwidth,page=1]{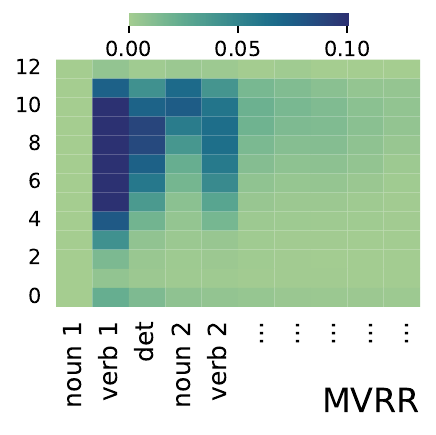}}
    
    \caption{Distance between the causal embeddings of the stimulus and the baseline, for all layers of GPT-2.}
    \label{fig:causal-gpt2-dist}
\end{figure}

Figure \ref{fig:causal-opt} shows the analysis of the causal embeddings for OPT. Like GPT-2, it creates dissimilar representations for the ambiguous and the unambiguous prefixes, but here the effect last until the final layer, \textit{i.e.} it will influence the subsequent predictions. Figures \ref{fig:causal-opt-dist} and \ref{fig:causal-gpt2-dist} show directly the distance between stimulus and baseline, without subtracting the counterpart pair. This is to show that, indeed, for GPT-2, the representations of the last layer are very similar, despite the disambiguation in advance.

\section{Details: Incremental Dependency Parsing}
\label{sec:appendix_parsing}
We use the biaffine parser implemented in \texttt{SuPar} (\url{https://github.com/yzhangcs/parser}) with embeddings from RoBERTa large. For the DiaParser, we follow the original implementation (\url{https://github.com/Unipisa/diaparser}) using ELECTRA base. Both parsers use \texttt{stanza} \citep{qi2020stanza} as tokenizers.

\subsection{Computation}
Our method is inspired by the approach of  \citet{hrycyk-etal-2021-fast}. Let us consider $\texttt{head}(i)^{t}$ as the head of token $i$ predicted at time step $t$. When $\texttt{head}(i)^{t} = \texttt{head}(i)^{ref}$, we can directly compare the label attention distribution between the current time step $t$ against the reference time step $ref \in \{0, t-1, T\}$. However when $\texttt{head}(i)^{t} \neq \texttt{head}(i)^{ref}$, either heads may or may not be observed yet. In the case where the head is already observed, we use the label distribution from the self-attention matrix at time step $t$ or $ref$ for comparison depending whether $t < ref$, otherwise we assume a uniform distribution. To be more precise, let assume that in an incremental scenario, the attention distribution for dependency labels at time step $n$ is also available at time step $m$ where $m > n$. We then compute the Jensen-Shannon divergence as the following when $\texttt{head}(i)^{n} \neq \texttt{head}(i)^{m}$:
\begin{align}
    JSD(p(y|\texttt{arc}(i,j)^{n}) || p(y|\overline{\texttt{arc}}(i, j)^{m})) \label{eq:js1} \\
    JSD(p(y|\overline{\texttt{arc}}(i, k)^{n})||p(y|\texttt{arc}(i,k)^{m})) \label{eq:js2}
\end{align}
where $\texttt{arc}(i,j)$ is the arc between token $i$ and $j$, and $p(y|\overline{\texttt{arc}}(i, \cdot))$ is obtained from the self-attention matrix if the token is already observed and uniform distribution otherwise. The distance between the label attention distribution at time step $n$ and $m$ is then defined as the average of (\ref{eq:js1}) and (\ref{eq:js2}).

\subsection{Additional Results}

\begin{figure}[h!]
    \centering

    \begin{subfigure}{\columnwidth}
    \frame{\includegraphics[trim={0cm 0cm 0cm 0cm},clip,width=0.49\columnwidth,page=1]{figures-dep/nps_roberta_last_diff.pdf}}
    \frame{\includegraphics[trim={0cm 0cm 0cm 0cm},clip,width=0.49\columnwidth,page=1]{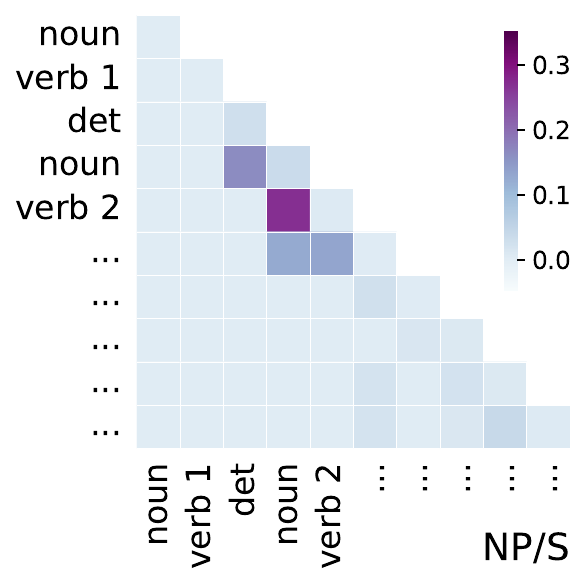}}
    \caption{Biaffine parser}
    \vspace*{.25cm}
    \end{subfigure}
    
    \begin{subfigure}{\columnwidth}
    \frame{\includegraphics[trim={0cm 0cm 0cm 0cm},clip,width=0.49\columnwidth,page=1]{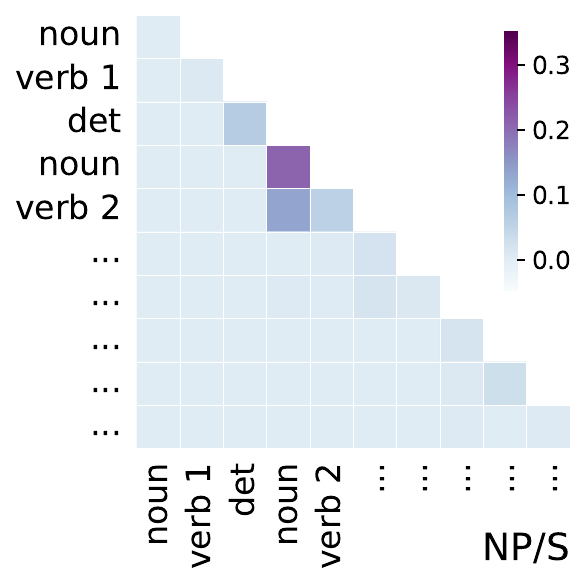}}
    \frame{\includegraphics[trim={0cm 0cm 0cm 0cm},clip,width=0.49\columnwidth,page=1]{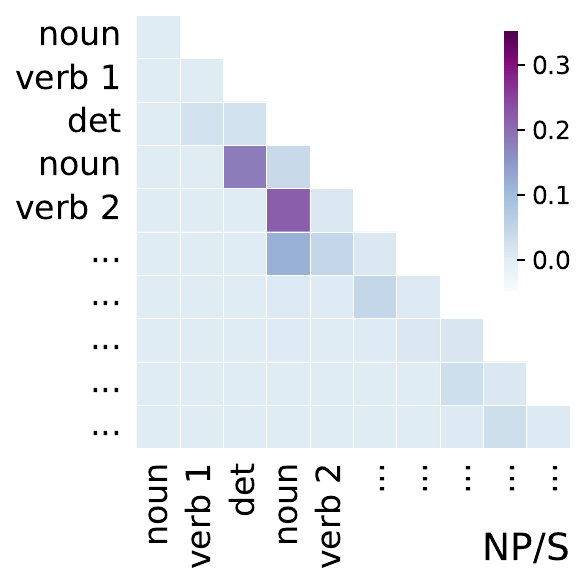}}
    \caption{DiaParser (ELECTRA-EWT)}
    \vspace*{.25cm}
    \end{subfigure}

    \begin{subfigure}{\columnwidth}
    \frame{\includegraphics[trim={0cm 0cm 0cm 0cm},clip,width=0.49\columnwidth,page=1]{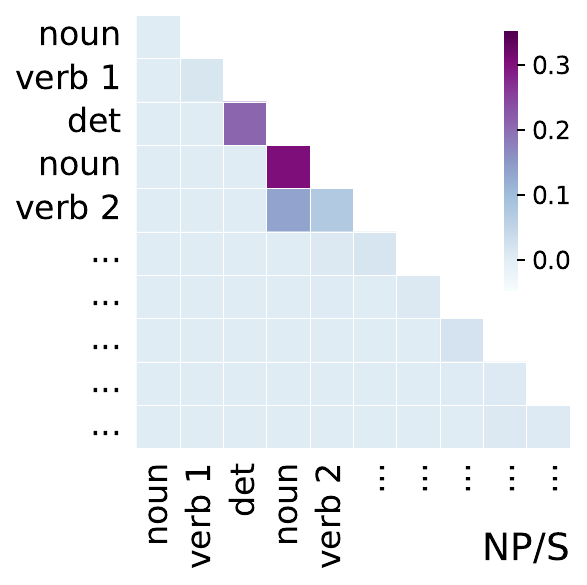}}
    \frame{\includegraphics[trim={0cm 0cm 0cm 0cm},clip,width=0.49\columnwidth,page=1]{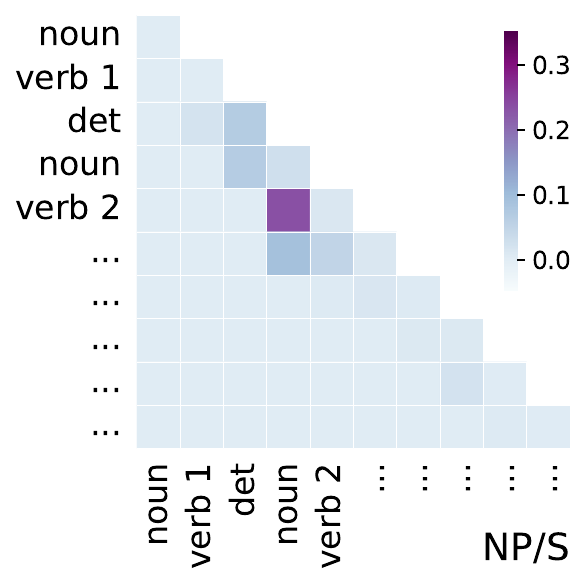}}
    \caption{DiaParser (ELECTRA-PTB)}
    \end{subfigure}

    \caption{Average absolute difference of JSD between stimulus and baseline on NP/S. Left: last step as reference. Right: previous step as reference. We also include the results for biaffine parser for the ease of comparison.}
    \label{fig:nnc-dep-appendix3}
\end{figure}

\begin{figure*}[t]
\centering

\begin{subfigure}[t]{\textwidth}
   \centering
   \frame{\includegraphics[width=0.3\linewidth]{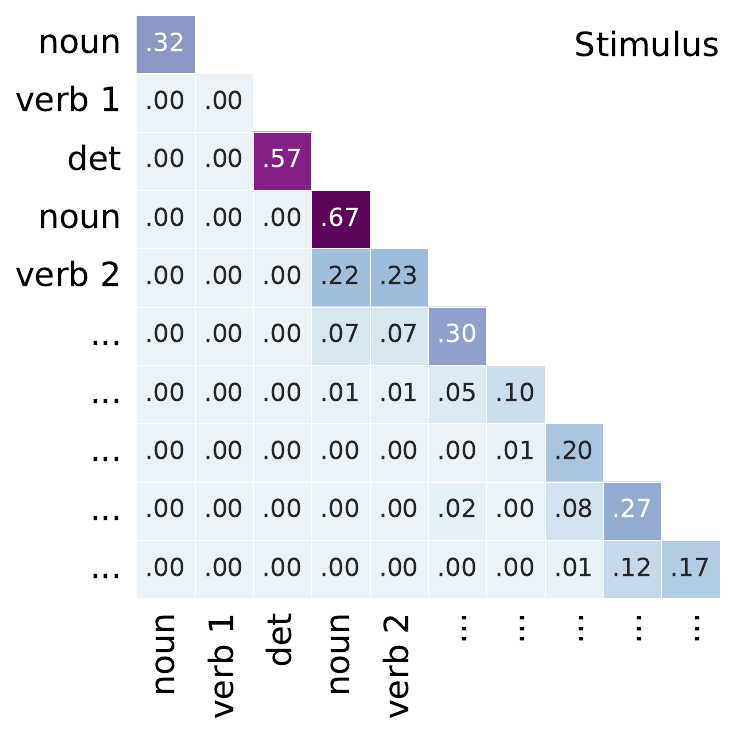}}
    \hfill
  \frame{\includegraphics[width=0.3\linewidth]{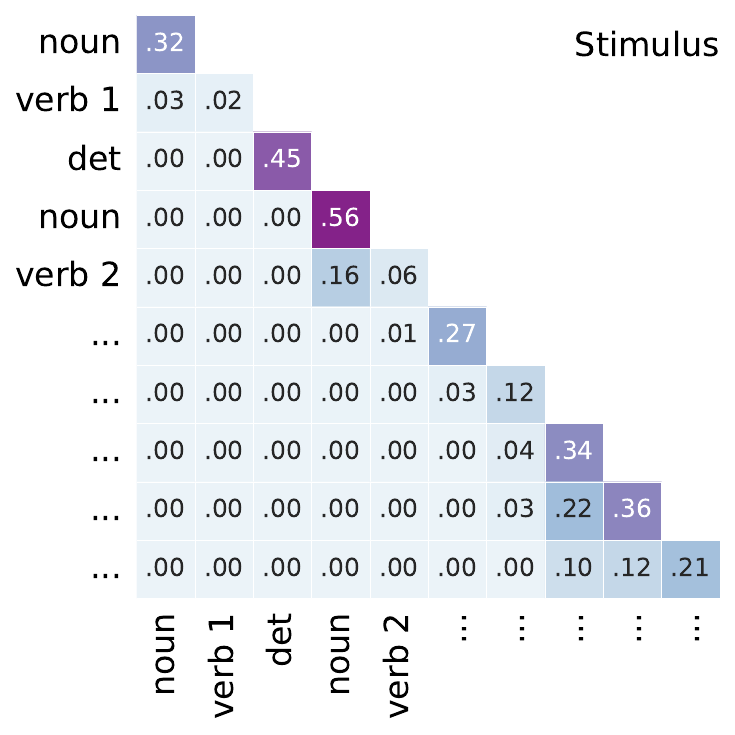}}
   \hfill
   \frame{\includegraphics[width=0.3\linewidth]{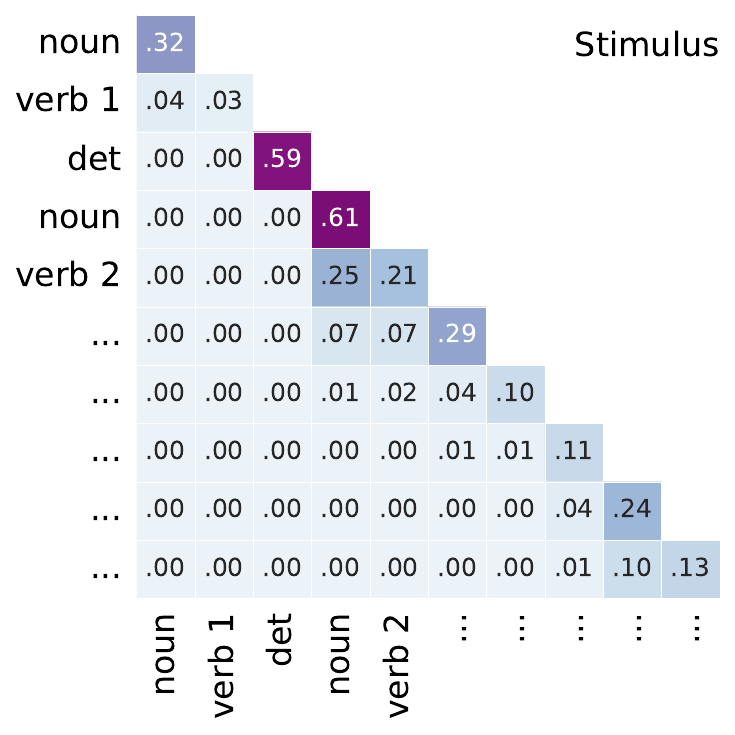}}

   \vspace*{.5cm}
   \frame{\includegraphics[width=0.3\linewidth]{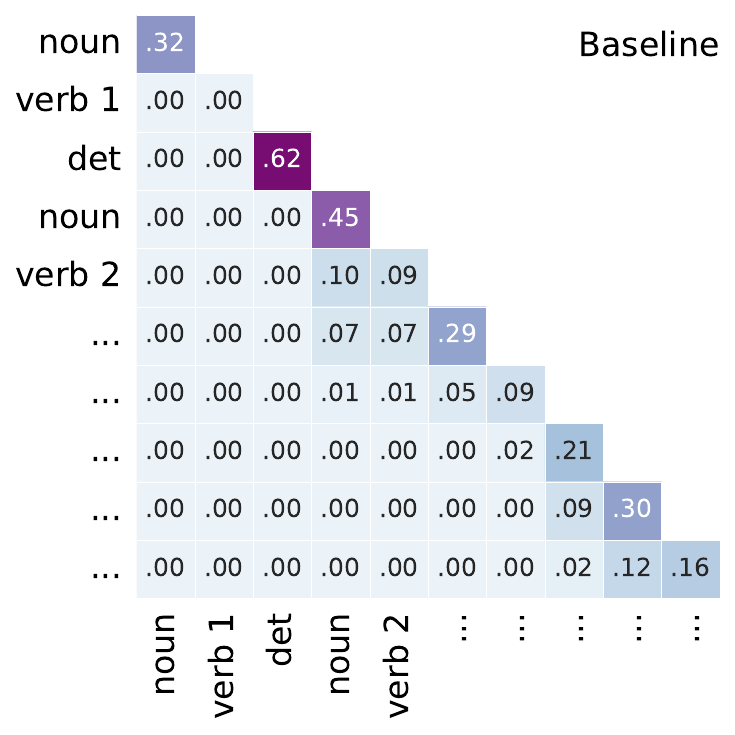}}
   \hfill
   \frame{\includegraphics[width=0.3\linewidth]{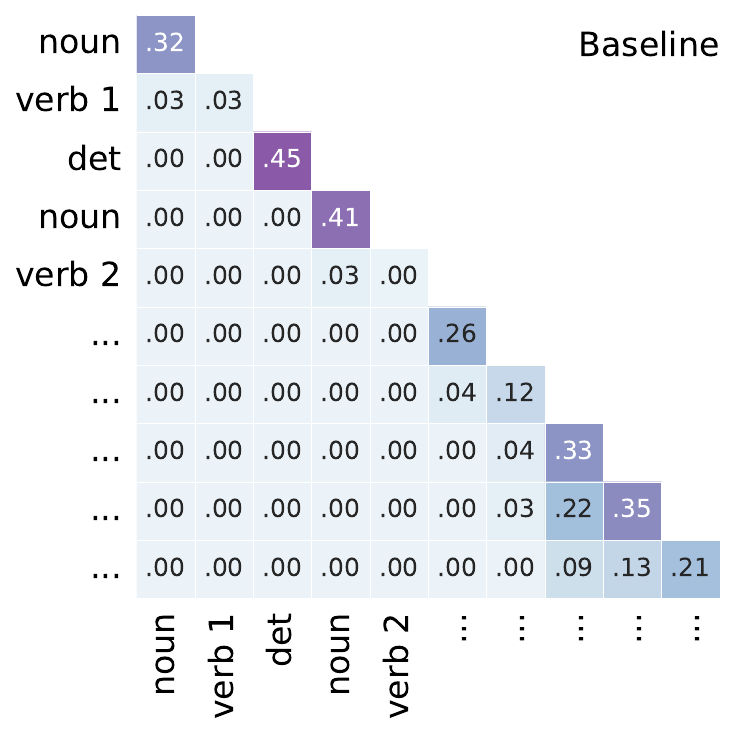}}
   \hfill
   \frame{\includegraphics[width=0.3\linewidth]{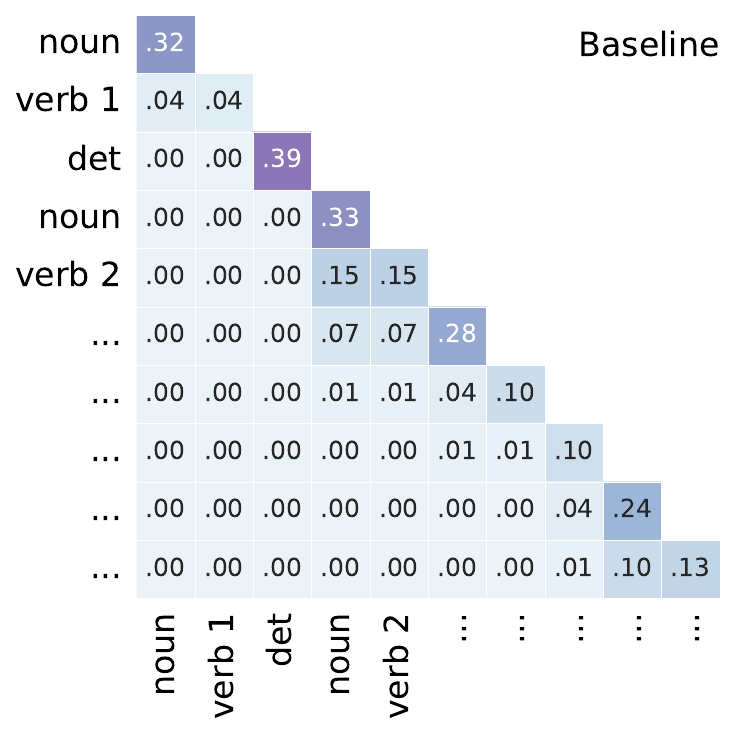}}
   \caption{JS divergence wrt. the last time step}
   \vspace*{.5cm}
\end{subfigure}
\begin{subfigure}[t]{\textwidth}
   \centering
   \frame{\includegraphics[width=0.3\linewidth]{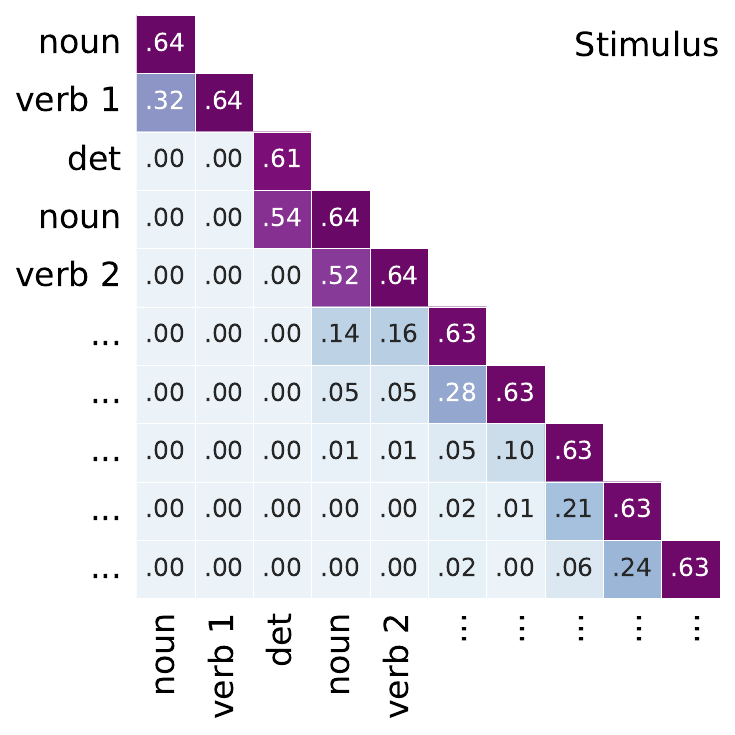}}
    \hfill
  \frame{\includegraphics[width=0.3\linewidth]{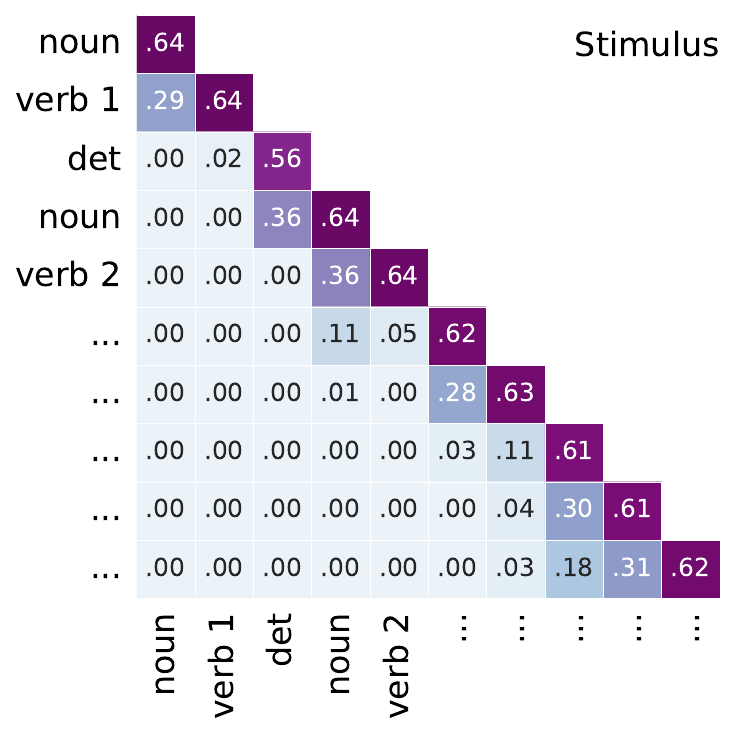}}
   \hfill
   \frame{\includegraphics[width=0.3\linewidth]{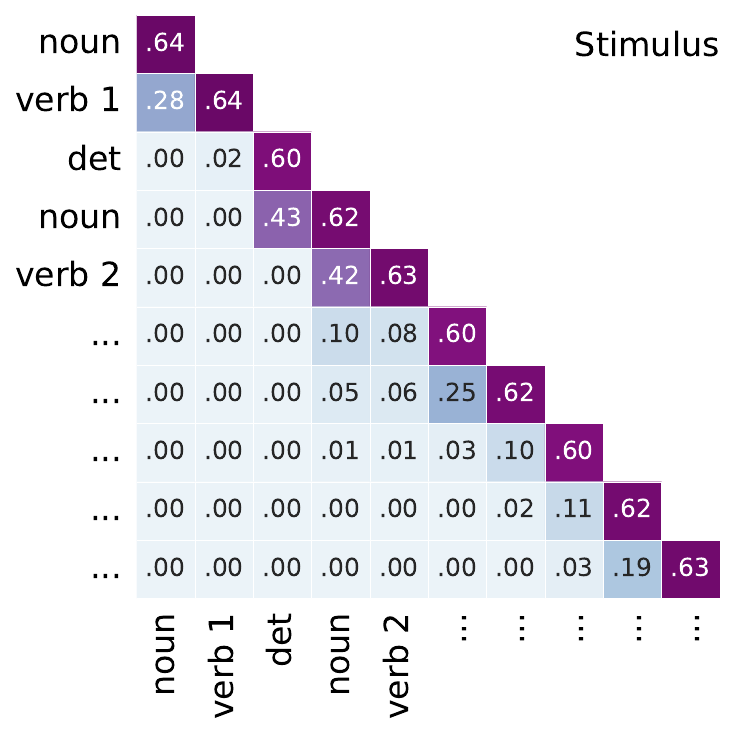}}

   \vspace*{.5cm}
   \frame{\includegraphics[width=0.3\linewidth]{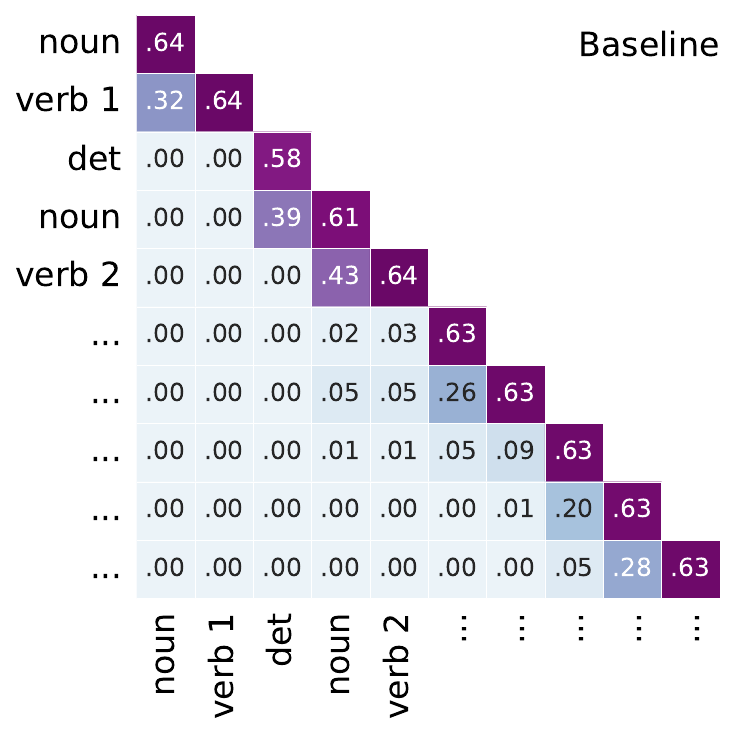}}
   \hfill
   \frame{\includegraphics[width=0.3\linewidth]{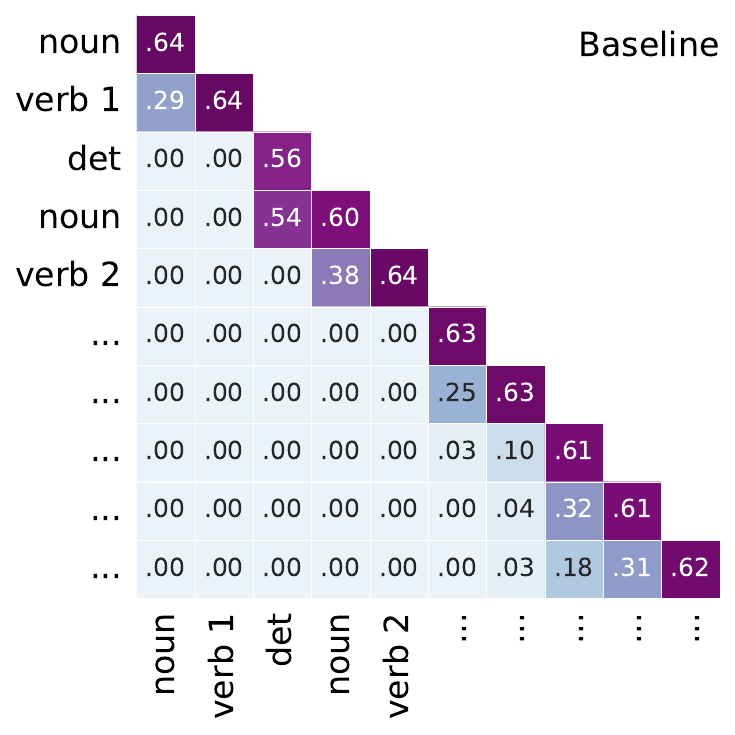}}
   \hfill
   \frame{\includegraphics[width=0.3\linewidth]{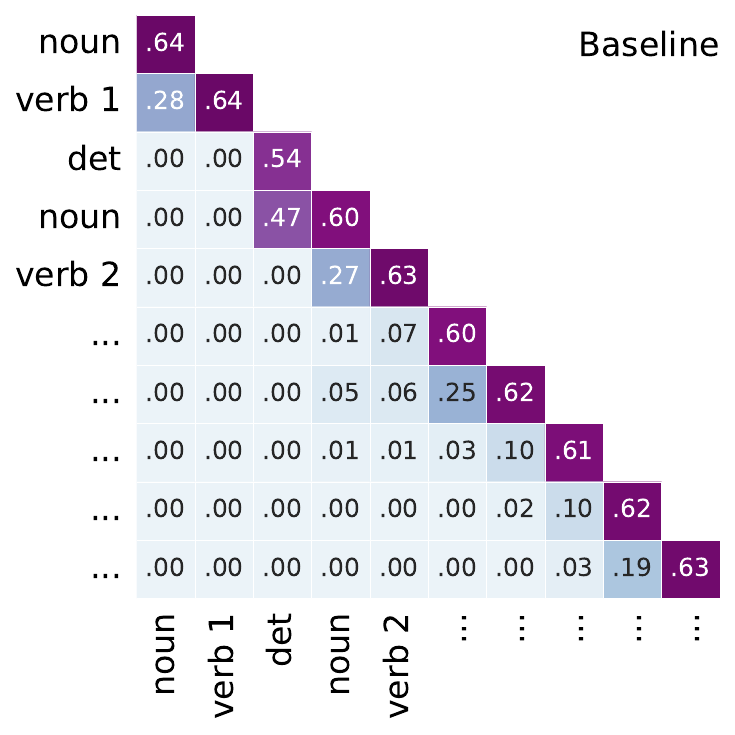}}
   \caption{JS divergence wrt. the previous time step}
   \vspace*{.5cm}
\end{subfigure}
  \caption{Overview of the JSD for the label attention distributions wrt. the last and previous time steps as reference, averaged over all NP/S stimuli and baselines. From left to right: biaffine parser, DiaParser (ELECTRA-EWT and ELECTRA-PTB).}
  
  \label{fig:nnc-dep-appendix4} 
\end{figure*}

\paragraph{NP/S} For the last time step as a reference, we see that the complement's distribution is closer to the final for the baseline compared to the stimulus, showing that the complementiser helps in processing the temporary ambiguity. The overview is depicted in Figure \ref{fig:nnc-dep-appendix4}. We also measure how the label distribution changes as more token is observed by taking the absolute difference of JSD between the stimulus and the baseline wrt. the previous time step. In Figure \ref{fig:nnc-dep-appendix3} (right), we observe a noticeable difference in the divergence of the complement when the disambiguation token (the second verb) is encountered. After that point, the gap diminishes rapidly. 

\vspace{1.22cm}
\begin{figure}[ht]
    \centering

    \begin{subfigure}{\columnwidth}
    \frame{\includegraphics[trim={0cm 0cm 0cm 0cm},clip,width=0.49\columnwidth,page=1]{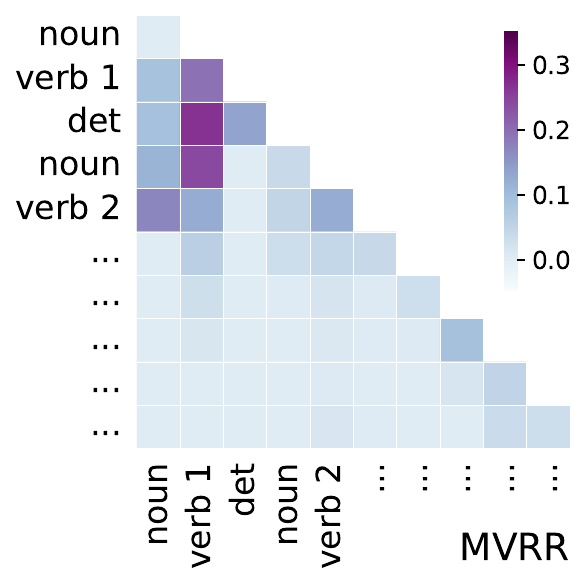}}
    \frame{\includegraphics[trim={0cm 0cm 0cm 0cm},clip,width=0.49\columnwidth,page=1]{figures-dep/mvrr_roberta_prev_diff.pdf}}
    \caption{Biaffine parser}
    \vspace*{.25cm}
    \end{subfigure}
    
    \begin{subfigure}{\columnwidth}
    \frame{\includegraphics[trim={0cm 0cm 0cm 0cm},clip,width=0.49\columnwidth,page=1]{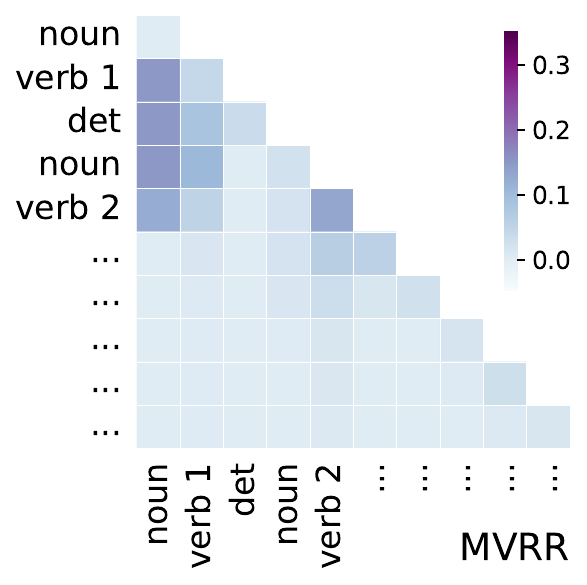}}
    \frame{\includegraphics[trim={0cm 0cm 0cm 0cm},clip,width=0.49\columnwidth,page=1]{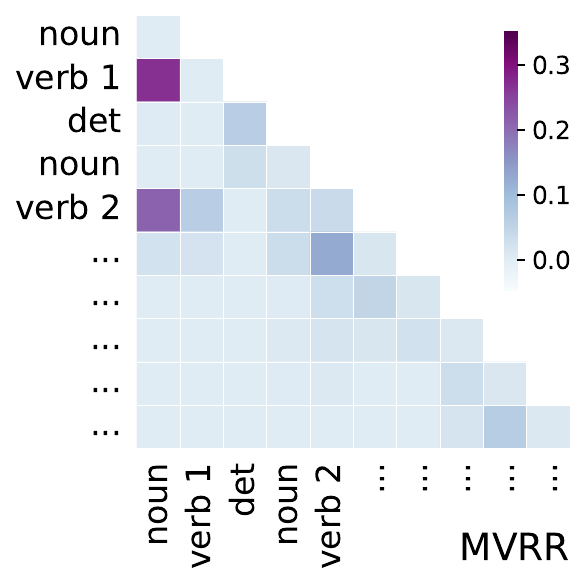}}
    \caption{DiaParser (ELECTRA-EWT)}
    \vspace*{.25cm}
    \end{subfigure}

    \begin{subfigure}{\columnwidth}
    \frame{\includegraphics[trim={0cm 0cm 0cm 0cm},clip,width=0.49\columnwidth,page=1]{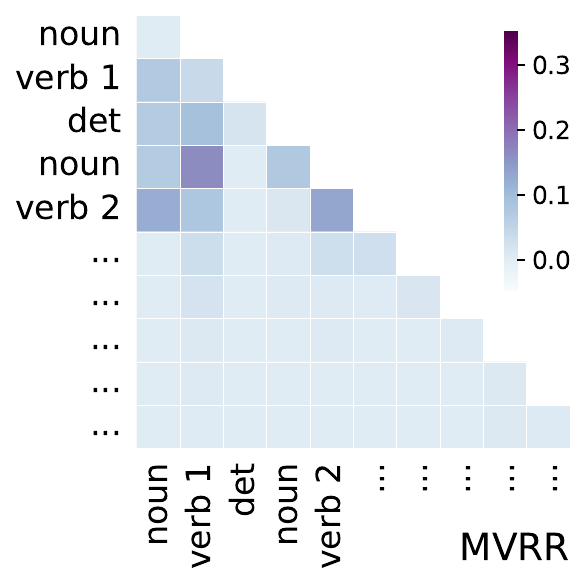}}
    \frame{\includegraphics[trim={0cm 0cm 0cm 0cm},clip,width=0.49\columnwidth,page=1]{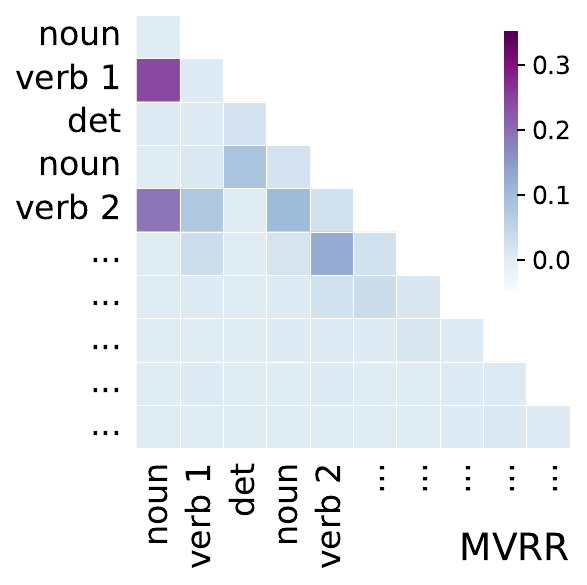}}
    \caption{DiaParser (ELECTRA-PTB)}
    \end{subfigure}

    \caption{Average absolute difference of JSD between stimulus and baseline on MVRR. Left: last step as reference. Right: previous step as reference. We also include the results for biaffine parser for the ease of comparison.}
    \label{fig:nnc-dep-appendix6}
\end{figure}

\paragraph{MVRR} We compute the absolute difference of JSD wrt. the last step as the stimulus and the baseline have the same final interpretation. In Figure \ref{fig:nnc-dep-appendix6} (left), we observe that the first noun and the first verb are processed differently in both cases. This is highly likely due to the fact that the first verb can be interpreted as the main verb or as a reduced relative, which also affects how the first noun is understood. We see that both the stimulus and the baseline converge to the same interpretation after the disambiguation token (the second verb) is observed.

\paragraph{Alignment} In addition to MCC, we also compute the average precision (AP) score for both the stimulus and the baseline. We do this by treating edits as the ground truth and the JSD wrt. previous step as the prediction to see if edits can be predicted just from JSD alone (Table \ref{tab:alignment-appendix1}). We observe that the AP for dependency arcs are high in general, while it is lower for the labels. We also see that the average edit ratio is low for all stimuli except NNC, as shown in Table \ref{tab:alignment-appendix2}.

\begin{table}[ht]
\centering
\small
{\setlength{\tabcolsep}{3pt}
\begin{tabular}{l c c c c c c}
    \toprule
         & \multicolumn{2}{c}{NNC} & \multicolumn{2}{c}{NP/S} & \multicolumn{2}{c}{MVRR} \\
    \midrule
    \textbf{Arc} & S & B & S & B & S & B \\
        \cmidrule(lr){2-3} \cmidrule(lr){4-5} \cmidrule(lr){6-7}
        \hspace{0.2cm} Biaffine & 0.99 & 0.99 & 0.92 & 0.88 & 0.88 & 0.89  \\
        \hspace{0.2cm} DiaParser (EWT) & 0.99 & 1.00 & 0.98 & 0.96 & 0.91 & 0.97 \\
        \hspace{0.2cm} DiaParser (PTB) & 0.99 & 0.99 & 0.97 & 0.98 & 0.92 & 0.99 \\
    \midrule
    \textbf{Label} & S & B & S & B & S & B \\
        \cmidrule(lr){2-3} \cmidrule(lr){4-5} \cmidrule(lr){6-7}
        \hspace{0.2cm} Biaffine & 0.41 & 0.90 & 0.92 & 0.78 & 0.80 & 0.85  \\
        \hspace{0.2cm} DiaParser (EWT) & 0.30 & 0.48 & 0.86 & 0.76 & 0.75 & 0.87 \\
        \hspace{0.2cm} DiaParser (PTB) & 0.55 & 0.80 & 0.89 & 0.74 & 0.78 & 0.91 \\
    \bottomrule
\end{tabular}
}
\caption{Average Precision between JSD wrt. previous time step and edits for dependency arcs and labels. S: stimulus \& B: baseline.
}
\label{tab:alignment-appendix1}
\end{table}

\begin{table}[ht]
\centering
\small
{\setlength{\tabcolsep}{3pt}
\begin{tabular}{l c c c c c c}
    \toprule
         & \multicolumn{2}{c}{NNC} & \multicolumn{2}{c}{NP/S} & \multicolumn{2}{c}{MVRR} \\
    \midrule
    \textbf{Arc} & S & B & S & B & S & B \\
        \cmidrule(lr){2-3} \cmidrule(lr){4-5} \cmidrule(lr){6-7}
        \hspace{0.2cm} Biaffine & 0.53 & 0.41 & 0.11 & 0.11 & 0.12 & 0.10  \\
        \hspace{0.2cm} DiaParser (EWT) & 0.67 & 0.60 & 0.13 & 0.13 & 0.15 & 0.11 \\
        \hspace{0.2cm} DiaParser (PTB) & 0.53 & 0.40 & 0.11 & 0.12 & 0.12 & 0.09 \\
    \midrule
    \textbf{Label} & S & B & S & B & S & B \\
        \cmidrule(lr){2-3} \cmidrule(lr){4-5} \cmidrule(lr){6-7}
        \hspace{0.2cm} Biaffine & 0.27 & 0.31 & 0.09 & 0.08 & 0.09 & 0.09  \\
        \hspace{0.2cm} DiaParser (EWT) & 0.27 & 0.30 & 0.09 & 0.08 & 0.11 & 0.08 \\
        \hspace{0.2cm} DiaParser (PTB) & 0.27 & 0.30 & 0.09 & 0.08 & 0.10 & 0.08 \\
    \bottomrule
\end{tabular}
}
\caption{Average edit ratio for dependency arcs and labels. S: stimulus \& B: baseline.
}
\label{tab:alignment-appendix2}
\end{table}

\newpage

\begin{figure}[ht]
    \centering
    \begin{subfigure}{\columnwidth}
    \frame{\includegraphics[trim={0cm 0cm 0cm 0cm},clip,width=0.49\columnwidth,page=1]{figures-dep/nnc_roberta_first_diff.pdf}}
    \frame{\includegraphics[trim={0cm 0cm 0cm 0cm},clip,width=0.49\columnwidth,page=1]{figures-dep/nnc_roberta_prev_diff.pdf}}
    \caption{Biaffine parser}
    \vspace*{.25cm}
    \end{subfigure}
    
    \begin{subfigure}{\columnwidth}
    \frame{\includegraphics[trim={0cm 0cm 0cm 0cm},clip,width=0.49\columnwidth,page=1]{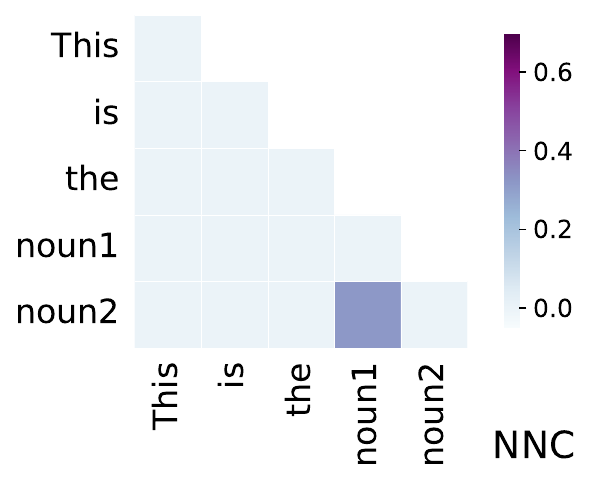}}
    \frame{\includegraphics[trim={0cm 0cm 0cm 0cm},clip,width=0.49\columnwidth,page=1]{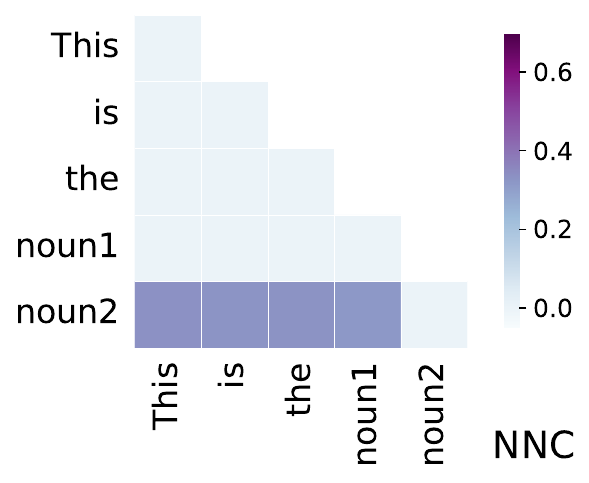}}
    \caption{DiaParser (ELECTRA-EWT)}
    \vspace*{.25cm}
    \end{subfigure}

    \begin{subfigure}{\columnwidth}
    \frame{\includegraphics[trim={0cm 0cm 0cm 0cm},clip,width=0.49\columnwidth,page=1]{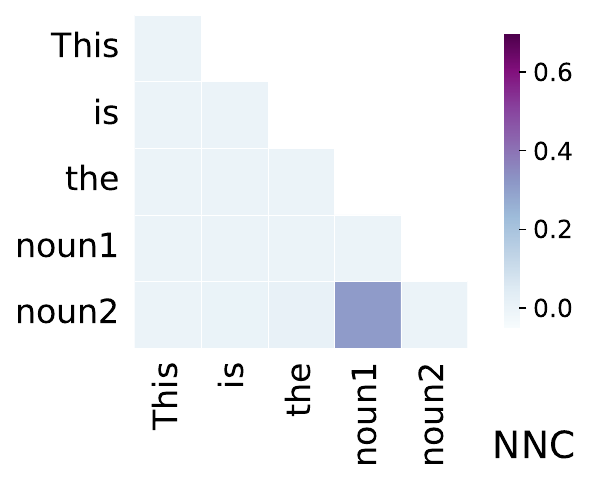}}
    \frame{\includegraphics[trim={0cm 0cm 0cm 0cm},clip,width=0.49\columnwidth,page=1]{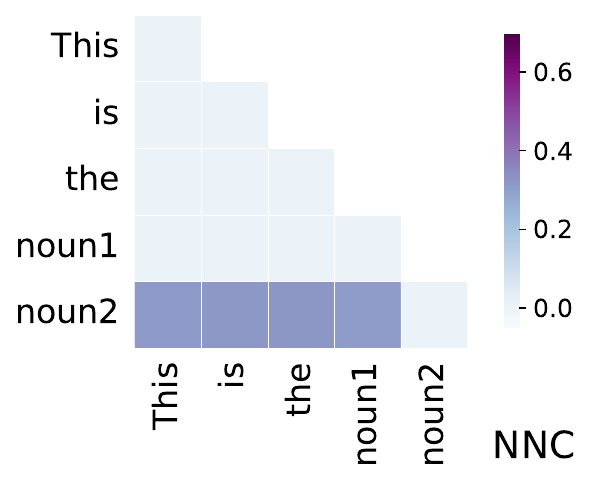}}
    \caption{DiaParser (ELECTRA-PTB)}
    \end{subfigure}

    \caption{Average absolute difference of JSD between stimulus and baseline on NNC. Left: first step as reference. Right: previous step as reference. We also include the results for biaffine parser for the ease of comparison and find similar results to hold across both parsers.}
    \label{fig:nnc-dep-appendix1}
\end{figure}

\begin{figure*}[t]
\centering

\begin{subfigure}[t]{\textwidth}
   \centering
   \frame{\includegraphics[width=0.3\linewidth]{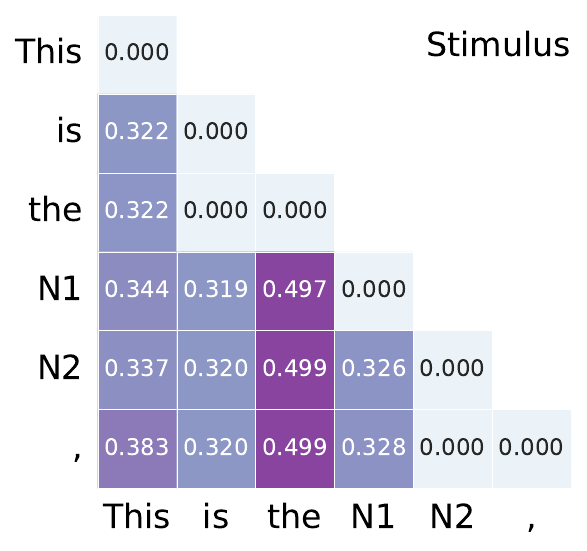}}
    \hfill
  \frame{\includegraphics[width=0.3\linewidth]{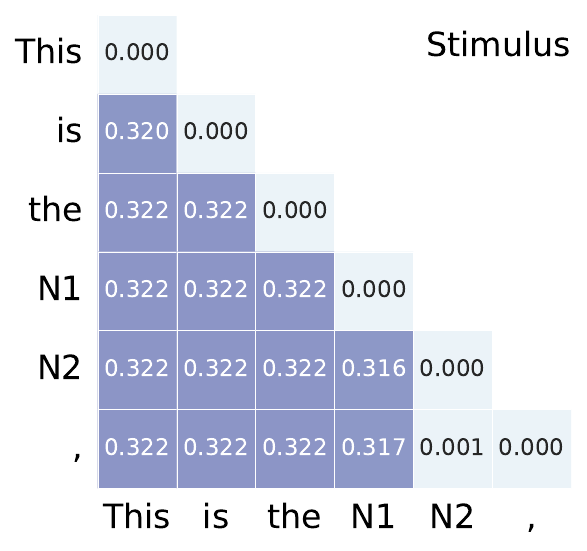}}
   \hfill
   \frame{\includegraphics[width=0.3\linewidth]{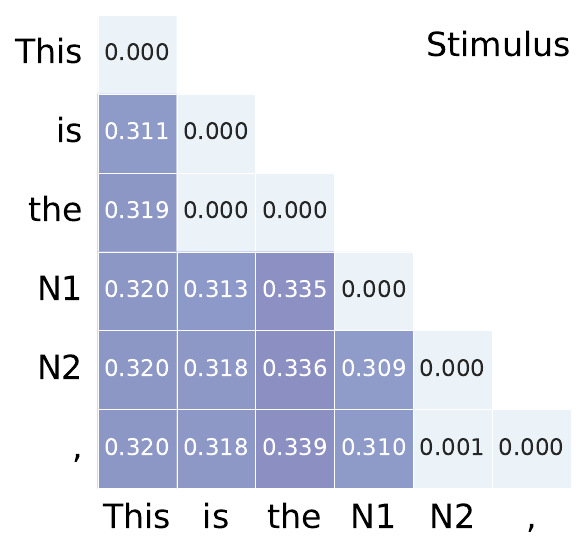}}

   \vspace*{.5cm}
   \frame{\includegraphics[width=0.3\linewidth]{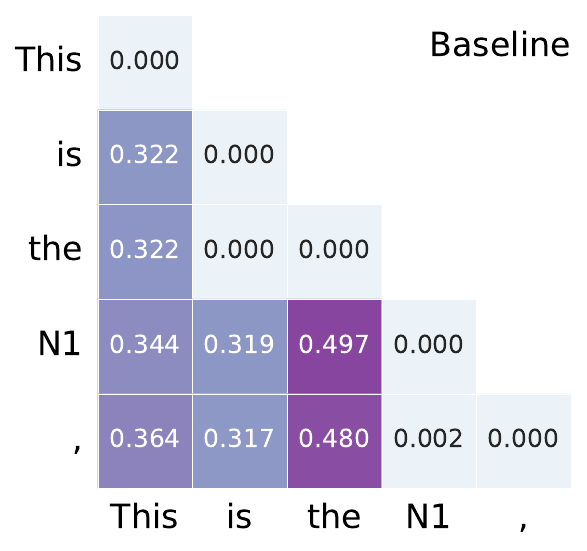}}
   \hfill
   \frame{\includegraphics[width=0.3\linewidth]{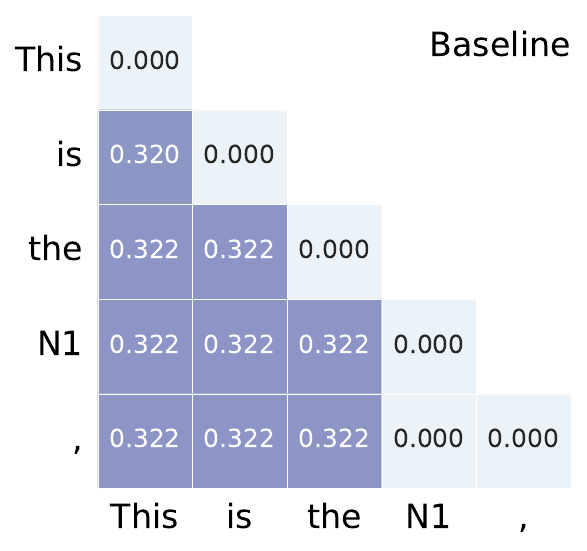}}
   \hfill
   \frame{\includegraphics[width=0.3\linewidth]{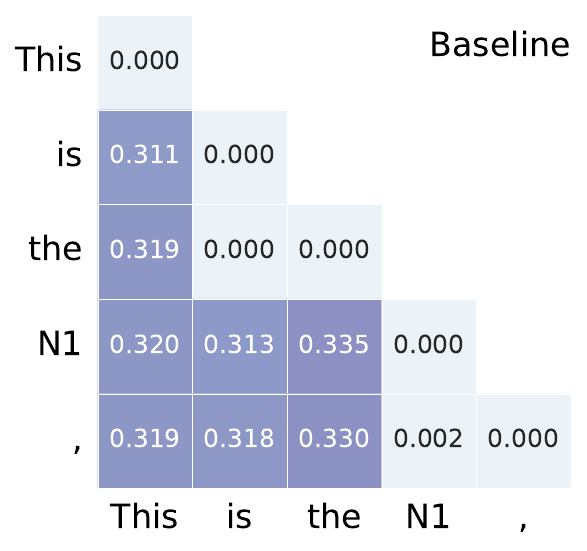}}
   \caption{JS divergence wrt. the first time step}
   \vspace*{.5cm}
\end{subfigure}
\begin{subfigure}[t]{\textwidth}
   \centering
   \frame{\includegraphics[width=0.3\linewidth]{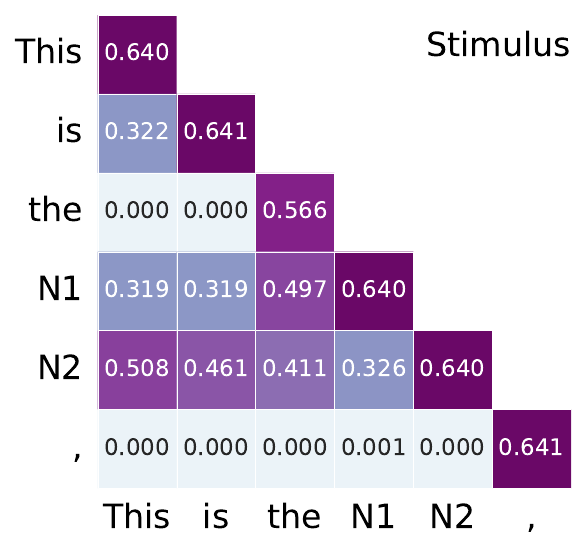}}
    \hfill
  \frame{\includegraphics[width=0.3\linewidth]{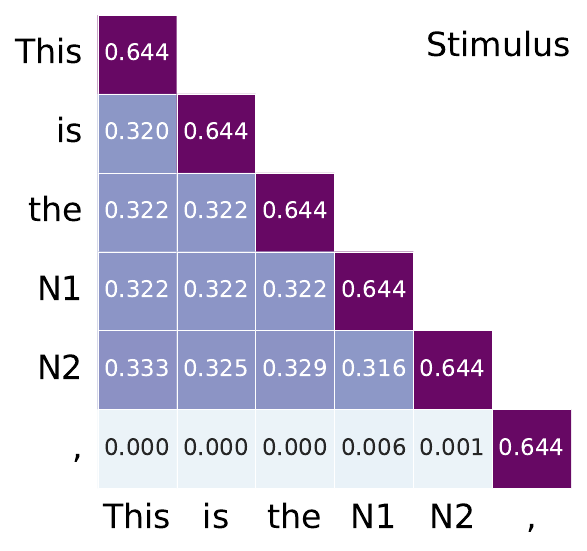}}
   \hfill
   \frame{\includegraphics[width=0.3\linewidth]{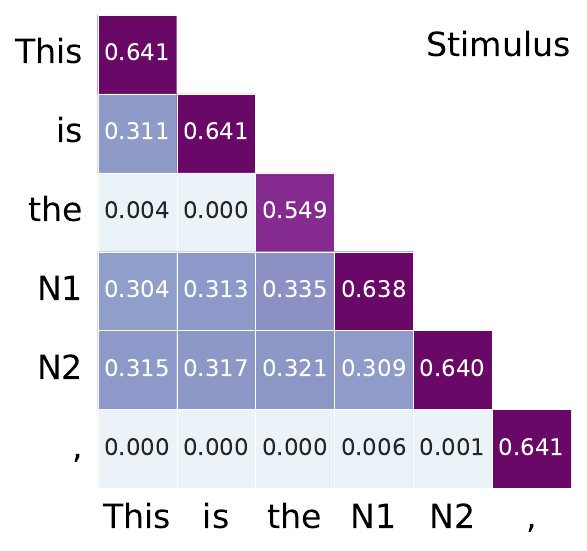}}

   \vspace*{.5cm}
   \frame{\includegraphics[width=0.3\linewidth]{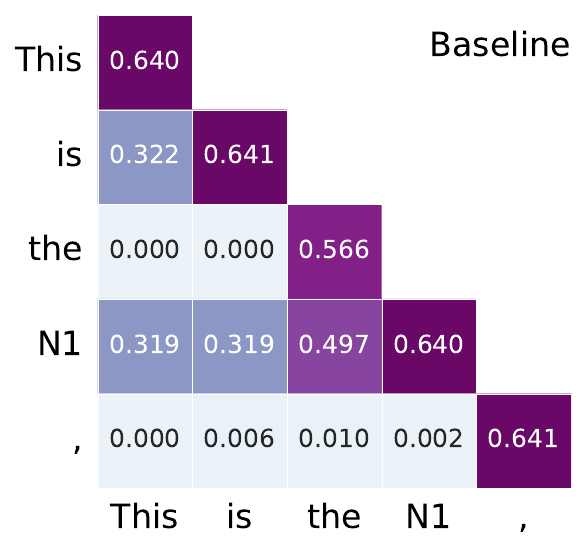}}
   \hfill
   \frame{\includegraphics[width=0.3\linewidth]{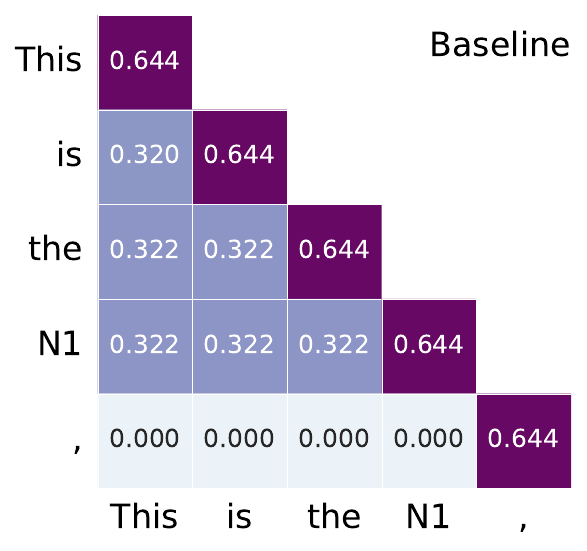}}
   \hfill
   \frame{\includegraphics[width=0.3\linewidth]{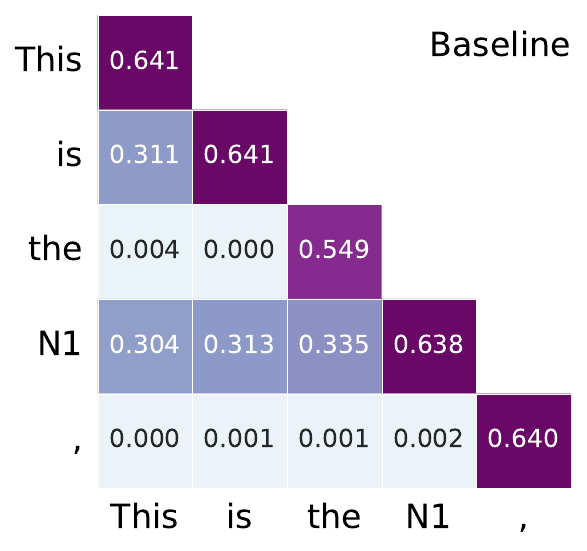}}
   \caption{JS divergence wrt. the previous time step}
   \vspace*{.5cm}
\end{subfigure}
  \caption{Overview of the JSD for the label attention distributions wrt. the first and previous time steps as reference, averaged over all NNC stimuli and baselines. From left to right: biaffine parser, DiaParser (ELECTRA-EWT and ELECTRA-PTB).}
  
  \label{fig:nnc-dep-appendix2} 
\end{figure*}

\begin{figure*}[t]
\centering

\begin{subfigure}[t]{\textwidth}
   \centering
   \frame{\includegraphics[width=0.3\linewidth]{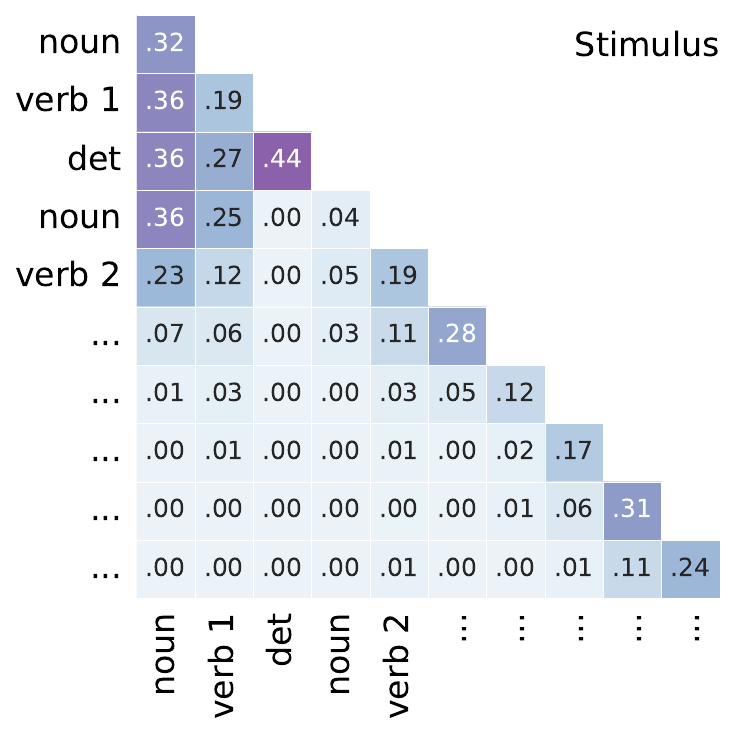}}
    \hfill
  \frame{\includegraphics[width=0.3\linewidth]{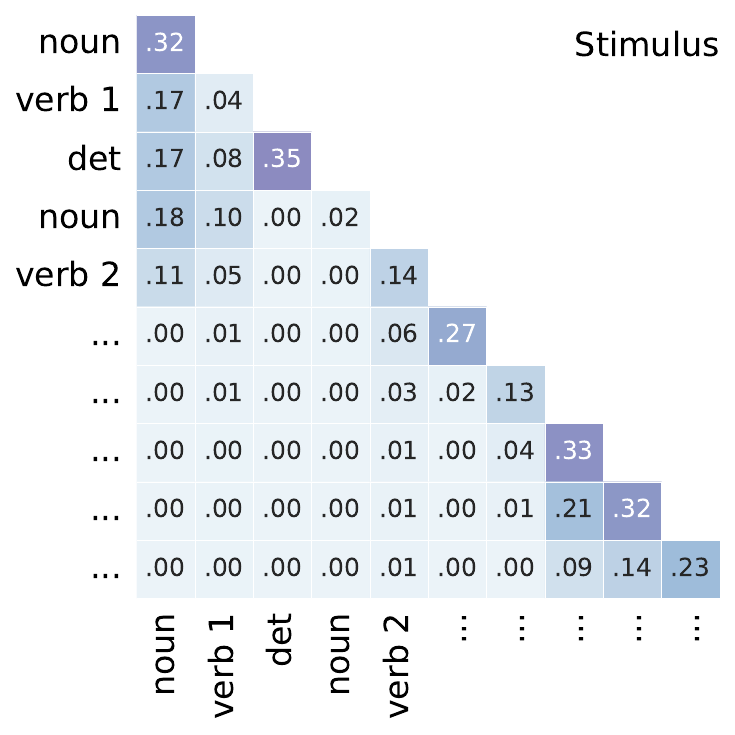}}
   \hfill
   \frame{\includegraphics[width=0.3\linewidth]{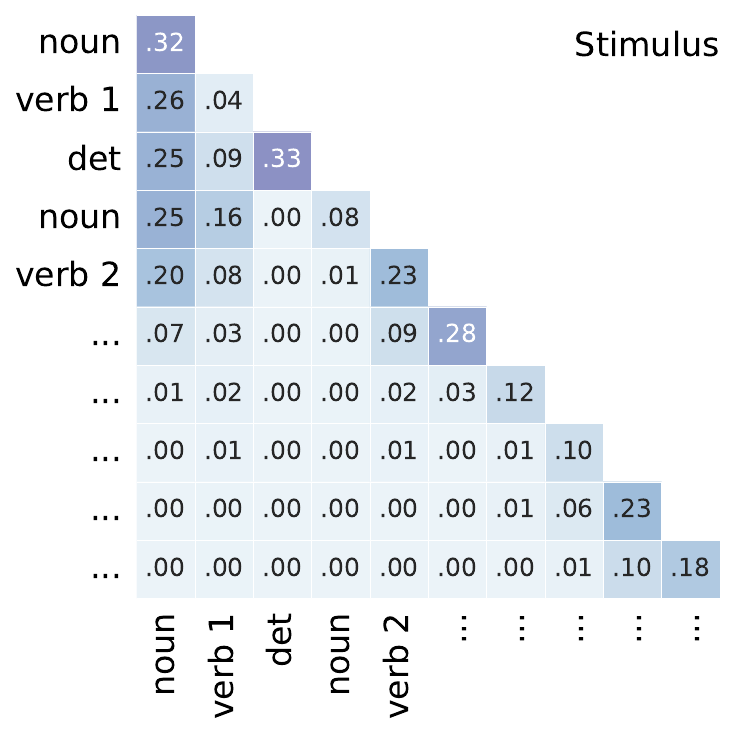}}

   \vspace*{.5cm}
   \frame{\includegraphics[width=0.3\linewidth]{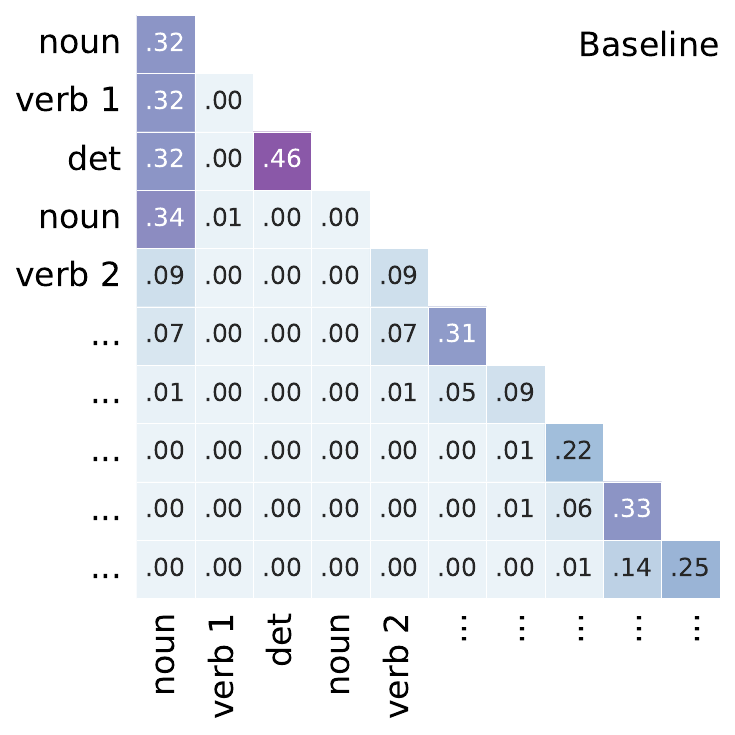}}
   \hfill
   \frame{\includegraphics[width=0.3\linewidth]{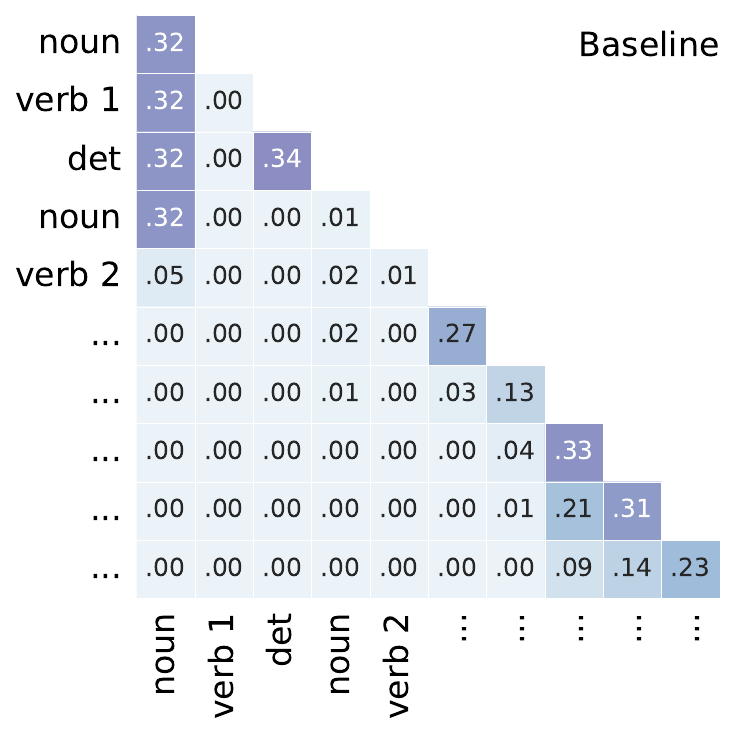}}
   \hfill
   \frame{\includegraphics[width=0.3\linewidth]{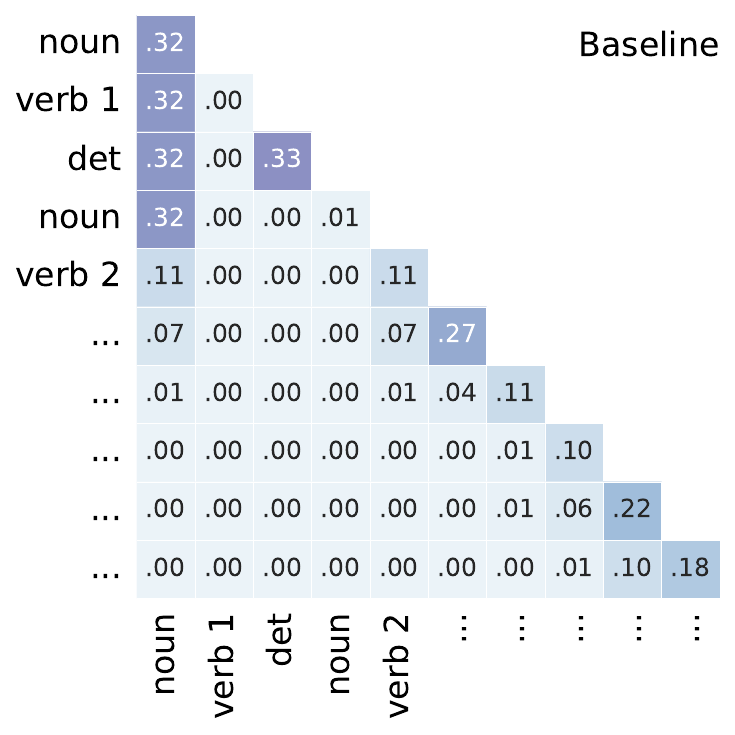}}
   \caption{JS divergence wrt. the last time step}
   \vspace*{.5cm}
\end{subfigure}
\begin{subfigure}[t]{\textwidth}
   \centering
   \frame{\includegraphics[width=0.3\linewidth]{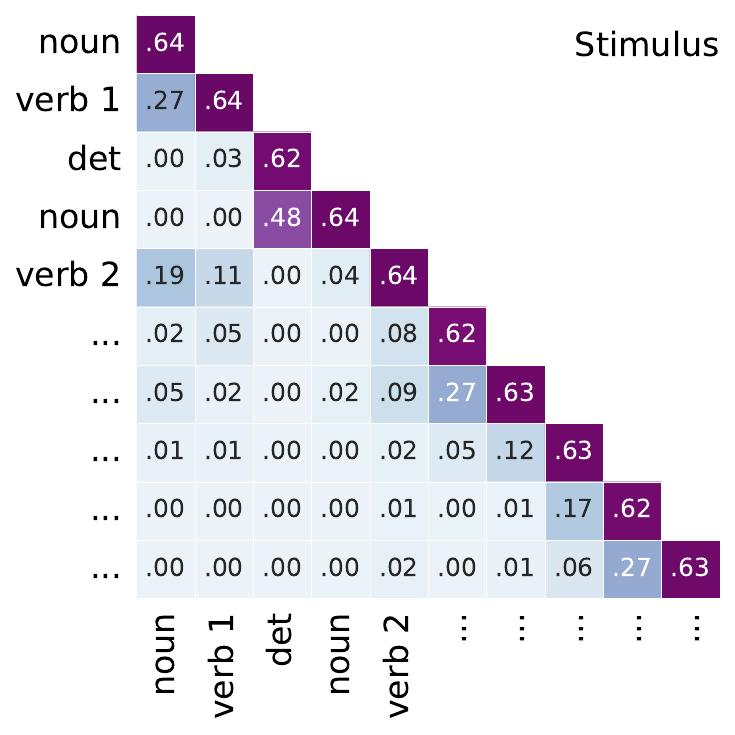}}
    \hfill
  \frame{\includegraphics[width=0.3\linewidth]{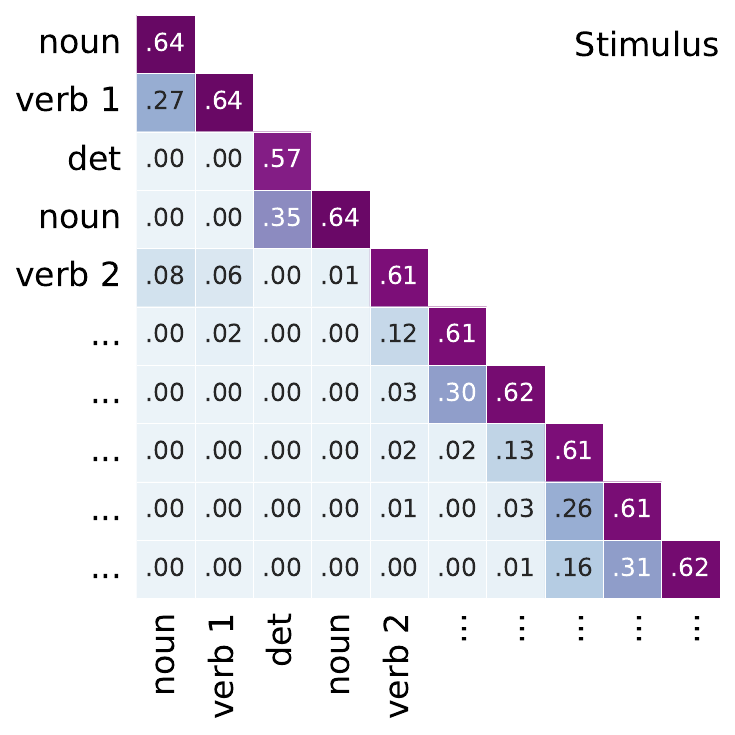}}
   \hfill
   \frame{\includegraphics[width=0.3\linewidth]{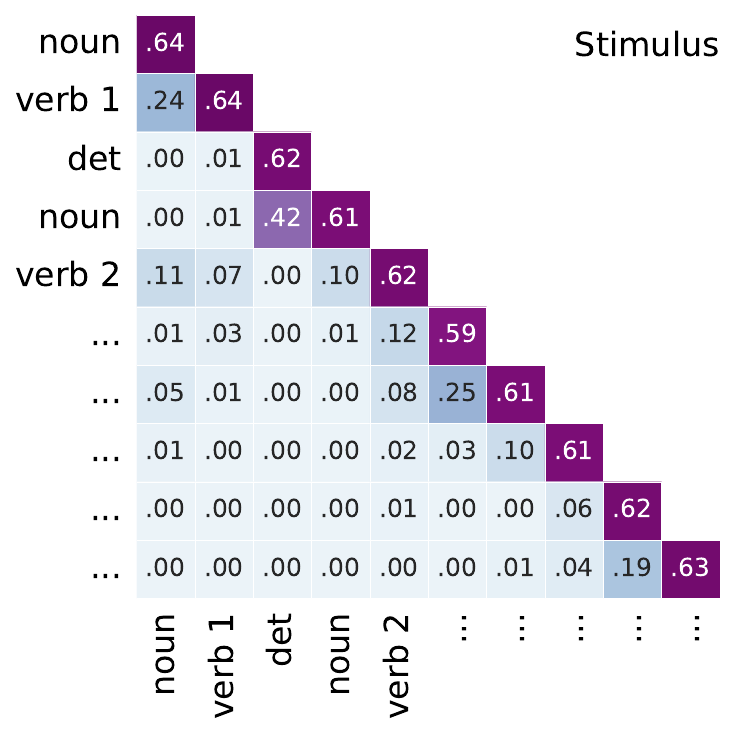}}

   \vspace*{.5cm}
   \frame{\includegraphics[width=0.3\linewidth]{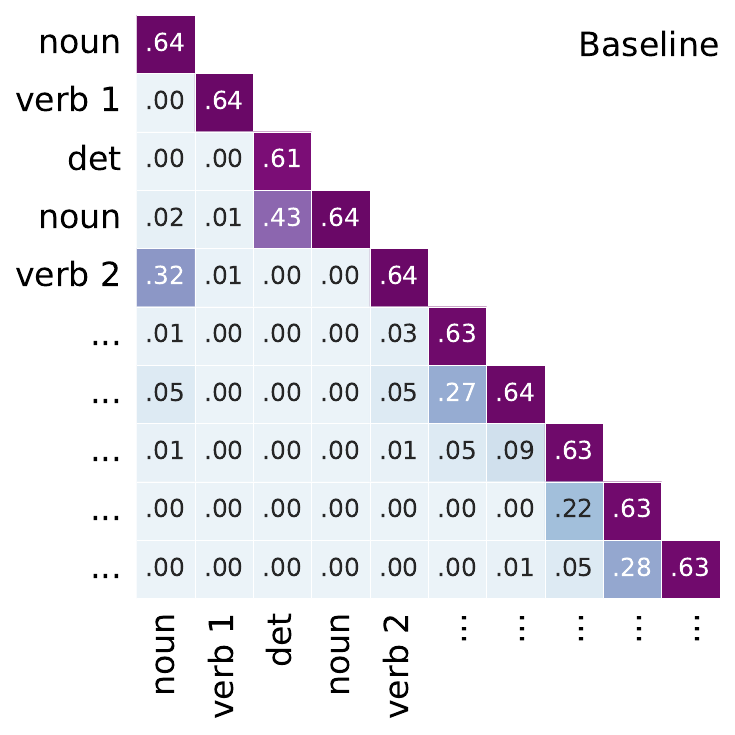}}
   \hfill
   \frame{\includegraphics[width=0.3\linewidth]{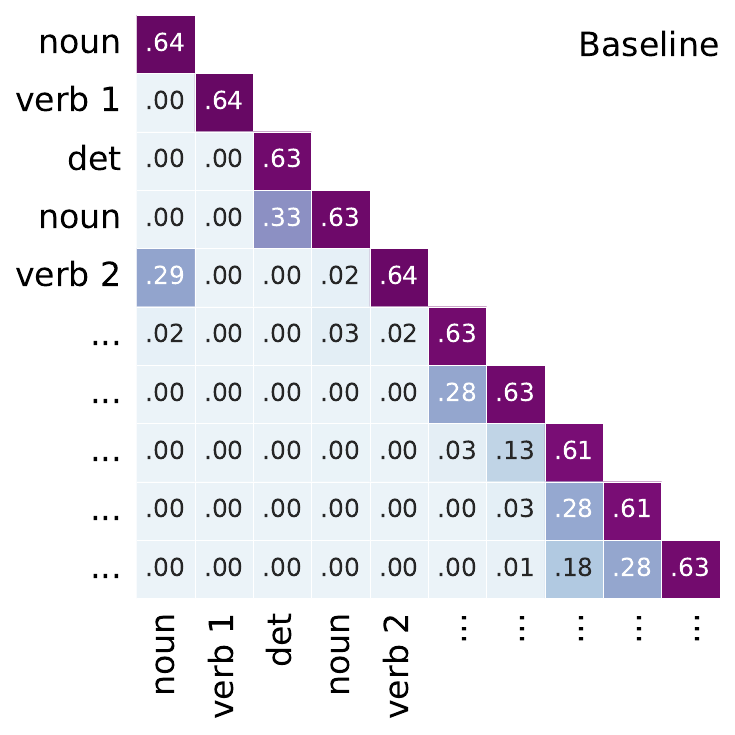}}
   \hfill
   \frame{\includegraphics[width=0.3\linewidth]{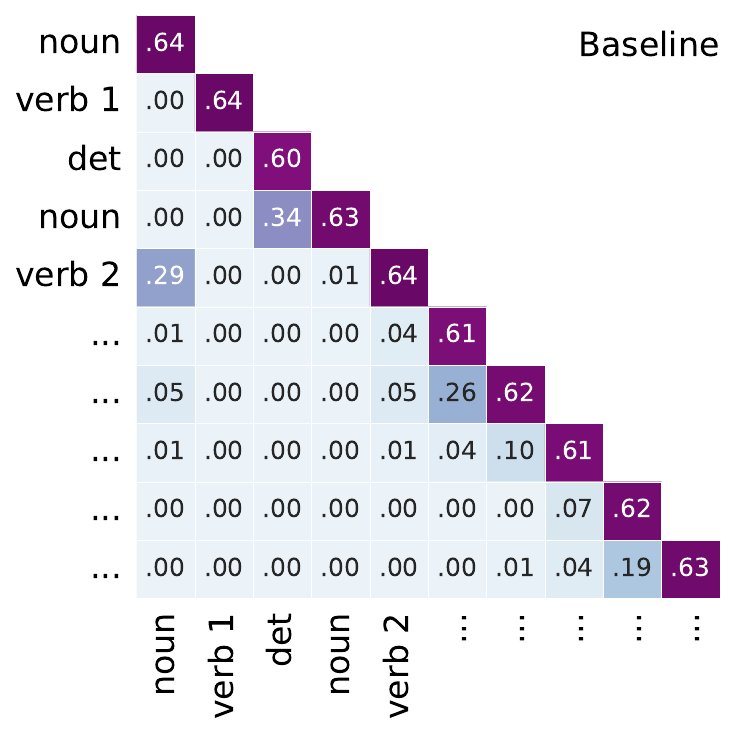}}
   \caption{JS divergence wrt. the previous time step}
   \vspace*{.5cm}
\end{subfigure}
  \caption{Overview of the JSD for the label attention distributions wrt. the last and previous time steps as reference, averaged over all MVRR stimuli and baselines. From left to right: biaffine parser, DiaParser (ELECTRA-EWT and ELECTRA-PTB).}
  
  \label{fig:nnc-dep-appendix5} 
\end{figure*}

\end{document}